\pdfoutput=1
\documentclass[11pt,a4paper,logo]{googledeepmind}
\theoremstyle{remark}
\newtheorem{remark}{Remark}
\usepackage[authoryear,sort&compress,round]{natbib}
\usepackage{amsmath,amsfonts,bm}

\def\eqref#1{equation~\ref{#1}}

\def\1{\bm{1}}

\DeclareMathAlphabet{\mathsfit}{\encodingdefault}{\sfdefault}{m}{sl}
\SetMathAlphabet{\mathsfit}{bold}{\encodingdefault}{\sfdefault}{bx}{n}

\usepackage{multirow}
\usepackage{subcaption}
\usepackage{placeins}
\usepackage{wrapfig}
\usepackage{tikz}
\usetikzlibrary{arrows.meta, positioning, fit}
\usepackage{longtable}
\usepackage{algorithm}
\usepackage{algpseudocode}
\usepackage{makecell}

\definecolor{gblue}{HTML}{4285F4}
\definecolor{gred}{HTML}{EA4335}
\definecolor{gyellow}{HTML}{F9AB00}
\definecolor{ggreen}{HTML}{34A853}
\definecolor{okgreen}{HTML}{188038}
\definecolor{nored}{HTML}{C5221F}
\definecolor{royalblue}{HTML}{4169E1}
\definecolor{sectiongray}{HTML}{EDEDED}
\newcommand{\cmark}{\textcolor{okgreen}{\ding{51}}}
\newcommand{\hib}{\textcolor{okgreen}{$\boldsymbol{\uparrow}$}}
\newcommand{\lob}{\textcolor{okgreen}{$\boldsymbol{\downarrow}$}}
\newcommand{\xmark}{\textcolor{nored}{\ding{55}}}

\newcommand{\cci}{\textsc{CCI}}
\newcommand{\buggymle}{\textsc{BuggyMLE}}
\newcommand{\bmbl}{Buggy\-MLE\-Bench-Lite}
\title{\buggymle{}: Scaling Machine Learning Engineering Difficulty with Bug Injection}

\author[1,3]{John Wu}
\author[2,4]{Rushi Qiang}
\author[2]{Yuchen Zhuang}
\author[2]{Guowang Li}
\author[1]{Girish Narayanswamy}
\author[1]{Max Xu}
\author[2]{Sam Schmidgall}
\author[1]{Mark Malhotra}
\author[1]{Shwetak Patel}
\author[3]{Jimeng Sun}
\author[1]{Daniel McDuff}
\author[2,4,$\dagger$]{Bo Dai}
\author[1,$\dagger$]{Xin Liu}
\affil[1]{Google Research}
\affil[2]{Google DeepMind}
\affil[3]{Siebel School of Computing and Data Science, University of Illinois Urbana-Champaign}
\affil[4]{Georgia Institute of Technology College of Computing}
\affil[$\dagger$]{Co-senior authors}
\correspondingauthor{johnwu3@illinois.edu, xliucs@google.com}
\reportnumber{}

\begin{abstract}
Enabling AI systems to autonomously improve AI models requires agents capable of performing machine learning engineering (MLE).
In practice, designing, training, and evaluating complex models demands modifying multi-file repositories to reproduce results, debug training runs, integrate datasets, and port frameworks. However, existing MLE benchmarks limit agents to single-file edits without full repository or shell access, and common repository-level software engineering benchmarks have rapidly saturated. To address this, we introduce the Code Complexity Index (\cci{}), a difficulty metric based on ten structural features spanning repository scale, dependency depth, and patch size. \cci{} strongly predicts average solve rates across eight public coding benchmarks ($R^2 = 0.951$), identifying patch size and complexity as top predictors. Guided by \cci{}, we develop \buggymle{}, a scalable framework where agents inject controlled mutations into working ML repositories, synthesizing over 6{,}600 bug-injected MLE repository variants and nearly 260K lines of injected code based on 222 source repositories. Using \buggymle{}, we make two contributions. First, we construct \bmbl{}, the first difficulty-graded, repository-level variant of MLE-Bench Lite; across 14 evaluated models, the top solve rate drops from 95\% to 9\% as difficulty is scaled from the easiest to the hardest tier. Second, we synthesize a training curriculum across three difficulty levels over 200 additional ML repositories, fine-tuning a Qwen3.5-9B model to rival frontier models on \bmbl{}. Our trained model transfers beyond ML, raising its SWE-bench Lite resolve rate from 24\% to 33\% and its LiveCodeBench v6 pass@1 from 69\% to 71\%.
\end{abstract}

\begin{document}
\maketitle

\section{Introduction}
\label{sec:intro}

\begin{figure}[!t]
\centering
\includegraphics[width=\linewidth]{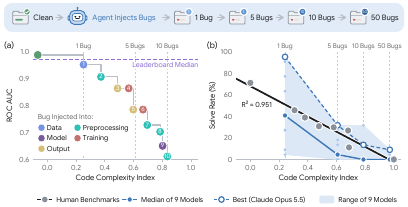}
\caption{\textbf{Code complexity predicts task difficulty across \bmbl{} and eight public coding benchmarks.} \textbf{Top:} An agent injects bugs into the clean repositories of all 22 \bmbl{} tasks to form the 1 Bug, 5 Bugs, 10 Bugs, and 50 Bugs tiers. \textbf{(a)} A representative \bmbl{} repository (Tabular Playground Series) is plotted at the \cci{} of its code after each injected bug. Its ROC AUC falls with each bug, dropping below the Kaggle competition leaderboard median after the first injection, with dots colored by the pipeline stage each bug modifies. \textbf{(b)} On the same \cci{} axis, \textbf{gray points} show eight public coding benchmarks at their \emph{average solve rate across all evaluated models}, along with their linear fit ($R^2 = 0.951$; \cci{} uses no model information). The \textbf{blue band} spans judge-verified Full Repair across nine baseline models from Table~\ref{tab:bmbl_results} (excluding Qwen3.5-9B and its trained variants) on the four \bmbl{} tiers (all 22 tasks per tier, each plotted at its median \cci{}), highlighting the median model (solid) and best model (dashed; Claude Opus 5.5). Solve rates decline with rising \cci{} along the same trajectory as the public benchmarks. We further verify this trend on 200 additional \buggymle{} tasks in Appendix~\ref{app:cci_sweep}.}
\label{fig:headline}
\end{figure}

Autonomous AI systems are increasingly asked to carry out the open-ended work of research and engineering rather than isolated question answering, a direction broadly framed as AI-for-Research \citep{chen2025ai4researchsurveyartificialintelligence_for_research}. One subset of AI-for-Research, the study of AI developing AI training pipelines and models, has emerged as a cornerstone of recursive self-improvement~\citep{chen2026recursive}, as agents that autonomously conduct ML research can improve the foundations on which future research agents are built. Machine learning engineering (MLE) tasks are a natural proving ground for this ambition: here, success requires navigating real codebases, running training pipelines, and iteratively improving measurable metrics. Benchmarks such as MLE-Bench~\citep{chan2025mlebench} and FrontierCS~\citep{mang2025frontiercs} measure how far frontier models have progressed on these open-ended problems, yet they face three recurring limitations. First, they predominantly grade single-file solutions, sidestepping the multi-file code \emph{repositories} that characterize real-world MLE projects. Second, many existing suites are largely saturated, with frontier models consistently outperforming human reference solutions \citep{mang2025frontiercs,yang2026frontis_Openmle}. Third, manual curation is costly: each task demands a vetted dataset, an execution harness capable of running and scoring training, and a human-verified reference outcome. As frontier models advance, hand-built suites saturate within months, driving a surge of interest in synthetically generating MLE tasks at scale \citep{qiang2026mlesmith}.

Synthetic task generation aims to break this curation bottleneck through \textbf{mutation-based} fault injection into existing codebases \citep{shen2026sera, yang2026swe} or \textbf{generation-based} synthesis, where new tasks and their evaluation setups are generated directly from underlying datasets \citep{qiang2026mlesmith, he2026frontiersmith}. While both paradigms have scaled available task \emph{quantity} by an order of magnitude, from hundreds \citep{qiang2026mle} to thousands \citep{qiang2026mlesmith, yang2026frontis_Openmle}, generating large volumes of tasks at flat or unmeasured complexities may leave benchmarks saturated and offer no principled curriculum from easy to hard tasks. Because existing pipelines lack a systematic measure of what makes a task hard, they cannot tell whether newly generated benchmarks remain useful or further scale difficulty as the frontier moves, underscoring the need to both \emph{measure} complexity and \emph{control} task difficulty (Table~\ref{tab:environments}).

Existing MLE agent frameworks compound this limitation by relying on highly specialized, ML-specific code-generation harnesses, such as evolutionary program search \citep{novikov2025alphaevolvecodingagentscientific, yang2026frontis_Openmle} or multi-agent orchestrators \citep{qiang2026matryoshka}, designed to rewrite single standalone scripts. Although these custom scaffolds may attain state-of-the-art benchmark scores via heavy test-time search, they make it difficult to disambiguate gains driven by the harness from the model's intrinsic capabilities. Moreover, regenerating a single script poorly reflects modern MLE practice, where engineers diagnose and modify interconnected multi-file repositories in an open shell environment. Together, these limitations raise three central questions: \emph{what makes a repository-level MLE task challenging for frontier models, how can this be measured, and how can we then systematically control its difficulty?}

To this end, we develop the Code Complexity Index (\cci{}), a principled metric that quantifies the difficulty of a coding benchmark using only structural properties of the code. 
The \cci{} predicts a benchmark's average solve rate from three feature groups (repository scale, dependency complexity, and bug/patch complexity) using a four-parameter logistic regression.
Across eight established coding benchmarks, \cci{} explains 95.1\% of the variance in average solve rate ($R^2 = 0.951$; Figure~\ref{fig:headline}), confirming the intuition that as required solution complexity (\cci{}) increases, solve rates decline.

Inspired by this finding, we present \buggymle{}, a synthetic, mutation-based generation framework for MLE tasks. \buggymle{} uses a large language model (LLM) agent to explore an existing ML task's repository and inject breaking changes, then measures whether frontier LLM solvers can diagnose and fix the injected faults. Unlike prior mutation-based approaches that focus primarily on software engineering tasks, \buggymle{} targets the ML domain specifically, where evaluation requires running training pipelines and measuring task-specific performance metrics. We validate our framework on both synthetically generated repositories from MLE-Smith \citep{qiang2026mlesmith} and human-engineered repositories from PyHealth \citep{wu2026pyhealth20comprehensiveopensource}, demonstrating that our approach can produce realistic, difficulty-controlled MLE tasks regardless of the source repository's origin. Bug-injected tasks follow the same difficulty trajectory as human-curated benchmarks: as we inject more bugs into a repository, its \cci{} rises and all models solve fewer tasks, matching the public-benchmark trend (Figure~\ref{fig:headline}b). 

In summary, we make the following contributions:

\begin{enumerate}
    \item \textbf{Model-Agnostic Complexity Metric (\cci{}):} We develop \cci{}, a ten-feature structural metric that predicts average benchmark solve rates from code properties alone, explaining 95.1\% of the variance across eight public coding benchmarks ($R^2 = 0.951$) without using any features of the evaluated model.

    \item 
    \textbf{Curriculum Learning through Scaling Task Difficulty (\buggymle{}):}
    We introduce \buggymle{}, a scalable environment providing multi-file repository access and an unrestricted shell where bug-injection agents scale task difficulty. At scale, \buggymle{} synthesizes 6{,}674 execution-\allowbreak verified, bug-injected repository variants (${\sim}$260K lines of injected code) from 222 source repositories, predictably lowering solve rates across models while providing dense, verifiable rewards for supervised fine-tuning (SFT) and reinforcement learning (RL).
    \item 
    \textbf{Difficulty-Graded Repository Benchmark (\bmbl{}):}
    We construct \bmbl{}, which, to the best of our knowledge, is the first repository-level variant of MLE-Bench Lite~\citep{chan2025mlebench}: 88 oracle-verified tasks spanning four controlled difficulty tiers (1 to 60 bugs per repository) over 22 multi-file MLE repositories. Across 14 models, the top solve rate falls from 95\% to 9\% from the easiest to the hardest tier.
    
    \item \textbf{Out-of-Distribution (OOD) Validation:} Using \buggymle{} to generate a difficulty-graded curriculum across 200 ML repositories, we train Qwen3.5-9B and carry its gains beyond ML repair, lifting SWE-bench Lite resolve rates from 24\% to 33\% and LiveCodeBench~v6 pass@1 from 69\% to 71\%.
\end{enumerate}

\section{Related Work}
\label{sec:related}

\subsection{Coding and ML Benchmarks}

The rapid advancement of LLM-based coding agents has driven the development of increasingly challenging evaluation benchmarks across several domains. In \textbf{software engineering}, SWE-bench \citep{jimenez2024swe} introduced the paradigm of evaluating agents on real-world GitHub issues, spawning a family of successors including SWE-Gym \citep{pan2024training_swegym} for agent training, SWE-bench Pro \citep{deng2025swebenchproaiagents} for long-horizon enterprise-scale issues, DeepSWE \citep{huang2026deepswe} and FrontierSWE \citep{chu2026frontierswe} for frontier-difficulty tasks, SWE-Marathon \citep{desai2026swe_swe_marathon} for long-horizon software engineering challenges, ProgramBench \citep{yang2026programbench}, and WebGen-Bench \citep{lu2026webgen} for generating entire applications and websites from scratch. In the \textbf{computer science task} space, FrontierCS \citep{mang2025frontiercs} evaluates algorithmic and research-style code generation and FrontierCode \citep{lu2026introducing} probes the limits of LLMs for competitive programming.

For \textbf{machine learning} specifically, MLAgentBench \citep{huang2023mlagentbench} introduced a curated set of ML experimentation tasks, followed by MLGym \citep{nathani2025mlgym} with a gym-style RL interface, MLE-Dojo \citep{qiang2026mle} with over 200 executable ML tasks, and MLS-Bench \citep{lyu2026mls}, which provides additional coverage of ML system building. In the \textbf{debugging} and \textbf{defect detection} space, DSDBench \citep{yang2025stop_dsdbench} targets data science debugging, MLDebugging \citep{huang2025mldebugging} covers multi-library debugging scenarios, and curated defect repositories such as Defect4ML \citep{morovati2023bugs_defect4ML} and DenchMark \citep{kim2021denchmark} provide collections of real-world ML framework bugs.

A common thread across all of these benchmarks is their reliance on \emph{human curation}: each task, test case, and evaluation criterion is manually authored or verified, limiting scalability and making it difficult to systematically vary difficulty. Our work addresses this gap by providing both a scalable generation pipeline and a quantitative model of what makes tasks hard.

\subsection{Environment Scope and Tooling}

Frameworks for ML engineering differ in the scope of code an agent may touch and in the generality of the tools it is given. AlphaEvolve \citep{novikov2025alphaevolvecodingagentscientific} and OpenMLE \citep{yang2026frontis_Openmle} improve on benchmarks by iteratively regenerating candidate solutions under an evolutionary search, in which the model proposes a program and the harness scores it. Matryoshka \citep{qiang2026matryoshka} distributes the work across sub-LLMs that generate and rewrite code, coordinated by an orchestrator with specialized tooling. MLE-Dojo \citep{qiang2026mle}, MLGym \citep{nathani2025mlgym}, and MLAgentBench \citep{huang2023mlagentbench} expose structured task interfaces over which such loops run. In each case the agent acts through a fixed set of operations, and in most cases on a single solution artifact. \buggymle{} instead places the agent in a container with an unrestricted shell, where it may run any command, inspect and patch any file in the repository, and decide for itself whether an experiment is worth its remaining budget (Table~\ref{tab:environments}). 
This freedom mirrors real ML engineering, where practitioners inherit a working repository and change one part of it at a time while leaving the rest of the pipeline intact.
Evolutionary harnesses produce impressive ML performance, but repeatedly regenerating a single code artifact under a scoring loop ignores the complexity of real-world ML workflows, which span multiple files.

\begin{table}[!t]
\caption{Work on agentic MLE falls into three lines: \emph{synthetic data
generators}, which emit new tasks; \emph{training frameworks}, which provide the harness in which an agent is trained or evaluated; and \emph{sandbox environments}, which define what an agent may
touch while solving. A check mark indicates the capability is present. \buggymle{} spans all
three, and is the only work giving the agent an unrestricted shell over a multi-file
repository together with explicit control over task difficulty.}
\label{tab:environments}
\centering
\footnotesize
\setlength{\tabcolsep}{4pt}
\resizebox{\linewidth}{!}{%
\begin{tabular}{lcccccc}
\toprule
 & \multicolumn{2}{c}{\textbf{Line of work}} & \multicolumn{3}{c}{\textbf{Sandbox environment}} & \\
\cmidrule(lr){2-3} \cmidrule(lr){4-6}
\textbf{Work} & \textbf{Synthetic} & \textbf{Training} & \textbf{Multi-file} & \textbf{Unrestricted} & \textbf{Agentic} & \textbf{Difficulty} \\
              & \textbf{data gen.} & \textbf{framework} & \textbf{repository} & \textbf{shell} & \textbf{evaluation} & \textbf{scaling} \\
\midrule
MLAgentBench \citep{huang2023mlagentbench}                     & \xmark & \xmark & \xmark & \xmark & \cmark & \xmark \\
MLGym \citep{nathani2025mlgym}                                 & \xmark & \cmark & \cmark & \xmark & \cmark & \xmark \\
MLE-Dojo \citep{qiang2026mle}                                  & \xmark & \xmark & \xmark & \xmark & \cmark & \xmark \\
MLE-Smith \citep{qiang2026mlesmith}                            & \cmark & \xmark & \xmark & \xmark & \xmark & \xmark \\
OpenMLE \citep{yang2026frontis_Openmle}                        & \cmark & \cmark & \xmark & \xmark & \xmark & \xmark \\
Matryoshka \citep{qiang2026matryoshka}                         & \cmark & \cmark & \xmark & \xmark & \cmark & \xmark \\
\midrule
\buggymle{}                                                    & \cmark & \cmark & \cmark & \cmark & \cmark & \cmark \\
\bottomrule
\end{tabular}}
\end{table}

\subsection{Synthetic Task and Data Generation}

An emerging line of work addresses the curation bottleneck through automated task generation. \textbf{Mutation-based} approaches take existing codebases and systematically introduce faults. SERA \citep{shen2026sera} generates reproducible assessment environments through code mutation, and SWE-smith \citep{yang2026swe} scales bug generation to thousands of Python repositories using automated fault injection. These approaches inherit the realism of their source codebases but typically lack principled control over the resulting task difficulty.

\textbf{Generation-based} approaches synthesize entirely new tasks and evaluation setups. MLE-Smith \citep{qiang2026mlesmith} converts raw datasets into competition-style ML challenges using a multi-agent pipeline, FrontierSmith \citep{he2026frontiersmith} transforms closed-form solution tasks into open-ended problems, TaskCraft \citep{shi2026taskcraft} generates multi-tool agentic tasks through compositional extensions, and AutoCodeBench \citep{zhou2026autocodebench} creates multilingual code problems via LLM-driven reverse synthesis. Complementary to both paradigms, \textbf{self-play} methods such as Self-Questioning Language Models \citep{chen2025self} and Self-Challenging Language Models \citep{zhou2026self} use LLMs to generate challenges for themselves, framing task generation as asymmetric self-play.

\buggymle{} sits in the mutation-based camp but introduces three key differentiators. First, rather than sampling from a fixed fault catalog, our Injector agent explores each repository and introduces breaking changes freely, and we control task difficulty directly by scaling the magnitude of the injected change. Second, \cci{} offers a \emph{quantitative} predictor of benchmark difficulty computed from code structure alone, and is, to our knowledge, the first such metric that relates structural code properties to expected solver performance with high explanatory power ($R^2 = 0.951$). Third, agents solve tasks within a general-purpose terminal environment across the full repository rather than through a fixed action set on a single file, mirroring real-world ML engineering workflows.

\section{Methodology}
\label{sec:method}

\subsection{Code Complexity Index}
\label{sec:cci}

What makes a coding benchmark hard? We answer with \cci{}, a composite score that maps structural properties of a repository and its patch to expected solver performance. We fit our \cci{} across eight established coding benchmarks whose leaderboard results are public.

\paragraph{Raw features.} For each benchmark, we compute the task-level median (p50) of ten structural features and group them into three conceptual dimensions:
\begin{equation}
\begin{aligned}
\mathcal{G}_{\text{repo}} &= \{\text{lines\_of\_code},\; \text{file\_count},\; \text{max\_dir\_depth}\}, \\
\mathcal{G}_{\text{dep}} &= \{\text{dependency\_count},\; \text{max\_dep\_chain}\}, \\
\mathcal{G}_{\text{patch}} &= \{\text{files\_changed},\; \text{hunks},\; \text{lines\_added},\; \text{lines\_removed},\; \text{total\_lines\_delta}\}.
\end{aligned}
\label{eq:groups}
\end{equation}
\emph{Repo scale} ($\mathcal{G}_{\text{repo}}$) captures codebase size and directory structure. \emph{Dependency complexity} ($\mathcal{G}_{\text{dep}}$) measures the depth and connectivity of the inter-file dependency graph, and \emph{bug/patch complexity} ($\mathcal{G}_{\text{patch}}$) describes how far the required fix reaches across files, lines, and hunks. A \emph{hunk} is a contiguous block of modified lines in a unified diff, so a single fix can span several hunks in one or more files.

\begin{figure}[!t]
\centering
\includegraphics[width=\linewidth]{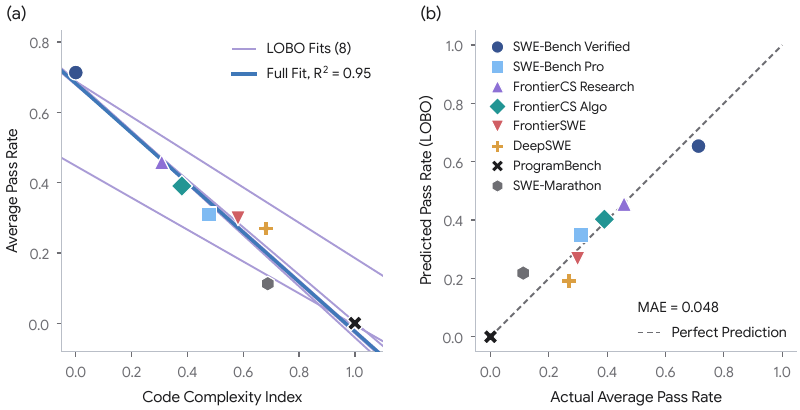}
\caption{\textbf{The \cci{} robustly predicts average solve rates across eight public coding benchmarks.} \textbf{(a)} Linear regressions of average solve rate on \cci{}: both the full fit (blue, $R^2 = 0.951$) and all eight leave-one-benchmark-out refits (lavender) slope downward consistently. \textbf{(b)} Each held-out benchmark's actual pass rate versus the pass rate predicted from the remaining seven benchmarks stays close to the identity line ($\text{MAE} = 0.048$); each benchmark retains the same color and marker across both panels.}
\label{fig:lobo_robust}
\end{figure}
\paragraph{Preprocessing.} All quantities below are computed per benchmark. Let $f_i$ be a benchmark's value of feature $i$. The raw features are right-skewed counts, so we log-transform each one, z-score it across benchmarks, and score each group $\mathcal{G}$ by the mean of its features' z-scores:
\begin{equation}
z_i = \frac{\log(1 + f_i) - \mu_i}{\sigma_i}, \qquad
\bar{z}_{\mathcal{G}} = \frac{1}{|\mathcal{G}|} \sum_{i \in \mathcal{G}} z_i,
\label{eq:zscore}
\end{equation}
where $\mu_i$ and $\sigma_i$ are the mean and standard deviation of $\log(1 + f_i)$ across benchmarks. Each group score is rescaled to unit variance across benchmarks; for benchmark $b$, we write $\bar{z}_{\text{repo},b}$ for the score of $\mathcal{G}_{\text{repo}}$, and likewise for the other two.

\paragraph{Modeling benchmark difficulty.} We regress model $m$'s continuous solve rate $\pi_{m,b} \in [0, 1]$ on benchmark $b$ onto its three group scores via logistic regression:
\begin{equation}
\text{logit}(\pi_{m,b}) = \beta_0 + \beta_{\text{repo}} \, \bar{z}_{\text{repo},b} + \beta_{\text{dep}} \, \bar{z}_{\text{dep},b} + \beta_{\text{patch}} \, \bar{z}_{\text{patch},b},
\label{eq:glm}
\end{equation}
where $\text{logit}(\pi) = \log\frac{\pi}{1-\pi}$, which keeps predicted solve rates between 0 and 1. Each (model, benchmark) pair is weighted by the number of tasks in the benchmark. Pairs from the same benchmark are correlated, so we cluster standard errors by benchmark.

\paragraph{\cci{} derivation.} A benchmark is harder when its features lower the predicted solve rate, so its raw complexity score is the weighted sum of its group scores with the sign flipped:
\begin{equation}
\text{CCI}_b = -\left(\beta_{\text{repo}} \, \bar{z}_{\text{repo},b} + \beta_{\text{dep}} \, \bar{z}_{\text{dep},b} + \beta_{\text{patch}} \, \bar{z}_{\text{patch},b}\right).
\label{eq:cci_raw}
\end{equation}
Finally, we min-max scale the raw scores across the eight reference benchmarks:
\begin{equation}
\cci_b = \frac{\text{CCI}_b - \min_{b'} \text{CCI}_{b'}}{\max_{b'} \text{CCI}_{b'} - \min_{b'} \text{CCI}_{b'}}.
\label{eq:cci_scaled}
\end{equation}
New repositories and tiers (including \bmbl{}) are projected onto this scale using the fixed $(\mu_i, \sigma_i)$, $\beta$, and min--max bounds from the eight reference benchmarks without refitting.

This four-parameter specification survives every ablation we tried: alternative preprocessing transformations, alternative link functions, and leave-one-benchmark-out (LOBO) cross-validation (Figure~\ref{fig:lobo_robust}). It also beats both single-group subsets and richer parameterizations among specifications that predict valid solve rates (Appendix~\ref{app:validation}). Adding model identity, whether as fixed effects or as an external intelligence index, makes out-of-sample predictions \emph{worse}, because some benchmarks are completely saturated and others not at all (Appendix~\ref{app:intelligence}). 

\begin{remark}[Patch features dominate benchmark difficulty]
Bug/patch complexity is the primary driver of benchmark difficulty. Its coefficient is the largest in magnitude of the three, ahead of repo scale and about three times that of dependency complexity in \cci{} (Appendices~\ref{app:signstab} and~\ref{app:groups}; Appendix Figure~\ref{fig:lobo_coefs}). Intuitively, even a large repository remains tractable when a fix is localized, whereas larger patches require diagnosing and coordinating edits across multiple files and hunks.
\end{remark}

\begin{remark}[\buggymle{} leverages \cci{} to control task difficulty] All models follow the \cci{} trend (Figure~\ref{fig:headline}b). Regardless of model scale, solve rates drop predictably as we increase \cci{} with our \buggymle{} framework, not only validating our \cci{} model but also enabling the construction of tasks at any target difficulty.
\end{remark}

\subsection{\buggymle{}}
\label{sec:pipeline}

\begin{figure}[!t]
\centering
\includegraphics[width=\linewidth]{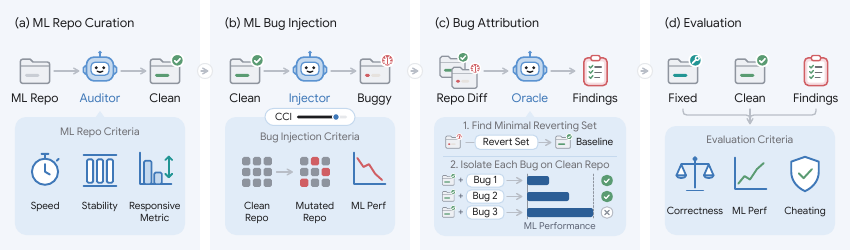}
\caption{\textbf{\buggymle{} data curation workflow: every multi-bug repository is built from a vetted clean repository and decomposed into verified, individually impactful bugs.} \textbf{(a) Repository curation:} an auditor keeps candidate ML repositories that run fast, give a stable metric across runs, and have a responsive metric (one that moves when the code changes). \textbf{(b) Bug injection:} an injector agent turns the clean repository into a mutated one at a target complexity (the \cci{} slider); a mutation is kept only if automated checks confirm it changes ML performance. \textbf{(c) Bug attribution:} in two passes, an oracle first cumulatively reverts mutations until baseline performance recovers (identifying a minimal reverting set), then applies each mutation alone onto the clean repository to measure isolated ML performance against the clean baseline (dashed), confirming impactful bugs and marking no-effect edits as decoys. \textbf{(d) Evaluation:} a solver's fixed repository is graded against the clean repository and the oracle's findings for correctness, recovered ML performance, and cheating (anti-bypass checks and an oracle-grounded LLM judge).}
\label{fig:overview}
\end{figure}

\buggymle{} is a four-stage, configuration-driven pipeline that turns clean machine learning repositories into graded debugging tasks and evaluates agent repairs (Figure~\ref{fig:overview}). YAML configurations set task selection, mutation budgets, model backends, and grading criteria for each stage, and stages run in parallel for large-scale generation:
\begin{enumerate}
    \item \textbf{Injector:} An LLM agent explores a clean ML repository inside a Docker sandbox and introduces breaking changes across the pipeline, subject to automated verification checks.
    \item \textbf{Oracle:} Starting from the diff between the clean and buggy repositories, an ablation procedure isolates which mutations actually drive performance degradation.
    \item \textbf{Solver:} A debugging agent receives the buggy codebase and tries to diagnose and repair the bugs without the clean reference code or any bug specification.
    \item \textbf{Evaluator:} The repaired candidate solution is executed in a clean container and graded on quantitative performance recovery and on an oracle-guided code review.
\end{enumerate}

\paragraph{Bug injection and oracle attribution.}\label{sec:injection} The Injector works without a fixed bug taxonomy. It inspects the target repository and introduces plausible bugs wherever they fit, in data loading, preprocessing, model architecture, training, or evaluation code. Difficulty comes from the mutation budget (the required number of modified files, hunks, and lines), so task complexity can be raised while the dataset and codebase stay fixed (Appendix~\ref{app:cci_sweep}). 

Deterministic scripts in the Docker sandbox validate every candidate injection against three criteria. (a) The diff against the clean repository must meet the mutation budget across files, hunks, and lines. (b) End-to-end execution must either crash or degrade the evaluation metric by more than three standard deviations of the clean repository's run-to-run variance, or by at least 10\% when that variance is unmeasured. (c) If execution completes, it must still produce a valid metric. A failed injection goes back to the agent with diagnostic feedback, up to five times.

A multi-hunk injection can mix benign edits with true bugs. The \textbf{Oracle} separates them in two passes (Algorithm~\ref{alg:oracle}, Appendix~\ref{app:oracle}). It first reverts candidate mutations in ranked order until baseline performance returns, which identifies a minimal \emph{reverting set}. It then applies each mutation alone to the clean repository and measures its isolated effect on the metric. The result is a verified checklist of \emph{impactful mutations}, kept apart from inert \emph{decoy mutations}, for downstream grading.

\paragraph{Agent harness.}\label{sec:harness} Injector, Oracle, and Solver share one execution harness inside a GPU-enabled Docker container. Agents interact through a multi-turn tool-calling loop with unrestricted \texttt{bash} and file-editing tools (\texttt{read\_file}, \texttt{write\_file}, \texttt{modify\_file}, and \texttt{list\_dir}), enabling arbitrary multi-file edits and script execution within a fixed turn budget (Figure~\ref{fig:overview}; Appendix~\ref{app:harness}).

\paragraph{Evaluation framework.} \label{sec:evaluation} Candidate repairs are graded by execution-based checks and oracle-\allowbreak grounded semantic verification. The \textbf{execution check} runs the solver's modified repository in a fresh container and records the output metric (scoring 0 if execution fails). The \textbf{metric check} then measures the \emph{recovery ratio}:
\begin{equation}
\text{recovery} = \frac{m_{\text{fixed}}}{m_{\text{baseline}}}\quad\text{(or } m_{\text{baseline}}/m_{\text{fixed}} \text{ for lower-is-better metrics)},
\label{eq:recovery}
\end{equation}
where $m_{\text{baseline}}$ and $m_{\text{fixed}}$ denote the performance of the clean and repaired repositories. A candidate achieves \emph{full repair} only if it satisfies three conditions:
\begin{enumerate}
    \item \textbf{Metric recovery:} $\text{recovery} \geq 0.99$.
    \item \textbf{Routing integrity:} The repository's entry point still executes the previously mutated modules, so an agent cannot bypass broken components.
    \item \textbf{Per-bug verification:} Grounded in the Oracle's verified bug records (ignoring decoys), an LLM judge inspects each bug site against its ground-truth diff and mechanism, confirming it was resolved by a genuine fix or refactor rather than a symptom-masking workaround. \bmbl{} grading requires every verified bug to pass, whereas RL rewards use the fraction of resolved sites as a dense training signal (Section~\ref{sec:training}; judge configurations in Appendix~\ref{app:reviewer}).
\end{enumerate}
Beyond binary solve rates, we report \emph{progress} $p = \mathrm{clip}\big((m_{\text{fixed}}-m_{\text{buggy}})/(m_{\text{baseline}}-m_{\text{buggy}}),\,0,\,1\big)$, the fraction of the degradation recovered, and the fraction of tasks where $m_{\text{fixed}}$ strictly improves over $m_{\text{buggy}}$. For \bmbl{} tasks we also map $m_{\text{fixed}}$ to its Kaggle leaderboard percentile.

\subsection{Curriculum Learning with \buggymle{}}
\label{sec:training}

\begin{figure}[!t]
\centering
\includegraphics[width=\linewidth]{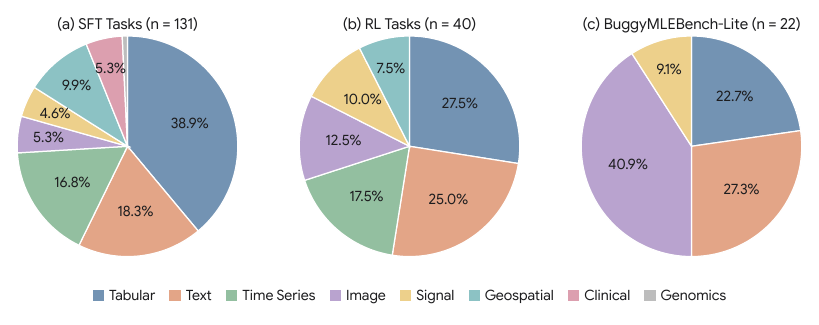}
\caption{\textbf{Distribution of tasks across data modalities.} Proportion of tasks by modality for \textbf{(a)} SFT + RFT training, \textbf{(b)} RL training, and \textbf{(c)} the \bmbl{} evaluation benchmark, where image tasks form the largest share. Detailed recipe statistics are provided in Appendix Figure~\ref{fig:training_recipe}.}
\label{fig:training_stats}
\end{figure}

Because \buggymle{} can synthesize tasks at any target difficulty by scaling the injected hunk budget on the same underlying codebase, we can construct a difficulty-graded curriculum while holding the repository distribution and modalities fixed across 200 training repositories (Figure~\ref{fig:training_stats}). Using this capability, our \textbf{All Difficulties} recipe trains Qwen3.5-9B~\citep{qwen3.5} with low-rank adaptation (LoRA) adapters in two sequential phases across the full complexity spectrum: SFT followed by RL (Appendix~\ref{app:rl_recipe}). The SFT phase grounds the model in the sandbox harness using 2{,}724 evaluator-verified repair trajectories across all difficulty bands from five frontier teacher models, including Gemini~3.6 Flash and Gemini~3.7 Flash~\citep{deepmind2026gemini36flash, deepmind2026gemini37flash}, supplemented via rejection-sampling fine-tuning (RFT) on 411 verified rollouts from a teacher-tuned Qwen3.8-27B~\citep{qwen38} model to incorporate explicit reasoning traces (Appendix~\ref{app:splits}). In practice, because issuing separate per-bug judge calls across thousands of multi-bug rollouts is prohibitively expensive when curating training data at scale, we filter candidate SFT trajectories (as well as our 200-task scaling sweep; Appendix~\ref{app:cci_sweep}) with a single repository-level LLM review conditioned on the full Oracle report (Appendix~\ref{app:reviewer}), rejecting patches that edit only decoys, bypass faulty logic, or introduce new defects.

Initialized from this SFT checkpoint, we apply Group Relative Policy Optimization (GRPO) \citep{shao2024deepseekmathpushinglimitsmathematical_grpo} across a three-stage curriculum of increasing difficulty on a fixed pool of 40 repositories: Easy (1--5 hunks), Medium (10--12), and Hard (30--34). Each rollout runs as a single interactive episode without test-time search and is scored using all evaluation criteria from Section~\ref{sec:evaluation}: continuous metric progress, execution and routing-integrity gates, and per-bug resolution judged against the injected defects (formalized in Appendix~\ref{app:rl_recipe}). Incorporating every criterion into the reward prevents superficial workarounds and enables the policy to keep climbing on residual or silent bugs even when raw ML performance practically saturates.

\subsection{\bmbl{}}
\label{sec:benchmark}

\begin{figure}[!t]
\centering
\includegraphics[width=\linewidth]{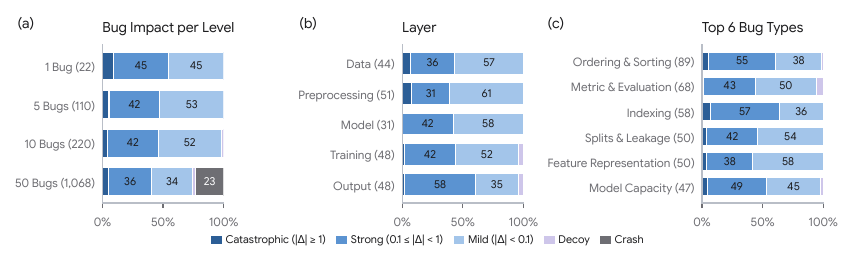}
\caption{\textbf{Distribution of injected bug impacts across complexity tiers, pipeline stages, and defect categories.} Proportion of bugs stratified by their isolated impact on the competition test metric. \textbf{(a)} Breakdown across complexity tiers, showing that mild and strong impacts dominate the 1, 5, and 10 Bugs tiers, whereas catastrophic failures are rare and execution crashes occur only at the 50 Bugs tier. \textbf{(b)} Distribution of unique bugs across the 1, 5, and 10 Bugs tiers by ML pipeline stage, and \textbf{(c)} across the six most frequent defect categories (total counts in parentheses; individual bugs may span multiple categories). A bug is classified as \emph{impactful} if applying it in isolation shifts the graded test score beyond the task's significance threshold, and is categorized by magnitude as \emph{mild} ($|\Delta| < 0.1$), \emph{strong} ($0.1 \le |\Delta| < 1$), or \emph{catastrophic} ($|\Delta| \ge 1$). A \emph{crash} fails to produce a valid submission file, whereas a \emph{decoy} alters at most the validation score.}
\label{fig:bug_outcomes_main}
\end{figure}

Because MLE-Bench Lite \citep{chan2025mlebench} provides only a raw dataset and grading script for each of its 22 competitions rather than an existing codebase, we first construct and verify a modular baseline repository for every competition prior to bug injection. Using \buggymle{}, we instantiate four complexity tiers across these repositories. The \textbf{1 Bug}, \textbf{5 Bugs}, and \textbf{10 Bugs} tiers contain the corresponding number of verified defects per repository, whereas the \textbf{50 Bugs} tier first expands each repository to roughly twice its original size before injecting 23--60 bugs (Appendix~\ref{app:splits}). For the first three tiers, oracle verification confirms that reverting all injected mutations exactly restores the baseline grading score on every task. To prevent data contamination, no evaluation datasets or repositories appear in the training corpus.

\begin{figure}[!t]
\centering
\includegraphics[width=\linewidth]{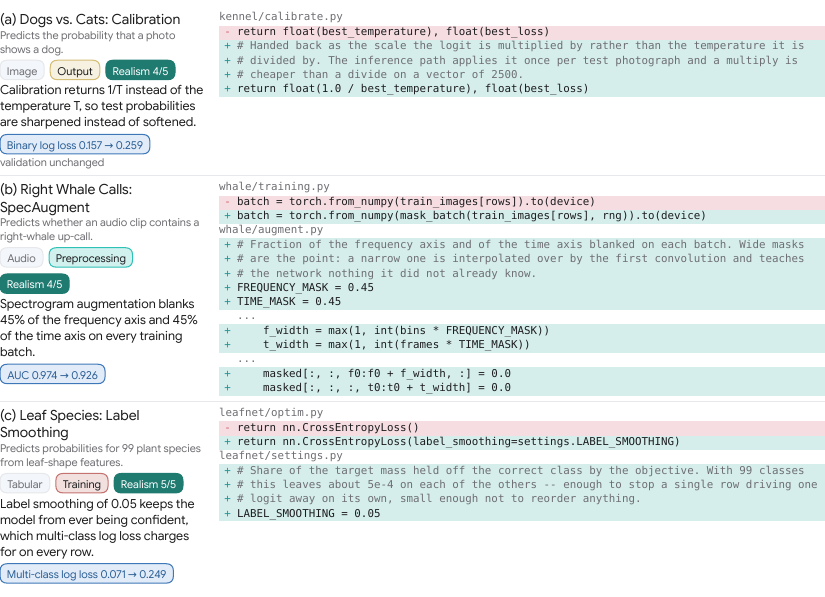}
\caption{\textbf{Representative examples of injected benchmark bugs.} \textbf{(a)--(c)} Three injected defects from \bmbl{} spanning image, audio, and tabular tasks, showing removed lines (red) and added lines (green), including the inline comments generated during injection to rationalize each modification. Reported scores reflect each defect's isolated impact on the competition test metric, alongside its 1--5 realism rating assigned by the LLM judge (RQ6, Section~\ref{sec:hard_bugs}).}
\label{fig:bug_examples}
\end{figure}

Empirical validation confirms that nearly all injected bugs significantly degrade test performance and span every stage of the ML pipeline (Figure~\ref{fig:bug_outcomes_main}). Furthermore, the vast majority of defects are rated as realistic by automated and human judges, with very few flagged as implausible (Appendix Figure~\ref{fig:bmbl_realism}). In practice, many injected defects resemble routine engineering modifications accompanied by plausible justifying comments (Figure~\ref{fig:bug_examples}); Appendix~\ref{app:bug_examples} presents nine additional examples spanning the full spectrum from overt self-sabotage to subtle, realistic modeling flaws.

\section{Experiments}
\label{sec:experiments}
\label{sec:setup}

\paragraph{Trained models.} To isolate the effect of curriculum difficulty, we compare our \textbf{All Difficulties} Qwen3.5-9B agent (Section~\ref{sec:training}) against three controlled ablations trained under the same LoRA setup: \textbf{Easy Only} (1--5 hunks; 936 teacher SFT trajectories) and \textbf{Hard Only} (30--34 hunks in RL and $\geq 30$ in SFT; 628 teacher SFT trajectories), each of which restricts both SFT and GRPO to its respective single difficulty subset, and \textbf{SFT Only} (the SFT + RFT checkpoint of All Difficulties prior to RL). From each RL run, we evaluate the exported checkpoint achieving the highest held-out validation score on its target difficulty band (Appendix~\ref{app:rl_curves}); corpus breakdowns and hyperparameters appear in Appendix Figure~\ref{fig:training_recipe} and Appendix~\ref{app:rl_recipe}.

\paragraph{Baselines.} External baselines fall into two tiers by model scale and inference cost (Appendix~\ref{app:eval_setup}). \textbf{Budget Models} are each provider's lowest-cost commercial tier or open-weight models of comparable size: Claude Haiku~5.5, Claude Haiku~4.5, Gemini~3.5 Flash-Lite, Qwen3.8-27B without thinking, and our untuned Qwen3.5-9B base model~\citep{anthropic2025claudehaiku45, anthropic2026claudehaiku55, deepmind2026gemini35flashlite, qwen3.5, qwen38}. They are the primary comparison class for our 9B agent. \textbf{Frontier Models} are larger flagship systems (Claude Opus~5.5, Claude Sonnet~5.5, Claude Sonnet~5, Claude Sonnet~4.6, and Gemini~3.8 Flash)~\citep{anthropic2026claudeopus55, anthropic2026claudesonnet46, anthropic2026claudesonnet5, anthropic2026claudesonnet55, deepmind2026gemini38flash} and serve as upper-bound reference points.

\paragraph{Evaluation.} All models, trained and hosted, run in the same sandbox with the same task prompt (Appendix~\ref{app:harness}) on the 22 tasks of each \bmbl{} tier. We report three metrics. \emph{Full Repair} (solve rate) is the percentage of tasks that reach full repair (Section~\ref{sec:evaluation}). \emph{Recovery} is the mean recovery ratio clipped to $[0, 1]$. \emph{Percentile} is the mean Kaggle leaderboard percentile of the repaired code's submission, scored 0 when no valid submission is produced.

\paragraph{OOD evaluation.} Transfer beyond ML debugging is measured on two general coding benchmarks that share nothing with our training corpus. SWE-bench Lite \citep{jimenez2024swe} poses 300 repository issues, scored by pass@1. LiveCodeBench~v6 \citep{jain2025livecodebench} poses 1{,}055 competitive programming problems, answered in a single turn without tools, with two samples per problem. To probe transfer across a broader task spectrum, Appendix~\ref{app:ood_full} evaluates four additional settings: three that synthesize ML solutions from scratch (as a standalone script without tools, a single file in the sandbox, or a full repository) and DeepSWE~\citep{huang2026deepswe}, which benchmarks longer-horizon software issues in general Python repositories.

\section{Results}
\label{sec:results}

We organize our empirical evaluation around seven research questions examining how agent performance scales with benchmark difficulty (\textbf{RQ1}), how training curriculum difficulty shapes agent capability (\textbf{RQ2}, \textbf{RQ3}), the distinct contributions of SFT and RL (\textbf{RQ4}), OOD generalization (\textbf{RQ5}), and how bug realism and observability govern task difficulty (\textbf{RQ6}, \textbf{RQ7}).

\begin{table}[!t]
\centering
\caption{\textbf{Evaluation results on the held-out \bmbl{} benchmark} (22 tasks per difficulty tier; all values in percentages). Reference denotes the clean baseline repository prior to bug injection. The 50 Bugs tier expands each repository to roughly twice its original size with 23--60 injected bugs per task. Arrows indicate the direction of better performance. Shading, \textbf{bold} (best), and \underline{underline} (second best) highlight rankings within the budget and 9B model comparison class. Model tiers and metrics are defined in Section~\ref{sec:setup}.}
\label{tab:bmbl_results}
\resizebox{\linewidth}{!}{
\setlength{\tabcolsep}{3pt}
\begin{tabular}{@{}l | ccc | ccc | ccc | ccc@{}}
\toprule[1.5pt]
 & \multicolumn{3}{c|}{\textbf{1 Bug}} & \multicolumn{3}{c|}{\textbf{5 Bugs}} & \multicolumn{3}{c|}{\textbf{10 Bugs}} & \multicolumn{3}{c}{\textbf{50 Bugs}} \\
\cmidrule(lr){2-4} \cmidrule(lr){5-7} \cmidrule(lr){8-10} \cmidrule(lr){11-13}
\textbf{Model} & Full Repair \hib & Recovery \hib & Percentile \hib & Full Repair \hib & Recovery \hib & Percentile \hib & Full Repair \hib & Recovery \hib & Percentile \hib & Full Repair \hib & Recovery \hib & Percentile \hib \\
\midrule\midrule
\rowcolor{sectiongray}\multicolumn{13}{@{}l}{\textit{Reference}} \\
Clean repository (no bugs) & -- & -- & \textit{49.1} & -- & -- & \textit{49.1} & -- & -- & \textit{49.1} & -- & -- & \textit{52.0} \\
\rowcolor{sectiongray}\multicolumn{13}{@{}l}{\textit{Frontier Models}} \\
Claude Opus 5.5 & 95.5 & 99.8 & 57.2 & 31.8 & 88.4 & 44.6 & 13.6 & 92.2 & 46.7 & 9.1 & 77.8 & 43.9 \\
Claude Sonnet 5.5 & 90.9 & 95.5 & 54.0 & 27.3 & 85.5 & 42.4 & 31.8 & 86.4 & 43.6 & 0.0 & 64.3 & 30.6 \\
Claude Sonnet 5 & 63.6 & 80.2 & 39.2 & 4.5 & 67.5 & 24.2 & 0.0 & 53.7 & 12.6 & 0.0 & 30.2 & 3.5 \\
Claude Sonnet 4.6 & 27.3 & 30.9 & 17.5 & 0.0 & 25.6 & 5.8 & 0.0 & 30.1 & 4.5 & 0.0 & 0.0 & 0.0 \\
Gemini 3.8 Flash & 27.3 & 27.3 & 15.6 & 13.6 & 41.2 & 21.9 & 13.6 & 18.5 & 14.1 & 4.5 & 22.4 & 9.2 \\
\rowcolor{sectiongray}\multicolumn{13}{@{}l}{\textit{Budget Models}} \\
Claude Haiku 5.5 & \cellcolor{royalblue!40}\textbf{77.3} & \cellcolor{royalblue!40}\textbf{89.5} & \cellcolor{royalblue!40}\textbf{46.1} & \cellcolor{royalblue!33}\underline{18.2} & \cellcolor{royalblue!39}\underline{75.2} & \cellcolor{royalblue!36}\underline{33.4} & \cellcolor{royalblue!6}0.0 & \cellcolor{royalblue!35}\underline{71.1} & \cellcolor{royalblue!21}\underline{19.3} & \cellcolor{royalblue!6}0.0 & \cellcolor{royalblue!40}\textbf{41.3} & \cellcolor{royalblue!40}\textbf{8.8} \\
Claude Haiku 4.5 & \cellcolor{royalblue!8}9.1 & \cellcolor{royalblue!16}33.3 & \cellcolor{royalblue!12}11.2 & \cellcolor{royalblue!6}0.0 & \cellcolor{royalblue!11}17.3 & \cellcolor{royalblue!7}4.1 & \cellcolor{royalblue!6}0.0 & \cellcolor{royalblue!11}18.7 & \cellcolor{royalblue!7}2.6 & \cellcolor{royalblue!6}0.0 & \cellcolor{royalblue!7}5.6 & \cellcolor{royalblue!26}\underline{5.5} \\
Gemini 3.5 Flash-Lite & \cellcolor{royalblue!6}4.5 & \cellcolor{royalblue!6}8.2 & \cellcolor{royalblue!6}3.8 & \cellcolor{royalblue!6}0.0 & \cellcolor{royalblue!6}8.0 & \cellcolor{royalblue!6}2.8 & \cellcolor{royalblue!6}0.0 & \cellcolor{royalblue!6}6.9 & \cellcolor{royalblue!6}1.3 & \cellcolor{royalblue!6}0.0 & \cellcolor{royalblue!9}7.0 & \cellcolor{royalblue!25}5.2 \\
Qwen3.8-27B & \cellcolor{royalblue!23}40.9 & \cellcolor{royalblue!33}72.7 & \cellcolor{royalblue!31}34.9 & \cellcolor{royalblue!13}4.5 & \cellcolor{royalblue!24}44.0 & \cellcolor{royalblue!18}14.8 & \cellcolor{royalblue!6}0.0 & \cellcolor{royalblue!27}53.1 & \cellcolor{royalblue!14}11.0 & \cellcolor{royalblue!6}0.0 & \cellcolor{royalblue!11}9.8 & \cellcolor{royalblue!11}1.6 \\
Qwen3.5-9B & \cellcolor{royalblue!10}13.6 & \cellcolor{royalblue!29}62.1 & \cellcolor{royalblue!23}24.7 & \cellcolor{royalblue!6}0.0 & \cellcolor{royalblue!26}49.3 & \cellcolor{royalblue!18}14.5 & \cellcolor{royalblue!6}0.0 & \cellcolor{royalblue!25}48.8 & \cellcolor{royalblue!12}8.1 & \cellcolor{royalblue!6}0.0 & \cellcolor{royalblue!6}4.1 & \cellcolor{royalblue!6}0.5 \\
\rowcolor{sectiongray}\multicolumn{13}{@{}l}{\textit{Ours: Curriculum Learning with \buggymle{} (9B)}} \\
SFT Only (9B) & \cellcolor{royalblue!27}50.0 & \cellcolor{royalblue!32}70.7 & \cellcolor{royalblue!33}36.9 & \cellcolor{royalblue!13}4.5 & \cellcolor{royalblue!33}62.7 & \cellcolor{royalblue!25}22.0 & \cellcolor{royalblue!6}0.0 & \cellcolor{royalblue!23}43.4 & \cellcolor{royalblue!13}9.0 & \cellcolor{royalblue!6}0.0 & \cellcolor{royalblue!9}7.2 & \cellcolor{royalblue!10}1.5 \\
\multicolumn{13}{@{}l}{\quad\textit{SFT+RL Curriculum:}} \\
\qquad Easy Only (1--5 Hunks) & \cellcolor{royalblue!27}50.0 & \cellcolor{royalblue!33}73.9 & \cellcolor{royalblue!32}35.6 & \cellcolor{royalblue!6}0.0 & \cellcolor{royalblue!29}54.6 & \cellcolor{royalblue!15}12.2 & \cellcolor{royalblue!6}0.0 & \cellcolor{royalblue!25}48.0 & \cellcolor{royalblue!12}8.2 & \cellcolor{royalblue!6}0.0 & \cellcolor{royalblue!11}9.9 & \cellcolor{royalblue!26}\underline{5.5} \\
\qquad Hard Only (30--34 Hunks) & \cellcolor{royalblue!14}22.7 & \cellcolor{royalblue!32}69.4 & \cellcolor{royalblue!30}34.2 & \cellcolor{royalblue!33}\underline{18.2} & \cellcolor{royalblue!35}67.3 & \cellcolor{royalblue!34}31.0 & \cellcolor{royalblue!23}\underline{9.1} & \cellcolor{royalblue!30}59.6 & \cellcolor{royalblue!21}19.1 & \cellcolor{royalblue!6}0.0 & \cellcolor{royalblue!13}12.3 & \cellcolor{royalblue!10}1.4 \\
\qquad All Difficulties & \cellcolor{royalblue!38}\underline{72.7} & \cellcolor{royalblue!37}\underline{81.5} & \cellcolor{royalblue!38}\underline{43.0} & \cellcolor{royalblue!40}\textbf{22.7} & \cellcolor{royalblue!40}\textbf{77.4} & \cellcolor{royalblue!40}\textbf{37.0} & \cellcolor{royalblue!40}\textbf{18.2} & \cellcolor{royalblue!40}\textbf{81.1} & \cellcolor{royalblue!40}\textbf{40.8} & \cellcolor{royalblue!6}0.0 & \cellcolor{royalblue!15}\underline{14.1} & \cellcolor{royalblue!25}5.1 \\
\bottomrule[1.5pt]
\end{tabular}
}
\end{table}

\begin{figure}[!t]
\centering
\includegraphics[width=\linewidth]{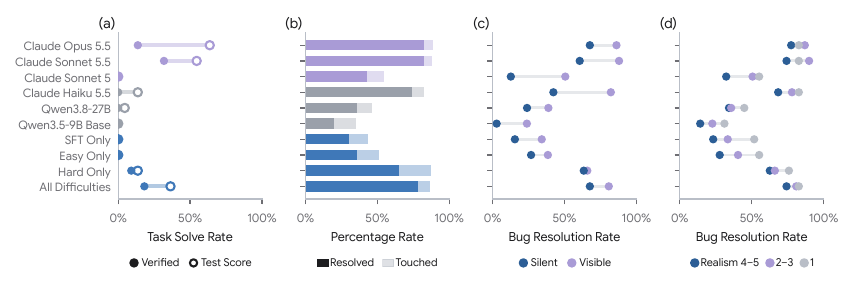}
\caption{\textbf{Impact of metric recovery, fault localization, observability, and realism on bug resolution.} Each row corresponds to one of the ten models audited at the per-bug level, ordered as in Table~\ref{tab:bmbl_results}, showing frontier models (lavender), budget baselines (gray), and our trained 9B models (blue). \textbf{(a)} 10 Bugs task solve rate under test-score recovery alone (open circles) versus judge-verified full repair requiring all injected bugs to be fixed (filled circles); the gap indicates tasks where the test metric recovered despite remaining bugs, which is widest for Claude Opus~5.5 and Sonnet~5.5. \textbf{(b)} Proportion of injected bugs modified (light) versus fully resolved (dark), showing that top agents (All Difficulties, Opus~5.5, and Sonnet~5.5) localize nearly all bugs and resolve most modified sites, whereas the 9B base model modifies roughly one-third. \textbf{(c)} Resolution rate for \emph{visible} bugs (which degrade validation metrics) versus \emph{silent} bugs (which affect only test predictions), demonstrating consistently lower resolution on silent bugs across all models. \textbf{(d)} Resolution rate stratified by LLM-judged realism score (4--5, 2--3, and 1). Panels (b)--(d) pool tasks across the 1, 5, and 10 Bugs tiers; all x-axes report percentages.}
\label{fig:repair_depth}
\end{figure}

\paragraph{RQ1: How do agents fare as \bmbl{} gets harder?} Consistent with \cci{} predictions, solve rates decline sharply across difficulty tiers (monotonically for 13 of the 14 models), with the top solve rate dropping from 95\% on the easiest tier to 9\% on the hardest (Table~\ref{tab:bmbl_results}). Model separation is clearest on the 10 Bugs tier, where our 9B All Difficulties agent outperforms Claude Opus~5.5 in end-to-end task solves and trails only Claude Sonnet~5.5. This ordering stems from strict per-bug verification: although Opus~5.5 recovers the target test metric on most tasks, it frequently leaves individual injected bugs unresolved, resulting in a per-bug resolution rate comparable to our 9B agent (Figure~\ref{fig:repair_depth}). At the 50 Bugs tier, end-to-end task resolution nearly vanishes, matching the trajectory predicted by human benchmarks at this \cci{} level; only Claude Opus~5.5 and Gemini~3.8 Flash complete any task in full. While our agent still resolves the second-highest share of individual bugs behind Sonnet~5.5, residual unfixed bugs suppress its overall metric recovery relative to Opus~5.5, as every 50 Bugs episode of our agent terminates at the turn or context limit (often before producing a valid submission).

\paragraph{RQ2: Do harder training tasks improve performance on easier tasks?} Incorporating high-\allowbreak complexity tasks improves single-bug repair, provided those tasks are embedded within a full curriculum. Specifically, adding high-complexity RL stages raises the 1 Bug solve rate of the All Difficulties agent substantially above both the SFT Only and Easy Only baselines (72.7\% vs.\ 50.0\%). Conversely, training exclusively on hard tasks hurts performance on the 1 Bug tier: because of distribution shift, the Hard Only model achieves fewer end-to-end 1 Bug solves (22.7\% vs.\ 50.0\%) despite resolving more individual bugs (59.1\% vs.\ 50.0\%) than the SFT baseline.

\begin{figure}[!t]
\centering
\includegraphics[width=\linewidth]{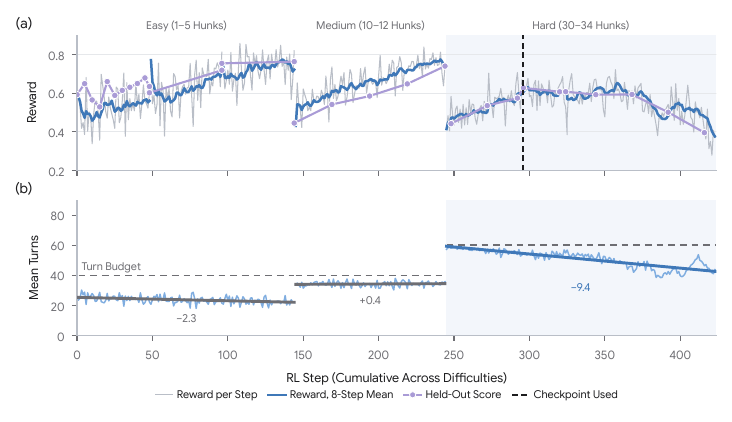}
\caption{\textbf{High-complexity curriculum stages drive turn efficiency.} Training dynamics of the All Difficulties curriculum across Easy (1--5 hunks), Medium (10--12), and Hard (30--34) RL stages (Hard stage shaded). \textbf{(a)} Per-step training reward (gray), 8-step moving average (blue), held-out evaluation score on the corresponding difficulty band (lavender), and the selected evaluation checkpoint (dashed). Reward recovers within each stage and remains stable through the Hard stage. \textbf{(b)} Mean turns per episode with stage-wise linear trends relative to the turn budget (dashed). Episode length remains nearly flat during the Easy and Medium stages but drops sharply during the Hard stage without degrading reward. An earlier turn-budget adjustment during the Hard stage is analyzed in Appendix Figure~\ref{fig:hard_budget}.}
\label{fig:rl_reward_turns}
\end{figure}

\paragraph{RQ3: Is there any benefit to training on easier tasks?} Although training solely on easy tasks fails to generalize upward (with Easy Only solving 0\% of tasks on the 5 and 10 Bugs tiers), easy stages serve as an essential foundation for curriculum learning. Including easy stages allows the All Difficulties model to consistently outperform Hard Only on complex tiers, doubling the solve rate on 10 Bugs. Furthermore, easy stages initiate a gradual reduction in episode length. However, the sharpest efficiency gains emerge during the budget-constrained hard stage, where the agent learns to complete repairs in substantially fewer turns without sacrificing reward (Figure~\ref{fig:rl_reward_turns}).

\paragraph{RQ4: What do SFT and RL each contribute?} SFT primarily establishes reliable tool use, eliminating syntax errors in tool calls and replacing the base model's full-file rewrites with targeted edits. Subsequent RL training improves fault localization and exploration efficiency. Across the 1, 5, and 10 Bugs tiers, the All Difficulties agent modifies the majority of bug-containing files (compared to fewer than half for SFT Only), executes substantially fewer exploratory throwaway scripts, and achieves higher mean metric recovery at every tier. Pipeline re-execution frequency does not distinguish the models, as the base, SFT Only, and All Difficulties agents all re-run the pipeline following their final edit in most episodes. Appendix Figure~\ref{fig:trajectory} illustrates this qualitative shift on a representative 5 Bugs task: whereas the base model repeatedly trains auxiliary models and leaves two bug sites untouched, the RL agent reverts four bugs simultaneously, diagnoses a fifth from a chance-level validation AUC, and resolves it.

\begin{table}[!t]
\centering
\small
\caption{\textbf{OOD generalization on general coding benchmarks} (experimental settings in Section~\ref{sec:experiments}). Arrows indicate the direction of better performance, with shading, \textbf{bold}, and \underline{underline} ranked across all rows. $\Delta$ denotes the paired difference relative to the untuned base model, where $\dagger$ indicates a 95\% bootstrap confidence interval excluding zero.}
\label{tab:regime_swe}
\resizebox{\linewidth}{!}{
\setlength{\tabcolsep}{4pt}
\begin{tabular}{@{}l | ccccc | ccccc@{}}
\toprule[1.5pt]
 & \multicolumn{5}{c|}{\textbf{SWE-bench Lite}} & \multicolumn{5}{c}{\textbf{LiveCodeBench v6}} \\
\cmidrule(lr){2-6} \cmidrule(lr){7-11}
\textbf{Model} & \makecell[c]{Resolved (\%)\\\hib} & \makecell[c]{Fail-to-Pass\\Tests (\%)\\\hib} & \makecell[c]{Tokens / Ep.\\(k)\\\lob} & \makecell[c]{Tool-Call\\Success (\%)\\\hib} & \makecell[c]{$\Delta$ vs.\\Qwen3.5-9B} & \makecell[c]{Pass@1\\All (\%)\\\hib} & \makecell[c]{Pass@1\\Easy (\%)\\\hib} & \makecell[c]{Pass@1\\Medium (\%)\\\hib} & \makecell[c]{Pass@1\\Hard (\%)\\\hib} & \makecell[c]{$\Delta$ vs.\\Qwen3.5-9B} \\
\midrule\midrule
\rowcolor{sectiongray}\multicolumn{11}{@{}l}{\textit{Base Model}} \\
Qwen3.5-9B & \cellcolor{royalblue!22}\underline{24.3} & \cellcolor{royalblue!15}26.4 & \cellcolor{royalblue!26}794 & \cellcolor{royalblue!6}75.5 & -- & \cellcolor{royalblue!33}69.3 & \cellcolor{royalblue!40}\textbf{97.4} & \cellcolor{royalblue!31}76.5 & \cellcolor{royalblue!34}\underline{35.7} & -- \\
\rowcolor{sectiongray}\multicolumn{11}{@{}l}{\textit{Ours: Curriculum Learning with \buggymle{} (9B)}} \\
SFT Only (9B) & \cellcolor{royalblue!9}18.3 & \cellcolor{royalblue!8}23.0 & \cellcolor{royalblue!40}\textbf{361} & \cellcolor{royalblue!34}95.1 & -- & \cellcolor{royalblue!6}61.3 & \cellcolor{royalblue!6}94.7 & \cellcolor{royalblue!6}65.7 & \cellcolor{royalblue!6}25.9 & -- \\
\multicolumn{11}{@{}l}{\quad\textit{SFT+RL Curriculum:}} \\
\qquad Easy Only (1--5 Hunks) & \cellcolor{royalblue!6}17.0 & \cellcolor{royalblue!6}21.7 & \cellcolor{royalblue!27}776 & \cellcolor{royalblue!32}93.6 & $-$7.3$^\dagger$ & \cellcolor{royalblue!30}68.3 & \cellcolor{royalblue!40}\textbf{97.4} & \cellcolor{royalblue!26}74.3 & \cellcolor{royalblue!32}35.0 & $-$1.0 \\
\qquad Hard Only (30--34 Hunks) & \cellcolor{royalblue!17}22.0 & \cellcolor{royalblue!20}\underline{29.2} & \cellcolor{royalblue!6}1428 & \cellcolor{royalblue!40}\textbf{99.2} & $-$2.3 & \cellcolor{royalblue!35}\underline{69.8} & \cellcolor{royalblue!34}\underline{96.9} & \cellcolor{royalblue!35}\underline{78.2} & \cellcolor{royalblue!34}35.6 & +0.4 \\
\qquad All Difficulties & \cellcolor{royalblue!40}\textbf{33.0} & \cellcolor{royalblue!40}\textbf{39.5} & \cellcolor{royalblue!32}\underline{616} & \cellcolor{royalblue!39}\underline{98.8} & +8.7$^\dagger$ & \cellcolor{royalblue!40}\textbf{71.3} & \cellcolor{royalblue!34}\underline{96.9} & \cellcolor{royalblue!40}\textbf{80.5} & \cellcolor{royalblue!40}\textbf{37.7} & +2.0$^\dagger$ \\
\bottomrule[1.5pt]
\end{tabular}
}
\end{table}

\paragraph{RQ5: Does training on \buggymle{} transfer beyond ML debugging?}\label{sec:ood} Training on \buggymle{} transfers positively to general software repair: the All Difficulties model resolves significantly more SWE-bench Lite issues than the untuned Qwen3.5-9B base model, whereas the narrower Easy Only and Hard Only curricula underperform the base model (Table~\ref{tab:regime_swe}). On single-turn algorithmic coding (LiveCodeBench~v6), SFT initially reduces pass@1 due to domain specialization, but subsequent GRPO training recovers this loss and slightly surpasses the base model. Taken together, these results show that reinforcement learning recovers general capabilities from a degraded SFT checkpoint, and that more diverse curricula can even improve those abilities beyond the base model. However, transfer remains bounded by task paradigm. When tasked with generating ML solutions from scratch, we observe mixed results on Kaggle leaderboards, with our model scoring slightly higher than the base model in some settings and lower in others, though tool-use reliability improves in every interactive setting (Appendix~\ref{app:ood_full}). Consistently improving from-scratch ML generation likely requires direct optimization on solution quality, such as a leaderboard-percentile reward.

\paragraph{RQ6: How realistic is our setup, and how does realism affect difficulty?}\label{sec:hard_bugs} Both automated and expert evaluations confirm that our injected bugs closely mirror real-world ML defects. An LLM judge rates nearly half of the evaluated bugs at 4 or 5 out of 5 for realism, and a domain expert judged 27 of 30 randomly sampled bugs to be plausible in production ML codebases (Appendix Figure~\ref{fig:bmbl_realism}). Higher realism directly increases repair difficulty, with bugs rated 4 or 5 exhibiting the lowest resolution rates across all evaluated models (Figure~\ref{fig:repair_depth}d). These realistic defects typically represent plausible but suboptimal modeling choices, such as misconfigured label smoothing or aggressive negative subsampling, that degrade performance on a specific dataset without introducing syntactic or logical errors detectable via static inspection.

\paragraph{RQ7: Which bug types are hardest to resolve, and does task modality matter?} Bug observability is the primary determinant of repair difficulty. Because official evaluation uses held-out test data, an agent's sandbox validation metric does not necessarily reflect test performance. We distinguish between \emph{visible} bugs, which directly degrade the printed validation metric, and \emph{silent} bugs, which leave validation scores intact and manifest only by corrupting test-time predictions (for instance, a label-indexing off-by-one error that preserves a 0.953 validation score while reducing the test score to 0.017). Across all models, silent bugs are resolved at substantially lower rates than visible ones (Figure~\ref{fig:repair_depth}c). Categorized by pipeline stage, output-formatting and submission defects prove most difficult for all models except those trained on high-complexity curricula (All Difficulties and Hard Only; Appendix Figure~\ref{fig:bug_resolve_by_type}). By contrast, data modality has minimal impact on frontier agents, with Claude Opus~5.5 and Sonnet~5.5 resolving comparable shares of individual bugs across all four \bmbl{} modalities (Appendix Figure~\ref{fig:bug_resolve_by_type}c).

\FloatBarrier

\section{Conclusion}
\label{sec:conclusion}
We show that code structure, rather than model identity, drives coding benchmark difficulty. The \cci{} captures this relationship using ten structural features and four parameters, accurately tracking solve rates across eight diverse public benchmarks. Operationalizing this insight, \buggymle{} scales the difficulty of fixed repositories via a controllable mutation budget, yielding \bmbl{}, a repository-level benchmark that remains unsaturated at its highest tiers. Training a 9B agent on an easy-to-hard \buggymle{} curriculum allows it to rival frontier models on 10 Bugs and generalize to broader code repair.
\paragraph{Limitations.} While \cci{} is stable under LOBO evaluation across eight human-curated benchmarks, validating it as a general law requires broader coverage, including independently generated synthetic sets (as \buggymle{} tasks were excluded to prevent circularity). Additionally, \cci{} relies purely on structural features under agentic evaluation; it does not capture semantic complexity and is less predictive on standalone-program benchmarks (e.g., FrontierCS Algo) than on the repository-scale tasks \buggymle{} targets. Finally, while \buggymle{} training strengthens software engineering and code repair, transfer to from-scratch ML solution generation yields mixed results.

\subsubsection*{Acknowledgments}

We thank the open-source benchmark community for making leaderboard data publicly available.

\bibliography{iclr2027_conference}

\begin{thebibliography}{49}
\providecommand{\natexlab}[1]{#1}
\providecommand{\url}[1]{\texttt{#1}}
\expandafter\ifx\csname urlstyle\endcsname\relax
  \providecommand{\doi}[1]{doi: #1}\else
  \providecommand{\doi}{doi: \begingroup \urlstyle{rm}\Url}\fi

\bibitem[{Anthropic}(2025)]{anthropic2025claudehaiku45}
{Anthropic}.
\newblock {Claude Haiku 4.5 System Card}, October 2025.
\newblock URL
  \url{https://www-cdn.anthropic.com/7aad69bf12627d42234e01ee7c36305dc2f6a970.pdf}.

\bibitem[{Anthropic}(2026{\natexlab{a}})]{anthropic2026claudehaiku55}
{Anthropic}.
\newblock {Claude Haiku 5.5 System Card}, October 7, 2026{\natexlab{a}}.
\newblock URL \url{https://www.anthropic.com/claude-haiku-5-5-system-card}.

\bibitem[{Anthropic}(2026{\natexlab{b}})]{anthropic2026claudeopus48}
{Anthropic}.
\newblock {Claude Opus 4.8 System Card}, May 28, 2026{\natexlab{b}}.
\newblock URL \url{https://www.anthropic.com/claude-opus-4-8-system-card}.

\bibitem[{Anthropic}(2026{\natexlab{c}})]{anthropic2026claudeopus5}
{Anthropic}.
\newblock {Claude Opus 5 System Card}, July 24, 2026{\natexlab{c}}.
\newblock URL \url{https://www.anthropic.com/claude-opus-5-system-card}.

\bibitem[{Anthropic}(2026{\natexlab{d}})]{anthropic2026claudeopus55}
{Anthropic}.
\newblock {Claude Opus 5.5 System Card}, September 22, 2026{\natexlab{d}}.
\newblock URL \url{https://www.anthropic.com/claude-opus-5-5-system-card}.

\bibitem[{Anthropic}(2026{\natexlab{e}})]{anthropic2026claudesonnet46}
{Anthropic}.
\newblock {Claude Sonnet 4.6 System Card}, February 17, 2026{\natexlab{e}}.
\newblock URL \url{https://www.anthropic.com/claude-sonnet-4-6-system-card}.

\bibitem[{Anthropic}(2026{\natexlab{f}})]{anthropic2026claudesonnet5}
{Anthropic}.
\newblock {Claude Sonnet 5 System Card}, June 30, 2026{\natexlab{f}}.
\newblock URL \url{https://www.anthropic.com/claude-sonnet-5-system-card}.

\bibitem[{Anthropic}(2026{\natexlab{g}})]{anthropic2026claudesonnet55}
{Anthropic}.
\newblock {Claude Sonnet 5.5 System Card}, September 28, 2026{\natexlab{g}}.
\newblock URL \url{https://www.anthropic.com/claude-sonnet-5-5-system-card}.

\bibitem[Chan et~al.(2025)Chan, Chowdhury, Jaffe, Aung, Sherburn, Mays,
  Starace, Liu, Maksin, Patwardhan, Weng, and M{\k{a}}dry]{chan2025mlebench}
J.~S. Chan, N.~Chowdhury, O.~Jaffe, J.~Aung, D.~Sherburn, E.~Mays, G.~Starace,
  K.~Liu, L.~Maksin, T.~Patwardhan, L.~Weng, and A.~M{\k{a}}dry.
\newblock {MLE-bench}: Evaluating machine learning agents on machine learning
  engineering.
\newblock In \emph{International Conference on Learning Representations
  (ICLR)}, 2025.

\bibitem[Chen et~al.(2025{\natexlab{a}})Chen, Prabhudesai, Fragkiadaki, Liu,
  and Pathak]{chen2025self}
L.~Chen, M.~Prabhudesai, K.~Fragkiadaki, H.~Liu, and D.~Pathak.
\newblock Self-questioning language models.
\newblock \emph{arXiv preprint arXiv:2508.03682}, 2025{\natexlab{a}}.

\bibitem[Chen et~al.(2026)Chen, Wang, and Qu]{chen2026recursive}
M.~Chen, L.~Wang, and B.~Qu.
\newblock Recursive self-improvement in {AI}: From bounded self-refinement to
  autonomous research loops.
\newblock \emph{arXiv preprint arXiv:2607.07663}, 2026.

\bibitem[Chen et~al.(2025{\natexlab{b}})Chen, Yang, Qin, Liu, Yan, Guan, Peng,
  Ji, Li, Hu, Zhang, Liang, Zhou, Wang, Chen, and
  Che]{chen2025ai4researchsurveyartificialintelligence_for_research}
Q.~Chen, M.~Yang, L.~Qin, J.~Liu, Z.~Yan, J.~Guan, D.~Peng, Y.~Ji, H.~Li,
  M.~Hu, Y.~Zhang, Y.~Liang, Y.~Zhou, J.~Wang, Z.~Chen, and W.~Che.
\newblock {AI4Research}: A survey of artificial intelligence for scientific
  research.
\newblock \emph{arXiv preprint arXiv:2507.01903}, 2025{\natexlab{b}}.
\newblock URL \url{https://arxiv.org/abs/2507.01903}.

\bibitem[Chu et~al.(2026)Chu, Agarwal, Thangamuthu, Graham, and
  Mattern]{chu2026frontierswe}
E.~Chu, R.~Agarwal, A.~Thangamuthu, B.~Graham, and J.~Mattern.
\newblock {FrontierSWE}: Benchmarking coding agents at the limits of human
  abilities, 2026.
\newblock URL \url{https://frontierswe.com}.

\bibitem[Deng et~al.(2025)Deng, Da, Pan, He, Ide, Garg, Lauffer, Park, Pasari,
  Rane, Sampath, Krishnan, Kundurthy, Hendryx, Wang, Bharadwaj, Holm, Aluri,
  Zhang, Jacobson, Liu, and Kenstler]{deng2025swebenchproaiagents}
X.~Deng, J.~Da, E.~Pan, Y.~Y. He, C.~Ide, K.~Garg, N.~Lauffer, A.~Park,
  N.~Pasari, C.~Rane, K.~Sampath, M.~Krishnan, S.~Kundurthy, S.~Hendryx,
  Z.~Wang, V.~Bharadwaj, J.~Holm, R.~Aluri, C.~B.~C. Zhang, N.~Jacobson,
  B.~Liu, and B.~Kenstler.
\newblock {SWE-bench Pro}: Can {AI} agents solve long-horizon software
  engineering tasks?
\newblock \emph{arXiv preprint arXiv:2509.16941}, 2025.
\newblock URL \url{https://arxiv.org/abs/2509.16941}.

\bibitem[Desai et~al.(2026)Desai, Hu, Cabezas, Harsola, Shukla, Chaim, Assadi,
  Kamath, Faldu, Hebbar, et~al.]{desai2026swe_swe_marathon}
R.~Desai, J.~Hu, J.~Cabezas, N.~Harsola, P.~Shukla, R.~B. Chaim, A.~E. Assadi,
  O.~M. Kamath, F.~Faldu, P.~Hebbar, et~al.
\newblock {SWE-Marathon}: Can agents autonomously complete ultra-long-horizon
  software work?
\newblock \emph{arXiv preprint arXiv:2606.07682}, 2026.

\bibitem[{Google DeepMind}(2026{\natexlab{a}})]{deepmind2026gemini35flashlite}
{Google DeepMind}.
\newblock {Gemini 3.5 Flash-Lite Model Card}, July 21, 2026{\natexlab{a}}.
\newblock URL
  \url{https://deepmind.google/models/model-cards/gemini-3-5-flash-lite/}.

\bibitem[{Google DeepMind}(2026{\natexlab{b}})]{deepmind2026gemini36flash}
{Google DeepMind}.
\newblock {Gemini 3.6 Flash Model Card}, July 21, 2026{\natexlab{b}}.
\newblock URL
  \url{https://deepmind.google/models/model-cards/gemini-3-6-flash/}.

\bibitem[{Google DeepMind}(2026{\natexlab{c}})]{deepmind2026gemini37flash}
{Google DeepMind}.
\newblock {Gemini 3.7 Flash Model Card}, August 13, 2026{\natexlab{c}}.
\newblock URL
  \url{https://deepmind.google/models/model-cards/gemini-3-7-flash/}.

\bibitem[{Google DeepMind}(2026{\natexlab{d}})]{deepmind2026gemini38flash}
{Google DeepMind}.
\newblock {Gemini 3.8 Flash Model Card}, September 2, 2026{\natexlab{d}}.
\newblock URL
  \url{https://deepmind.google/models/model-cards/gemini-3-8-flash/}.

\bibitem[He et~al.(2026)He, Mang, Zhou, Liu, Li, Mao, Zhang, Li, Peng, Cheng,
  et~al.]{he2026frontiersmith}
R.~He, Q.~Mang, S.~Zhou, K.~Liu, H.~Li, H.~Mao, Q.~Zhang, Z.~Li, B.~Peng,
  L.~Cheng, et~al.
\newblock {FrontierSmith}: Synthesizing open-ended coding problems at scale.
\newblock \emph{arXiv preprint arXiv:2605.14445}, 2026.

\bibitem[Huang et~al.(2025)Huang, Feng, Chen, Zhao, Cheng, Bai, Zhou, Li, and
  Qin]{huang2025mldebugging}
J.~Huang, X.~Feng, Q.~Chen, H.~Zhao, Z.~Cheng, J.~Bai, J.~Zhou, M.~Li, and
  L.~Qin.
\newblock {MLDebugging}: Towards benchmarking code debugging across
  multi-library scenarios.
\newblock In \emph{Findings of the Association for Computational Linguistics:
  ACL 2025}, pages 5866--5879, 2025.

\bibitem[Huang et~al.(2023)Huang, Vora, Liang, and
  Leskovec]{huang2023mlagentbench}
Q.~Huang, J.~Vora, P.~Liang, and J.~Leskovec.
\newblock {MLAgentBench}: Evaluating language agents on machine learning
  experimentation.
\newblock \emph{arXiv preprint arXiv:2310.03302}, 2023.

\bibitem[Huang et~al.(2026)Huang, Lee, Tng, and Ge]{huang2026deepswe}
W.~Huang, C.~Lee, L.~Tng, and S.~Ge.
\newblock {DeepSWE}: Measuring frontier coding agents on original, long-horizon
  engineering tasks.
\newblock \emph{arXiv preprint arXiv:2607.07946}, 2026.

\bibitem[Jain et~al.(2025)Jain, Han, Gu, Li, Yan, Zhang, Wang, Solar-Lezama,
  Sen, and Stoica]{jain2025livecodebench}
N.~Jain, K.~Han, A.~Gu, W.-D. Li, F.~Yan, T.~Zhang, S.~Wang, A.~Solar-Lezama,
  K.~Sen, and I.~Stoica.
\newblock {LiveCodeBench}: Holistic and contamination free evaluation of large
  language models for code.
\newblock In \emph{The Thirteenth International Conference on Learning
  Representations}, 2025.
\newblock URL \url{https://openreview.net/forum?id=chfJJYC3iL}.

\bibitem[Jimenez et~al.(2024)Jimenez, Yang, Wettig, Yao, Pei, Press, and
  Narasimhan]{jimenez2024swe}
C.~E. Jimenez, J.~Yang, A.~Wettig, S.~Yao, K.~Pei, O.~Press, and K.~Narasimhan.
\newblock {SWE-bench}: Can language models resolve real-world {GitHub} issues?
\newblock In \emph{International Conference on Learning Representations},
  volume 2024, pages 54107--54157, 2024.

\bibitem[Kim et~al.(2021)Kim, Kim, and Lee]{kim2021denchmark}
M.~Kim, Y.~Kim, and E.~Lee.
\newblock {Denchmark}: A bug benchmark of deep learning-related software.
\newblock In \emph{2021 IEEE/ACM 18th International Conference on Mining
  Software Repositories (MSR)}, pages 540--544. IEEE, 2021.

\bibitem[Lu et~al.(2026)Lu, Pan, Birlikci, Lee, Wang, Choudhury, Ma, Qin,
  Baronio, Alberti, et~al.]{lu2026introducing}
E.~Lu, B.~Pan, D.~Birlikci, S.~Lee, R.~Wang, R.~Choudhury, F.~Ma, T.~Qin,
  C.~Baronio, S.~Alberti, et~al.
\newblock Introducing {FrontierCode}.
\newblock \emph{Cognition research blog}, 2026.

\bibitem[Lu et~al.(2025)Lu, Yang, Ren, Hou, Xiao, Wang, Shi, Zhou, Zhan, and
  Li]{lu2026webgen}
Z.~Lu, Y.~Yang, H.~Ren, H.~Hou, H.~Xiao, K.~Wang, W.~Shi, A.~Zhou, M.~Zhan, and
  H.~Li.
\newblock {WebGen-Bench}: Evaluating {LLMs} on generating interactive and
  functional websites from scratch.
\newblock \emph{Advances in Neural Information Processing Systems}, 38, 2025.

\bibitem[Lyu et~al.(2026)Lyu, Yang, Huang, Zhang, Xu, Li, Han, Zhang, Zhang,
  Huang, et~al.]{lyu2026mls}
B.~Lyu, Y.~Yang, S.~Huang, J.~Zhang, Q.~Xu, X.~Li, X.~Han, Y.~Zhang, H.~Zhang,
  R.~Huang, et~al.
\newblock {MLS-Bench}: A holistic and rigorous assessment of {AI} systems on
  building better {AI}.
\newblock \emph{arXiv preprint arXiv:2605.08678}, 2026.

\bibitem[Mang et~al.(2025)Mang, Chai, Li, Mao, Zhou, Du, Li, Liu, Chen, Wang,
  et~al.]{mang2025frontiercs}
Q.~Mang, W.~Chai, Z.~Li, H.~Mao, S.~Zhou, A.~Du, H.~Li, S.~Liu, E.~Chen,
  Y.~Wang, et~al.
\newblock {FrontierCS}: Evolving challenges for evolving intelligence.
\newblock \emph{arXiv preprint arXiv:2512.15699}, 2025.

\bibitem[Morovati et~al.(2023)Morovati, Nikanjam, Khomh, and
  Jiang]{morovati2023bugs_defect4ML}
M.~M. Morovati, A.~Nikanjam, F.~Khomh, and Z.~M. Jiang.
\newblock Bugs in machine learning-based systems: a faultload benchmark.
\newblock \emph{Empirical Software Engineering}, 28\penalty0 (3):\penalty0 62,
  2023.

\bibitem[Nathani et~al.(2025)Nathani, Madaan, Roberts, Bashlykov, Menon, Moens,
  Budhiraja, Magka, Vorotilov, Chaurasia, et~al.]{nathani2025mlgym}
D.~Nathani, L.~Madaan, N.~Roberts, N.~Bashlykov, A.~Menon, V.~Moens,
  A.~Budhiraja, D.~Magka, V.~Vorotilov, G.~Chaurasia, et~al.
\newblock {MLGym}: A new framework and benchmark for advancing {AI} research
  agents.
\newblock \emph{arXiv preprint arXiv:2502.14499}, 2025.

\bibitem[Novikov et~al.(2025)Novikov, V{\~u}, Eisenberger, Dupont, Huang,
  Wagner, Shirobokov, Kozlovskii, Ruiz, Mehrabian, Kumar, See, Chaudhuri,
  Holland, Davies, Nowozin, Kohli, and
  Balog]{novikov2025alphaevolvecodingagentscientific}
A.~Novikov, N.~V{\~u}, M.~Eisenberger, E.~Dupont, P.-S. Huang, A.~Z. Wagner,
  S.~Shirobokov, B.~Kozlovskii, F.~J.~R. Ruiz, A.~Mehrabian, M.~P. Kumar,
  A.~See, S.~Chaudhuri, G.~Holland, A.~Davies, S.~Nowozin, P.~Kohli, and
  M.~Balog.
\newblock {AlphaEvolve}: A coding agent for scientific and algorithmic
  discovery.
\newblock \emph{arXiv preprint arXiv:2506.13131}, 2025.
\newblock URL \url{https://arxiv.org/abs/2506.13131}.

\bibitem[Pan et~al.(2024)Pan, Wang, Neubig, Jaitly, Ji, Suhr, and
  Zhang]{pan2024training_swegym}
J.~Pan, X.~Wang, G.~Neubig, N.~Jaitly, H.~Ji, A.~Suhr, and Y.~Zhang.
\newblock Training software engineering agents and verifiers with {SWE-Gym}.
\newblock \emph{arXiv preprint arXiv:2412.21139}, 2024.

\bibitem[Qiang et~al.(2025)Qiang, Zhuang, Li, Sagar~VK, Zhang, Li, Wong, Yang,
  Liang, Zhang, and Dai]{qiang2026mle}
R.~Qiang, Y.~Zhuang, Y.~Li, D.~Sagar~VK, R.~Zhang, C.~Li, I.~Wong, S.~Yang,
  P.~Liang, C.~Zhang, and B.~Dai.
\newblock {MLE-Dojo}: Interactive environments for empowering {LLM} agents in
  machine learning engineering.
\newblock \emph{Advances in Neural Information Processing Systems}, 38, 2025.

\bibitem[Qiang et~al.(2026{\natexlab{a}})Qiang, Li, Sun, Zhuang, Zhang, and
  Dai]{qiang2026matryoshka}
R.~Qiang, C.~Li, H.~Sun, Y.~Zhuang, C.~Zhang, and B.~Dai.
\newblock {Matryoshka Agent}: Unfolding sub-agents for long-horizon machine
  learning engineering.
\newblock \emph{arXiv preprint arXiv:2607.25090}, 2026{\natexlab{a}}.

\bibitem[Qiang et~al.(2026{\natexlab{b}})Qiang, Zhuang, Singh, Liang, Zhang,
  Yang, and Dai]{qiang2026mlesmith}
R.~Qiang, Y.~Zhuang, A.~Singh, P.~Liang, C.~Zhang, S.~Yang, and B.~Dai.
\newblock {MLE-Smith}: Scaling {MLE} tasks with automated multi-agent pipeline.
\newblock In \emph{International Conference on Learning Representations},
  volume 2026, pages 5963--5990, 2026{\natexlab{b}}.

\bibitem[{Qwen Team}(2026{\natexlab{a}})]{qwen3.5}
{Qwen Team}.
\newblock {Qwen3.5}: Towards native multimodal agents, February
  2026{\natexlab{a}}.
\newblock URL \url{https://qwen.ai/blog?id=qwen3.5}.

\bibitem[{Qwen Team}(2026{\natexlab{b}})]{qwen38}
{Qwen Team}.
\newblock {Qwen3.8-Max}: A new bar for coding and cowork, August
  2026{\natexlab{b}}.
\newblock URL \url{https://qwen.ai/blog?id=qwen3.8}.

\bibitem[Shao et~al.(2024)Shao, Wang, Zhu, Xu, Song, Bi, Zhang, Zhang, Li, Wu,
  and Guo]{shao2024deepseekmathpushinglimitsmathematical_grpo}
Z.~Shao, P.~Wang, Q.~Zhu, R.~Xu, J.~Song, X.~Bi, H.~Zhang, M.~Zhang, Y.~K. Li,
  Y.~Wu, and D.~Guo.
\newblock {DeepSeekMath}: Pushing the limits of mathematical reasoning in open
  language models.
\newblock \emph{arXiv preprint arXiv:2402.03300}, 2024.
\newblock URL \url{https://arxiv.org/abs/2402.03300}.

\bibitem[Shen et~al.(2026)Shen, Tormoen, Shah, Farhadi, and
  Dettmers]{shen2026sera}
E.~Shen, D.~Tormoen, S.~Shah, A.~Farhadi, and T.~Dettmers.
\newblock {SERA}: Soft-verified efficient repository agents.
\newblock \emph{arXiv preprint arXiv:2601.20789}, 2026.

\bibitem[Shi et~al.(2026)Shi, Cao, Chen, Sun, Li, Lu, Dong, Qin, Liu, Jiang,
  et~al.]{shi2026taskcraft}
D.~Shi, J.~Cao, Q.~Chen, W.~Sun, W.~Li, H.~Lu, F.~Dong, T.~Qin, M.~Liu,
  Y.~Jiang, et~al.
\newblock {TaskCraft}: Automated generation of agentic tasks.
\newblock In \emph{International Conference on Learning Representations},
  volume 2026, pages 43714--43734, 2026.

\bibitem[Wu et~al.(2026)Wu, Fan, Wu, Landes, Schrock, Razin, Chatterjee,
  Baskaran, Steier, Fitzpatrick, Arif, Atri, Pradeepkumar, Laghuvarapu, Gao,
  Cross, and Sun]{wu2026pyhealth20comprehensiveopensource}
J.~Wu, Y.~Fan, Z.~Wu, P.~Landes, E.~Schrock, S.~S. Razin, A.~Chatterjee,
  N.~Baskaran, J.~Steier, A.~Fitzpatrick, B.~Arif, R.~Atri, J.~Pradeepkumar,
  S.~Laghuvarapu, J.~Gao, A.~R. Cross, and J.~Sun.
\newblock {PyHealth 2.0}: A comprehensive open-source toolkit for accessible
  and reproducible clinical deep learning.
\newblock \emph{arXiv preprint arXiv:2601.16414}, 2026.
\newblock URL \url{https://arxiv.org/abs/2601.16414}.

\bibitem[Yang et~al.(2025{\natexlab{a}})Yang, Lieret, Jimenez, Wettig,
  Khandpur, Zhang, Hui, Press, Schmidt, and Yang]{yang2026swe}
J.~Yang, K.~Lieret, C.~Jimenez, A.~Wettig, K.~Khandpur, Y.~Zhang, B.~Hui,
  O.~Press, L.~Schmidt, and D.~Yang.
\newblock {SWE-smith}: Scaling data for software engineering agents.
\newblock \emph{Advances in Neural Information Processing Systems}, 38,
  2025{\natexlab{a}}.

\bibitem[Yang et~al.(2026{\natexlab{a}})Yang, Jiang, Fu, Luo, Ren, Wang, Zhao,
  Liu, Zuo, Wang, et~al.]{yang2026frontis_Openmle}
J.~Yang, C.~Jiang, Y.~Fu, T.~Luo, C.~Ren, W.~Wang, K.~Zhao, H.~Liu, Y.~Zuo,
  Y.~Wang, et~al.
\newblock {Frontis-MA1}: Training an {AI4AI} model towards recursive
  self-improvement in machine learning engineering.
\newblock \emph{arXiv preprint arXiv:2607.28568}, 2026{\natexlab{a}}.

\bibitem[Yang et~al.(2026{\natexlab{b}})Yang, Lieret, Ma, Thakkar, Pedchenko,
  Sootla, McMilin, Yin, Hou, Synnaeve, et~al.]{yang2026programbench}
J.~Yang, K.~Lieret, J.~Ma, P.~Thakkar, D.~Pedchenko, S.~Sootla, E.~McMilin,
  P.~Yin, R.~Hou, G.~Synnaeve, et~al.
\newblock {ProgramBench}: Can language models rebuild programs from scratch?
\newblock \emph{arXiv preprint arXiv:2605.03546}, 2026{\natexlab{b}}.

\bibitem[Yang et~al.(2025{\natexlab{b}})Yang, Wang, Yan, and
  Deng]{yang2025stop_dsdbench}
Z.~Yang, S.~Wang, Y.~Yan, and Y.~Deng.
\newblock Why stop at one error? {Benchmarking} {LLMs} as data science code
  debuggers for multi-hop and multi-bug errors.
\newblock In \emph{Proceedings of the 2025 Conference on Empirical Methods in
  Natural Language Processing}, pages 17377--17392, 2025{\natexlab{b}}.

\bibitem[Zhou et~al.(2026)Zhou, Liu, Deng, Zeng, Zhang, Zhu, Cai, Mao, Zhang,
  Tan, et~al.]{zhou2026autocodebench}
C.~Zhou, A.~Liu, Y.~Deng, Z.~Zeng, T.~Zhang, H.~Zhu, J.~Cai, Y.~Mao, C.~Zhang,
  L.~Tan, et~al.
\newblock {AutoCodeBench}: Large language models are automatic code benchmark
  generators.
\newblock In \emph{International Conference on Learning Representations},
  volume 2026, pages 79116--79148, 2026.

\bibitem[Zhou et~al.(2025)Zhou, Levine, Weston, Li, and
  Sukhbaatar]{zhou2026self}
Y.~Zhou, S.~Levine, J.~Weston, X.~Li, and S.~Sukhbaatar.
\newblock Self-challenging language model agents.
\newblock \emph{Advances in Neural Information Processing Systems},
  38:\penalty0 113959--113991, 2025.

\end{thebibliography}

\clearpage
\appendix
\section{\cci{} Validation and Scaling Sweep}

\subsection{Feature Definitions and Benchmark Statistics}
\label{app:features}

To complement Section~\ref{sec:cci}, Table~\ref{tab:features-full} defines the ten task-level median (p50) structural features grouped into repository scale, dependency complexity, and bug/patch complexity.

\begin{table}[!htbp]
\caption{Raw features used in \cci{}.}
\label{tab:features-full}
\centering
\small
\begin{tabular}{lll}
\toprule
\textbf{Group} & \textbf{Feature (p50)} & \textbf{Description} \\
\midrule
\multirow{3}{*}{Repo Scale} & lines\_of\_code & Total source lines \\
 & file\_count & Number of source files \\
 & max\_dir\_depth & Deepest directory nesting \\
\midrule
\multirow{2}{*}{Dependency} & dependency\_count & Number of inter-file dependencies \\
 & max\_dep\_chain & Longest dependency chain \\
\midrule
\multirow{5}{*}{Patch} & files\_changed & Files modified by the bug fix \\
 & hunks & Number of diff hunks \\
 & lines\_added & Lines added in the fix \\
 & lines\_removed & Lines removed in the fix \\
 & total\_lines\_delta & Net line change \\
\bottomrule
\end{tabular}
\end{table}

Table~\ref{tab:benchmarks} reports the resulting \cci{} score, task count, task type, language, and mean frontier-model solve rate for each of the eight public coding benchmarks used to fit the index.

\begin{table}[!htbp]
\caption{Benchmarks ordered by \cci{}. Solve rates average all evaluated frontier models. SWE-Marathon counts the 17 of its 20 tasks with readable reference code.}
\label{tab:benchmarks}
\centering
\footnotesize
\begin{tabular}{lclcc c}
\toprule
\textbf{Benchmark} & \textbf{Tasks} & \textbf{Task type} & \textbf{Language} & \textbf{\cci{}} & \textbf{Solve rate} \\
\midrule
SWE-bench Verified & 500 & Repo issue repair & Python & 0.00 & 0.71 \\
FrontierCS Research & 61 & Research code generation & Python & 0.31 & 0.46 \\
FrontierCS Algo & 188 & Algorithmic & C++ & 0.38 & 0.39 \\
SWE-bench Pro & 731 & Long-horizon repo repair & Multi & 0.48 & 0.31 \\
FrontierSWE & 17 & From-scratch engineering & Multi & 0.58 & 0.30 \\
DeepSWE & 113 & Long-horizon engineering & Multi & 0.68 & 0.27 \\
SWE-Marathon & 17 & Long-horizon project engineering & Multi & 0.69 & 0.11 \\
ProgramBench & 201 & From-scratch reconstruction & Multi & 1.00 & 0.00 \\
\bottomrule
\end{tabular}
\end{table}

\FloatBarrier
\subsection{Pipeline Ablations and LOBO Stability}
\label{app:validation}

We ablate each design choice of the \cci{} pipeline across all eight public benchmarks ($105$ model--benchmark pairs) and verify that alternative specifications leave our conclusions unchanged.

\paragraph{Log transformation.}
\label{app:log}

Applying $\log(1 + f_i)$ reduces skewness on all 10 raw features (Table~\ref{tab:skew}). Refitting the regression on untransformed z-scores preserves the sign of every group coefficient, confirming that log transformation improves numerical conditioning without altering qualitative findings.

\begin{table}[!htbp]
\caption{Skewness before and after $\log(1+f_i)$.}
\label{tab:skew}
\centering
\small
\begin{tabular}{lccc}
\toprule
\textbf{Feature ($f_i$)} & \textbf{Raw skew} & \textbf{$\log(1{+}f_i)$ skew} & \textbf{Improved?} \\
\midrule
lines\_of\_code & 1.70 & $-0.64$ & yes \\
file\_count & 1.46 & $-0.05$ & yes \\
max\_dir\_depth & 0.27 & $-0.07$ & yes \\
dependency\_count & 1.14 & $-0.10$ & yes \\
max\_dep\_chain & 2.23 & 1.26 & yes \\
files\_changed & 2.66 & 1.32 & yes \\
hunks & 2.25 & 0.38 & yes \\
lines\_added & 1.94 & $-0.02$ & yes \\
lines\_removed & 0.79 & 0.44 & yes \\
total\_lines\_delta & 1.94 & $-0.01$ & yes \\
\bottomrule
\end{tabular}

\end{table}

\paragraph{Feature grouping.}
\label{app:group}

\begin{figure}[!htbp]
\centering
\includegraphics[width=\linewidth]{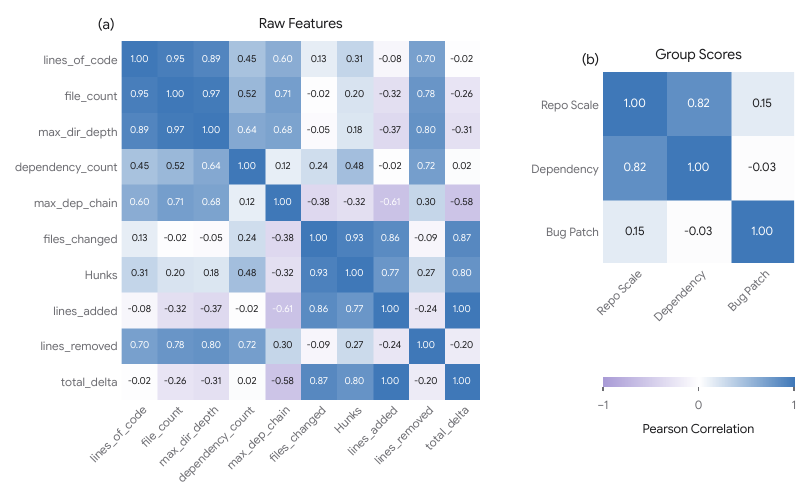}
\caption{\textbf{Grouping the ten structural features into three dimensions eliminates severe within-group collinearity.} Pairwise Pearson correlations across the eight public coding benchmarks (log-transformed task medians; blue: positive, lavender: negative). \textbf{(a)} Raw features exhibit near-collinearity within conceptual blocks (e.g., \texttt{file\_count} vs.\ \texttt{max\_dir\_depth}, $r = 0.97$; \texttt{files\_changed} vs.\ \texttt{hunks}, $r = 0.93$), which destabilizes an ungrouped regression. \textbf{(b)} Aggregating into Repo Scale, Dependency, and Bug Patch z-score averages lowers the maximum off-diagonal correlation to $r = 0.82$ while keeping Bug Patch nearly orthogonal to both repository-level groups ($r = 0.15$ and $-0.03$).}
\label{fig:collinearity}
\end{figure}

Because features within the same structural dimension covary strongly, fitting an ungrouped ten-feature logistic regression suffers from severe multicollinearity. Averaging z-scores within the three conceptual groups reduces the maximum pairwise correlation from $1.00$ to $0.82$ (Figure~\ref{fig:collinearity}), with balanced variance contributions across constituent features.

\paragraph{Link-function robustness.}
\label{app:link}

Both weighted least squares (WLS) with an identity link and a binomial generalized linear model (GLM) with a probit link yield identical coefficient signs to our logistic specification. Because binomial counts across tasks exhibit over-dispersion (Pearson $\chi^2 / \mathrm{df} \approx 30$), naive parametric standard errors underestimate uncertainty; we therefore evaluate generalization via non-parametric LOBO cross-validation rather than in-sample $p$-values.

\paragraph{LOBO coefficient stability.}
\label{app:signstab}

\begin{figure}[!htbp]
\centering
\includegraphics[width=\linewidth]{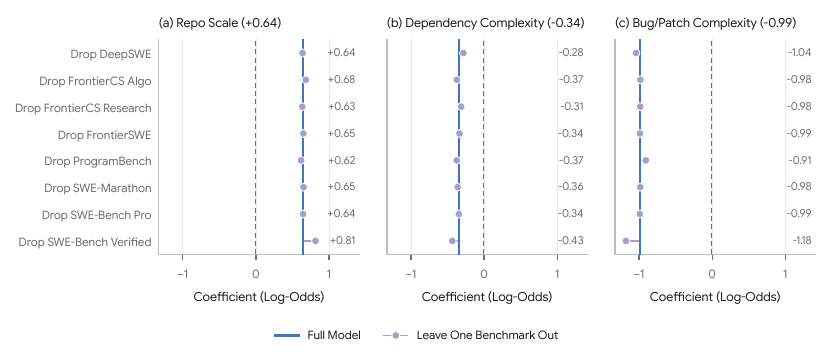}
\caption{\textbf{All three \cci{} group coefficients preserve their sign and relative magnitude across every LOBO refit.} Binomial GLM log-odds coefficients when each of the eight benchmarks is held out in turn (lavender dots and horizontal offsets, exact values at right) compared against the full eight-benchmark fit (vertical blue line; zero shown dashed) for \textbf{(a)} Repo Scale ($+0.64$), \textbf{(b)} Dependency Complexity ($-0.34$), and \textbf{(c)} Bug/Patch Complexity ($-0.99$). No single benchmark drives the fit or flips any coefficient's sign, and Bug/Patch Complexity remains the dominant predictor in every refit ($-0.91$ to $-1.18$).}
\label{fig:lobo_coefs}
\end{figure}

Given eight reference benchmarks, a natural concern is whether a single outlier benchmark drives the regression. LOBO refits rule this out: holding out any benchmark preserves all three coefficient signs and relative magnitudes (Figure~\ref{fig:lobo_coefs}), and held-out pass-rate predictions remain tightly aligned with the identity line (Figure~\ref{fig:lobo_robust}).

\paragraph{Model-specification ablation.}
\label{app:ablation}

Table~\ref{tab:ablation} compares nine regression specifications by held-out LOBO mean absolute error. Among specifications whose predictions remain valid solve rates in $[0, 1]$, the four-parameter, complexity-only logistic model achieves the lowest out-of-sample error ($\text{MAE} = 0.115$). OLS reaches $0.105$ only because 10 of its 82 held-out predictions (over models appearing in at least two benchmarks) fall below zero and are clipped to the $0\%$ observed on ProgramBench. Adding per-model fixed effects more than doubles held-out error ($0.239$), underperforming even an intercept-only baseline ($0.227$), because leaderboard model subsets vary across benchmarks and overfit benchmark-level intercepts; we therefore exclude model identity from \cci{}. Because \cci{} is model-agnostic, per-(model, benchmark) pair MAE in Table~\ref{tab:ablation} ($0.115$) reflects within-benchmark spread across models, whereas benchmark-average pass rates in Figure~\ref{fig:lobo_robust}b achieve $\text{MAE} = 0.048$; likewise, $\beta_{\text{repo}} = +0.64$ (Figure~\ref{fig:lobo_coefs}a) acts as a conditional adjustment for suites with large repositories alongside small localized patches (such as SWE-bench Verified).

\begin{table}[!htbp]
\caption{LOBO prediction error by model specification. MAE: mean absolute error of the held-out solve rate; OLS: ordinary least squares; WLS: weighted least squares. Linear specifications can predict solve rates outside $[0, 1]$ and are clipped before scoring. \textbf{Bold}: lowest error among specifications whose predictions stay within $[0, 1]$.}
\label{tab:ablation}
\centering
\small
\begin{tabular}{lccc}
\toprule
\textbf{Approach} & \textbf{Params} & \textbf{Predicts in $[0,1]$} & \textbf{LOBO MAE} \\
\midrule
3 complexity groups (logistic regression) & 4 & \cmark & \textbf{0.115} \\
+ model-family fixed effects & $\sim$17 & \cmark & 0.130 \\
Bug patch only & 2 & \cmark & 0.166 \\
Patch + dependency & 3 & \cmark & 0.167 \\
+ intelligence index & 5 & \cmark & 0.173 \\
Intercept only (baseline) & 1 & \cmark & 0.227 \\
+ per-model fixed effects & $\sim$54 & \cmark & 0.239 \\
\midrule
OLS + 3 groups & 4 & \xmark & 0.105 \\
WLS + 3 groups & 4 & \xmark & 0.133 \\
\bottomrule
\end{tabular}

\end{table}

\FloatBarrier
\subsection{Feature Group Contributions}
\label{app:groups}

\begin{figure}[!htbp]
\centering
\includegraphics[width=\linewidth]{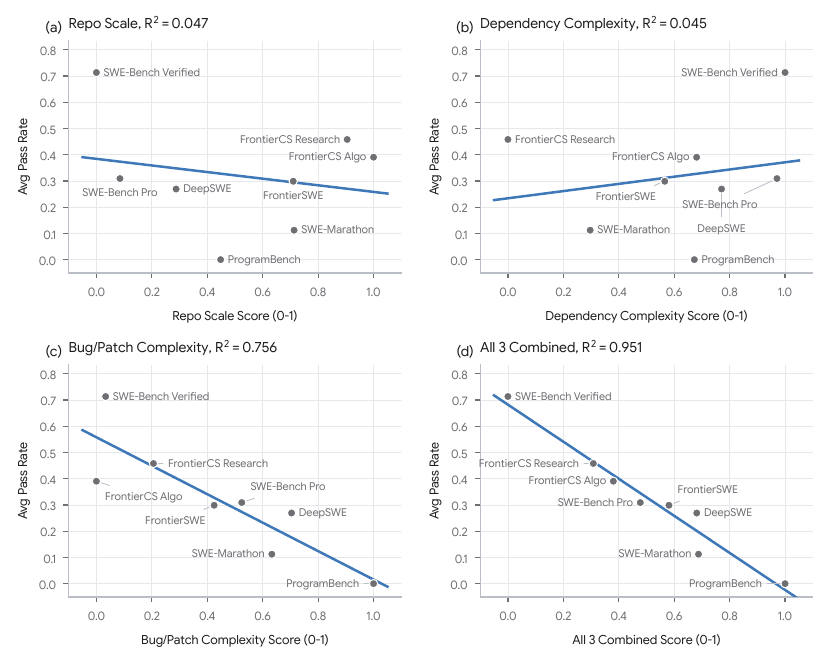}
\caption{\textbf{Bug/patch complexity alone explains three-quarters of benchmark difficulty variance, and combining all three groups reaches $R^2 = 0.951$.} Average benchmark pass rate across evaluated models (gray points, directly labeled) regressed via OLS (blue line) onto each min--max scaled score in $[0, 1]$: \textbf{(a)} Repo Scale ($R^2 = 0.047$), \textbf{(b)} Dependency Complexity ($R^2 = 0.045$), \textbf{(c)} Bug/Patch Complexity ($R^2 = 0.756$), and \textbf{(d)} the full three-group \cci{} ($R^2 = 0.951$). Repository size and dependency structure are weak standalone predictors because small single-file algorithmic tasks (e.g., FrontierCS Algo) can still be difficult, whereas combining them with patch scope aligns all eight benchmarks tightly along a single linear frontier.}
\label{fig:individual_r2}
\end{figure}

To isolate the predictive power of each structural dimension in Section~\ref{sec:cci}, Figure~\ref{fig:individual_r2} regresses average benchmark solve rate onto each group score individually.

Bug/patch complexity accounts for the largest share of variance ($R^2 = 0.756$), reflecting the difficulty of coordinating edits across multiple files and diff hunks. Dependency complexity and repository scale are weak standalone predictors ($R^2 = 0.045$ and $0.047$) because compact single-file algorithmic benchmarks remain challenging, yet both provide complementary signal that raises the joint fit to $R^2 = 0.951$.

\subsection{Model Capability vs.\ Benchmark Difficulty}
\label{app:intelligence}

\begin{figure}[!htbp]
\centering
\includegraphics[width=0.62\linewidth]{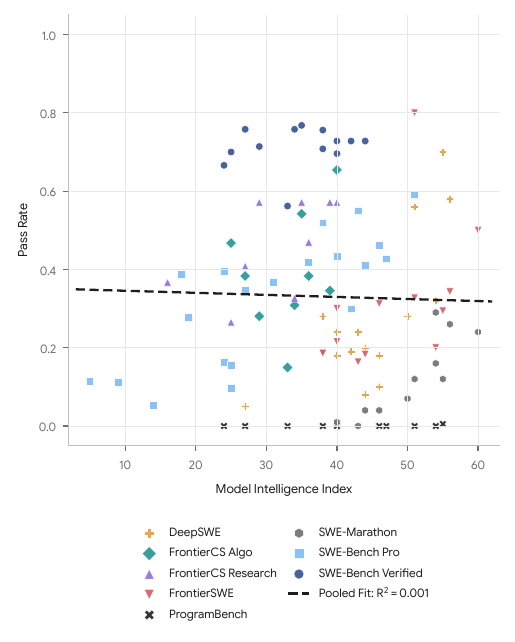}
\caption{\textbf{Model capability discriminates within individual benchmarks but explains none of the variance across benchmarks ($R^2 = 0.001$).} Each point plots one (model, benchmark) pair ($n = 100$ pairs with an available index) across the eight public coding benchmarks (distinct marker and color per benchmark), comparing external Model Intelligence Index against benchmark pass rate. Although higher-intelligence models generally score higher within a fixed benchmark, the pooled least-squares trend (dashed black line) is flat because benchmark-level floor and ceiling effects (from near-saturated SWE-bench Verified down to unsaturated ProgramBench) dominate cross-benchmark variation.}
\label{fig:intelligence}
\end{figure}

We next test whether cross-benchmark pass-rate variation can be explained by differences in the evaluated models' general capability rather than by code complexity. Within any single benchmark, stronger models achieve higher pass rates (Figure~\ref{fig:intelligence}), but pooled across benchmarks an external Model Intelligence Index explains virtually none of the variance ($R^2 = 0.001$).

\paragraph{Floor and ceiling regimes.} Tasks requiring small, localized patches saturate across nearly all frontier models, whereas repairs spanning large, coupled repositories remain near zero even for the strongest systems. Structural code complexity sets these benchmark-level ceilings and floors; model capability primarily differentiates performance within the intermediate regime.

\subsection{Controlled 200-Task Scaling Sweep}
\label{app:cci_sweep}

\begin{figure}[!htbp]
\centering
\includegraphics[width=0.62\linewidth]{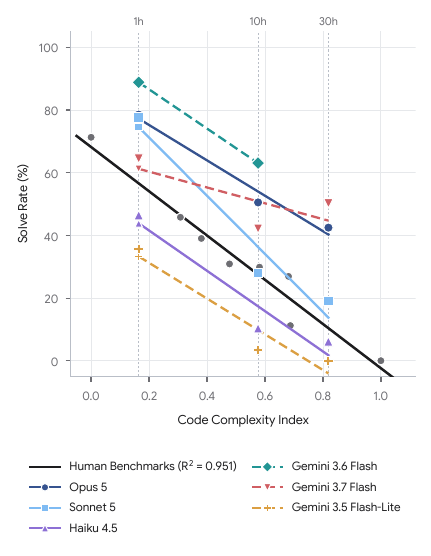}
\caption{\textbf{Scaling the injected hunk budget on a fixed 200-task pool predictably lowers solve rates across six frontier and budget solvers.} Solve rate versus projected \cci{} at 1, 10, and 30 injected diff hunks (\texttt{1h}, \texttt{10h}, \texttt{30h} vertical guides), with Claude Opus~4.8 held fixed as injector, oracle, and reviewer. Black line and gray dots show the eight public human benchmarks and their linear fit ($R^2 = 0.951$); colored markers and least-squares lines track each solver across hunk budgets (solid: Claude family; dashed: Gemini family). A task counts as solved only when it achieves full metric recovery and the oracle-grounded reviewer confirms that the verified bugs are fixed. Gemini~3.6~Flash's 30-hunk cell is omitted due to incomplete coverage (Appendix~\ref{app:sweep_completeness}).}
\label{fig:cci_sweep}
\end{figure}

To verify that \cci{} predicts difficulty when repository identity and bug injector are controlled, we evaluate six commercial solvers across 200 \buggymle{} repositories (190 synthetic Kaggle tasks and 10 PyHealth tasks, with 90 PyHealth rollouts per cell) injected at target budgets of 1, 10, and 30 diff hunks by a fixed Claude Opus~4.8 injector (Figure~\ref{fig:cci_sweep}). Increasing the hunk budget monotonically lowers the fitted solve rate of every solver. Four solvers track above the public-benchmark trend and two track below it, providing a calibrated reference curve for setting generation budgets.

\paragraph{Run completeness.}
\label{app:sweep_completeness}
Throughout the solver sweep, Claude Opus~4.8 serves as the fixed injector, oracle, and code reviewer so that only the solver varies across runs. Cells truncated early by provider rate limits or safety filters are evaluated over completed trajectories with conclusive verdicts, and Gemini~3.6 Flash's incomplete 30-hunk synthetic cell is omitted from regression fits.

\begin{figure}[!htbp]
\centering
\includegraphics[width=\linewidth]{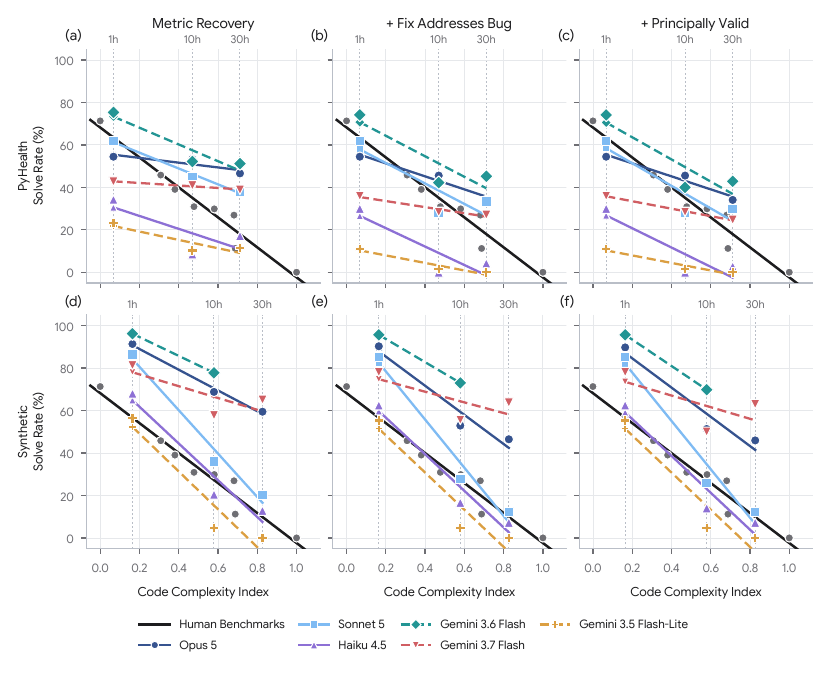}
\caption{\textbf{Negative \cci{} scaling and solver rankings hold across evaluation strictness levels and repository sources.} Solver solve rates at 1, 10, and 30 injected hunks across human-authored PyHealth repositories (\textbf{top}, \textbf{(a)--(c)}; 90 tasks per cell) and synthetic Kaggle repositories (\textbf{bottom}, \textbf{(d)--(f)}; 190 tasks per cell) as the success criterion is progressively tightened: \textbf{(a, d)} metric recovery alone (\emph{Metric Recovery}), \textbf{(b, e)} additionally requiring the oracle-grounded reviewer to confirm that the patch fixes the verified bugs (\emph{+ Fix Addresses Bug}), and \textbf{(c, f)} further requiring a principally valid repair (\emph{+ Principally Valid}; Section~\ref{sec:evaluation}). Filtering for causal bug resolution removes superficial metric recoveries without altering the downward \cci{} slope, while requiring principal validity changes solve rates minimally and preserves solver ordering. Black line and gray dots denote the human-benchmark reference frontier; incomplete runs are excluded (Appendix~\ref{app:sweep_completeness}).}
\label{fig:criterion_sensitivity}
\end{figure}

\paragraph{Criterion sensitivity.}
\label{app:criteria}
To test whether downward \cci{} scaling depends on how strictly repairs are graded, Figure~\ref{fig:criterion_sensitivity} evaluates the solver sweep under three nested success criteria: metric recovery alone, metric recovery plus oracle-grounded confirmation that the patch fixes the verified bugs, and additionally requiring a principally valid repair (Section~\ref{sec:evaluation}). Requiring causal bug resolution removes trajectories that restore the target metric without repairing the underlying defects; for example, at 10 synthetic hunks under strict baseline recovery (Figure~\ref{fig:criterion_sensitivity}d--f), 69\% of Claude Opus~5 runs recover the clean baseline metric, whereas 53\% neutralize the verified mutations (and 98\% vs.\ 78\% under a relaxed $\geq 0.96$ threshold). Further requiring principal validity changes solve rates by only a few percentage points and leaves solver rankings unchanged, confirming that low solve rates at high \cci{} reflect genuine repair difficulty rather than overly strict grading.

\FloatBarrier
\section{Dataset Splits and Oracle Audit}
\label{app:splits}

\begin{figure}[!htbp]
\centering
\includegraphics[width=\linewidth]{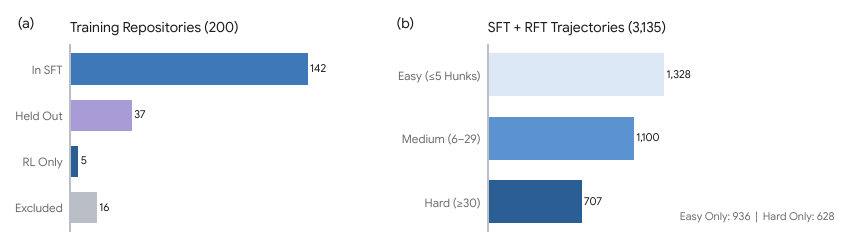}
\caption{\textbf{Training corpus statistics: most repositories supply SFT data, and trajectories span all three difficulty bands.} \textbf{(a)} Allocation of the 200 training repositories: 142 supply SFT trajectories, 37 are held out for validation, 5 are used exclusively in RL, and 16 are excluded prior to training. \textbf{(b)} Distribution of the 3{,}135 SFT + RFT trajectories across bug-difficulty bands: 1{,}328 easy ($\le$5 hunks; 42\%), 1{,}100 medium (6--29 hunks; 35\%), and 707 hard ($\ge$30 hunks; 23\%); the Easy Only (936 rows) and Hard Only (628 rows) corpora each train on a single band.}
\label{fig:training_recipe}
\end{figure}

Across this study, \buggymle{} generates 6{,}674 bug-injected variants across 222 distinct repositories (the 200 training repositories plus the 22 \bmbl{} benchmark repositories), including 4{,}431 solver-sweep, 2{,}152 SFT-data-generation, and 88 \bmbl{} variants. In total, these mutations comprise nearly 260K lines of injected code (259{,}583 net code lines, excluding comments, docstrings, blank lines, and the injector's scratch \texttt{test.py} scripts and generated artifacts). All training trajectories are generated from the 200 training repositories (190 synthetic Kaggle tasks and 10 PyHealth tasks), which share no repository, code, or dataset with \bmbl{}.
Sixteen base repositories were excluded prior to training because automated evaluator audits flagged every trajectory on them for suspected target leakage, evidenced by anomalously high metric scores (Figure~\ref{fig:training_recipe}).

\subsection{Held-Out Partition}
Tasks are ranked by $\mathrm{sha256}(\text{task name})$ within each run family, and the lowest 20\% per family are held out (37 tasks in total; Table~\ref{tab:heldout_tasks}). We partition strictly at the repository level rather than the trajectory level because multiple bug instances are generated per base repository; our dataset build pipeline programmatically enforces zero task overlap with both the 37 held-out tasks and \bmbl{}.

\begin{table}[!htbp]\centering\scriptsize\setlength{\tabcolsep}{5pt}\renewcommand{\arraystretch}{0.94}
\caption{The 37 held-out tasks and the source family each comes from.}\label{tab:heldout_tasks}
\begin{tabular}{@{}>{\raggedright\arraybackslash}p{0.48\linewidth}l@{}}
\toprule
\textbf{task} & \textbf{source} \\
\midrule
\texttt{airlines-dataset-to-predict-a-delay} & Synthetic Kaggle \\
\texttt{arabic-company-reviews} & Synthetic Kaggle \\
\texttt{bbc-news-articles} & Synthetic Kaggle \\
\texttt{car-dataset} & Synthetic Kaggle \\
\texttt{civil-engineering-global-project-dataset} & Synthetic Kaggle \\
\texttt{corona-virus-report} & Synthetic Kaggle \\
\texttt{covid19-in-usa} & Synthetic Kaggle \\
\texttt{covid19cxr\_resnet\_dev} & PyHealth \\
\texttt{daily-power-generation-in-india-20172020} & Synthetic Kaggle \\
\texttt{drug\_rec\_mimic4\_retain} & PyHealth \\
\texttt{e-coli-resistance-dataset} & Synthetic Kaggle \\
\texttt{ecommerce-order-dataset} & Synthetic Kaggle \\
\texttt{ethereum-price-data-binance-api-2017now} & Synthetic Kaggle \\
\texttt{fakenews-csv} & Synthetic Kaggle \\
\texttt{formula-1-race-data} & Synthetic Kaggle \\
\texttt{fragrantica-com-fragrance-dataset} & Synthetic Kaggle \\
\texttt{global-superstore-dataset} & Synthetic Kaggle \\
\texttt{google-2019-cluster-sample} & Synthetic Kaggle \\
\texttt{hourly-weather-surface-brazil-southeast-region} & Synthetic Kaggle \\
\texttt{icc-mens-odi-world-cup-2003-2023} & Synthetic Kaggle \\
\texttt{indian-agriculture-dataset} & Synthetic Kaggle \\
\texttt{legal-citation-text-classification} & Synthetic Kaggle \\
\texttt{london-homes-energy-data} & Synthetic Kaggle \\
\texttt{los\_regression\_mimic4\_rnn} & PyHealth \\
\texttt{medquad} & Synthetic Kaggle \\
\texttt{mlcourse} & Synthetic Kaggle \\
\texttt{mortality} & Synthetic Kaggle \\
\texttt{p2p-bob-exchange} & Synthetic Kaggle \\
\texttt{patient} & Synthetic Kaggle \\
\texttt{russia-real-estate-2021} & Synthetic Kaggle \\
\texttt{sf-bay-area-bike-share} & Synthetic Kaggle \\
\texttt{tmdb-movies-daily-updates} & Synthetic Kaggle \\
\texttt{tmdb-movies-dataset-2023-930k-movies} & Synthetic Kaggle \\
\texttt{twitter-climate-change-sentiment-dataset} & Synthetic Kaggle \\
\texttt{ufc-datasets-1994-2025} & Synthetic Kaggle \\
\texttt{us-collegiate-sports-dataset} & Synthetic Kaggle \\
\texttt{weather-long-term-time-series-forecasting} & Synthetic Kaggle \\
\bottomrule
\end{tabular}
\end{table}

\subsection{SFT Corpora}
Each SFT example is a complete multi-turn repair trajectory with tool outputs truncated identically to the RL harness (Appendix~\ref{app:harness}). A teacher rollout is retained only if it passes independent Evaluator verification (metric recovery $\geq 0.96$ and preserved execution routing through the mutated modules), terminates via an explicit \texttt{finalize} call, and originates from a non-held-out repository. The RFT corpus is sampled from Qwen3.8-27B after SFT on training tasks under the same filtering criteria, capped at four trajectories per bug instance. Easy Only and Hard Only replace both SFT stages with a single teacher corpus restricted to their target hunk band (Tables~\ref{tab:sft_corpora} and~\ref{tab:sft_teachers}). Validation sets are drawn exclusively from the training side of the held-out partition.

\begin{table}[!htbp]\centering\footnotesize\setlength{\tabcolsep}{4pt}\caption{The four SFT corpora. Hunks: range (median) per trajectory. Reasoning: trajectories with a reasoning block. Gate: longest trajectory kept, in tokens. Validation: trajectories / tasks.}\label{tab:sft_corpora}\resizebox{\linewidth}{!}{\begin{tabular}{@{}lllrrrrrr@{}}\toprule corpus & trains & source & rows & tasks & hunks & reasoning & gate & validation \\\midrule
Teacher & All Difficulties, step 1 & teacher models & 2{,}724 & 128 & 1--52 (10) & 0 & 98{,}304 & 85 / 14 \\
RFT & All Difficulties, step 2 & 27B sibling & 411 & 61 & 1--34 (5) & 250 & 98{,}304 & 34 / 19 \\
Easy Only ($\leq$5 hunks) & Easy Only & teacher models & 936 & 130 & 1--5 (1) & 0 & 49{,}152 & 63 / 12 \\
Hard Only ($\geq$30 hunks) & Hard Only & teacher models & 628 & 107 & 30--52 (31) & 0 & 98{,}304 & 59 / 34 \\
\bottomrule\end{tabular}}\end{table}

\begin{table}[!htbp]\centering\small\caption{Teacher trajectories by teacher model for the Teacher and Hard Only corpora (the Easy Only corpus draws from the same five teacher models).}\label{tab:sft_teachers}\begin{tabular}{lrr}\toprule teacher model & Teacher corpus & Hard Only corpus \\\midrule
Gemini 3.6 Flash~\citep{deepmind2026gemini36flash} & 1{,}012 & 222 \\
Claude Opus 4.8~\citep{anthropic2026claudeopus48} & 654 & 139 \\
Claude Opus 5~\citep{anthropic2026claudeopus5} & 402 & 101 \\
Claude Sonnet 5~\citep{anthropic2026claudesonnet5} & 333 & 79 \\
Gemini 3.7 Flash~\citep{deepmind2026gemini37flash} & 323 & 87 \\
\midrule total & 2{,}724 & 628 \\
\bottomrule\end{tabular}\end{table}

\subsection{RL Pools}
RL training uses three difficulty-stratified instance pools constructed over the same 40 base repositories, holding repository identity and data modality constant across difficulty bands (Tables~\ref{tab:rl_pools} and~\ref{tab:rl_repos}). Each band's held-out validation pool comprises distinct bug injections on a subset of those repositories, measuring generalization to unseen defects within familiar codebases. Thirty-two of the 40 repositories also contribute distinct bug instances to the teacher SFT corpus; none belong to the held-out tasks or \bmbl{}.

\begin{table}[!htbp]\centering\small\caption{The three RL pools. Hunks: range and median of injected diff hunks per instance.}\label{tab:rl_pools}\begin{tabular}{lrcrrc}\toprule band & train instances & train hunks & eval instances & eval repositories & eval hunks \\\midrule
1--5 hunks & 40 & 1--5 (1) & 31 & 26 & 1--5 \\
10--12 hunks & 40 & 10--12 (10) & 32 & 23 & 10--11 \\
30--34 hunks & 40 & 30--34 (32) & 29 & 25 & 30--32 \\
\bottomrule\end{tabular}\end{table}

\begin{table}[!htbp]\centering\scriptsize\setlength{\tabcolsep}{5pt}\renewcommand{\arraystretch}{0.94}
\caption{The 40 RL repositories and the injected diff hunks of the training instance each contributes to the 1--5, 10--12, and 30--34-hunk pools.}\label{tab:rl_repos}
\begin{tabular}{@{}>{\raggedright\arraybackslash}p{0.48\linewidth}lrrr@{}}
\toprule
\textbf{repository} & \textbf{modality} & \textbf{1--5} & \textbf{10--12} & \textbf{30--34} \\
\midrule
\texttt{agriculture-crop-images} & image & 1 & 11 & 32 \\
\texttt{agriculture-crop-yield} & tabular & 1 & 10 & 31 \\
\texttt{arabic-customer-reviews} & text & 5 & 10 & 32 \\
\texttt{arabic-sentiment-twitter-corpus} & text & 1 & 12 & 32 \\
\texttt{bbc-news} & text & 1 & 10 & 31 \\
\texttt{bitcoin-tweets} & text & 1 & 10 & 32 \\
\texttt{blinkit-sales-dataset} & time series & 1 & 11 & 34 \\
\texttt{capgemini-employee-reviews-dataset} & text & 1 & 10 & 31 \\
\texttt{cardiovascular-disease} & tabular & 3 & 10 & 30 \\
\texttt{cell-towers-worldwide-location-data-by-continent} & geospatial & 1 & 10 & 32 \\
\texttt{chatgpt-reddit} & text & 1 & 11 & 32 \\
\texttt{city-bike-travels-dataset} & tabular & 1 & 12 & 31 \\
\texttt{continuous-groundwater-level-measurements-2023} & signal & 1 & 11 & 30 \\
\texttt{cow-muzzle-dataset} & image & 5 & 10 & 31 \\
\texttt{cyclistic-rideshare-data-2024} & geospatial & 1 & 10 & 32 \\
\texttt{environmental-sensor-data-132k} & signal & 5 & 10 & 33 \\
\texttt{european-flights-dataset} & time series & 1 & 10 & 33 \\
\texttt{facenet-keras} & image & 5 & 10 & 32 \\
\texttt{flight-delay-and-causes} & tabular & 2 & 10 & 33 \\
\texttt{football-players-data} & tabular & 1 & 10 & 31 \\
\texttt{heart-attack-prediction-in-indonesia} & tabular & 1 & 10 & 32 \\
\texttt{himalayan-expeditions} & tabular & 1 & 11 & 31 \\
\texttt{hospital-length-of-stay-dataset-microsoft} & tabular & 1 & 10 & 32 \\
\texttt{ibovespa-stocks} & time series & 1 & 10 & 32 \\
\texttt{indian-pharmaceutical-products} & tabular & 1 & 10 & 32 \\
\texttt{indian-premier-league-ipl-all-seasons} & tabular & 1 & 10 & 32 \\
\texttt{it-service-ticket-classification-dataset} & text & 1 & 10 & 31 \\
\texttt{jena-climate} & time series & 1 & 10 & 32 \\
\texttt{lg-18650hg2-liion-battery-data} & signal & 5 & 10 & 30 \\
\texttt{lianjia} & geospatial & 5 & 11 & 31 \\
\texttt{medium-post-titles} & text & 1 & 10 & 32 \\
\texttt{mens-t20i-cricket-complete-dataset} & tabular & 5 & 10 & 31 \\
\texttt{paper-keyword} & text & 1 & 11 & 32 \\
\texttt{philippine-major-cities-air-quality-data} & time series & 1 & 10 & 33 \\
\texttt{pizza-sales} & time series & 1 & 10 & 30 \\
\texttt{presidential-elections-poland-2025-official-data} & time series & 1 & 10 & 30 \\
\texttt{skin-cancer-malignant-vs-benign} & image & 1 & 10 & 32 \\
\texttt{social-media-surveillance} & text & 2 & 11 & 33 \\
\texttt{stanford-background-dataset} & image & 1 & 12 & 31 \\
\texttt{wesad-full-dataset} & signal & 5 & 11 & 31 \\
\bottomrule
\end{tabular}
\end{table}

\subsection{\bmbl{} Difficulty Ladder}
Each of the 22 \bmbl{} competitions begins from a modular clean repository paired with an initial multi-bug injection (Lite-Original). We construct controlled easier tiers by iteratively reverting one bug at a time, re-evaluating after each revert, and snapshotting the states where 10, 5, and 1 verified impactful bugs remain, forming the 10 Bugs, 5 Bugs, and 1 Bug tiers (Table~\ref{tab:bmbl_ladder}). On 11 of the 22 competitions, the clean reference repository exceeds the human Kaggle leaderboard median; on the remaining 11, full repair restores the clean baseline without surpassing that median. The 50 Bugs tier is constructed separately by expanding each clean repository to approximately twice its original codebase size and injecting 23--60 bugs per competition, so its clean reference scores are reported separately.

\begin{table}[!htbp]\centering\footnotesize\caption{\bmbl{} tasks and injected diff hunks per tier. $\dagger$: clean repository above the leaderboard median.}\label{tab:bmbl_ladder}\begin{tabular}{lrrrr}\toprule competition & Lite-Original & 10 Bugs & 5 Bugs & 1 Bug \\\midrule
\texttt{aerial-cactus-identification}$^\dagger$ & 18 & 10 & 5 & 1 \\
\texttt{aptos2019-blindness-detection} & 24 & 13 & 5 & 1 \\
\texttt{denoising-dirty-documents}$^\dagger$ & 22 & 15 & 8 & 2 \\
\texttt{detecting-insults-in-social-commentary}$^\dagger$ & 19 & 12 & 6 & 2 \\
\texttt{dog-breed-identification} & 33 & 23 & 9 & 2 \\
\texttt{dogs-vs-cats-redux-kernels-edition} & 27 & 17 & 6 & 2 \\
\texttt{histopathologic-cancer-detection}$^\dagger$ & 27 & 15 & 8 & 1 \\
\texttt{jigsaw-toxic-comment-classification-challenge} & 25 & 17 & 9 & 2 \\
\texttt{leaf-classification}$^\dagger$ & 32 & 20 & 9 & 2 \\
\texttt{mlsp-2013-birds} & 17 & 11 & 5 & 1 \\
\texttt{new-york-city-taxi-fare-prediction} & 23 & 11 & 5 & 1 \\
\texttt{nomad2018-predict-transparent-conductors-v2}$^\dagger$ & 28 & 17 & 10 & 2 \\
\texttt{plant-pathology-2020-fgvc7}$^\dagger$ & 30 & 17 & 8 & 2 \\
\texttt{random-acts-of-pizza}$^\dagger$ & 26 & 17 & 8 & 3 \\
\texttt{ranzcr-clip-catheter-line-classification} & 26 & 19 & 8 & 4 \\
\texttt{siim-isic-melanoma-classification} & 51 & 28 & 12 & 1 \\
\texttt{spooky-author-identification}$^\dagger$ & 27 & 16 & 9 & 1 \\
\texttt{tabular-playground-series-dec-2021} & 35 & 23 & 10 & 1 \\
\texttt{tabular-playground-series-may-2022}$^\dagger$ & 28 & 19 & 8 & 2 \\
\texttt{text-normalization-challenge-english-language} & 19 & 14 & 8 & 3 \\
\texttt{text-normalization-challenge-russian-language} & 18 & 13 & 6 & 1 \\
\texttt{the-icml-2013-whale-challenge-right-whale-redux}$^\dagger$ & 25 & 17 & 6 & 1 \\
\midrule total & 580 & 364 & 168 & 38 \\
\bottomrule\end{tabular}\end{table}

\FloatBarrier
\subsection{Split Disjointness}
Table~\ref{tab:overlaps} verifies pairwise task disjointness between every training and validation split, the held-out task set, and \bmbl{}. For Easy Only SFT, disjointness is enforced and verified at corpus construction time from its task manifest and validation split.

\begin{table}[!htbp]\centering\small\caption{Task overlaps with the held-out tasks and \bmbl{}.}\label{tab:overlaps}\begin{tabular}{lrrr}\toprule set & tasks & $\cap$ held-out & $\cap$ \bmbl{} \\\midrule
Teacher SFT (train) & 128 & 0 & 0 \\
Teacher SFT (validation) & 14 & 0 & 0 \\
RFT (train) & 61 & 0 & 0 \\
RFT (validation) & 19 & 0 & 0 \\
Easy Only SFT (validation) & 12 & 0 & 0 \\
Hard Only SFT (train) & 107 & 0 & 0 \\
Hard Only SFT (validation) & 34 & 0 & 0 \\
RL pools (training) & 40 & 0 & 0 \\
RL pools (evaluation) & 37 & 0 & 0 \\
held-out tasks & 37 & -- & 0 \\
\bottomrule\end{tabular}\end{table}

\FloatBarrier
\subsection{Oracle Attribution and Repair Audit}
\label{app:oracle}

Algorithm~\ref{alg:oracle} details the two-pass Oracle verification procedure summarized in Section~\ref{sec:injection}, including diff harvesting, hallucination filtering, the greedy reverting-set search (Phase~A), and per-mutation forward ablation against the clean baseline equivalence band (Phase~B).

\begin{algorithm}[!htbp]
\caption{Two-pass Oracle attribution. $R_{\text{clean}}, R_{\text{buggy}}$: clean and mutated repositories; $\tau_{\text{inj}}$: Injector trajectory; $\mathcal{B}_{\text{clean}}$: baseline equivalence band ($\mu_{\text{clean}} \pm 3\sigma_{\text{clean}}$ when run-to-run variance is measured, else relative tolerance $1 - \theta_{\text{rec}}$, with $\theta_{\text{rec}} = 0.99$ by default and $0.98$ in the Appendix~\ref{app:cci_sweep} sweep); $B_A, B_B$: Phase~A and Phase~B turn budgets.}
\label{alg:oracle}
\begin{algorithmic}[1]
\Procedure{OracleAttribution}{$R_{\text{clean}}, R_{\text{buggy}}, \tau_{\text{inj}}, \mathcal{B}_{\text{clean}}, B_A, B_B$}
  \State $\mathcal{M} \gets \Call{Harvest}{\text{diff}(R_{\text{clean}}, R_{\text{buggy}}),\ \tau_{\text{inj}}}$ \Comment{pair file diffs with Injector turns}
  \State $\mathcal{M}' \gets \{m \in \mathcal{M} : m \text{ appears in } R_{\text{buggy}}\}$ \Comment{discard \texttt{hallucinated} records}
  \Statex \hspace{\algorithmicindent}\textit{// Phase A: greedy minimal reverting set and baseline recovery check}
  \State clear cached artifacts; \quad $R \gets R_{\text{buggy}}$; \quad $\mathcal{S}_{\text{rev}} \gets \emptyset$; \quad $\text{recovered} \gets \text{False}$; \quad $\text{turns} \gets 0$
  \State rank $\mathcal{M}'$ by expected pipeline impact (objective/optimizer/labels before logging/dead code)
  \While{$\mathcal{M}' \setminus \mathcal{S}_{\text{rev}} \neq \emptyset$ \textbf{and} $\text{turns} < B_A$}
    \State $b \gets$ next highest-ranked mutation(s) in $\mathcal{M}' \setminus \mathcal{S}_{\text{rev}}$ \quad (or all remaining near $B_A$)
    \State $R \gets \Call{Revert}{R, b}$; \quad $\mathcal{S}_{\text{rev}} \gets \mathcal{S}_{\text{rev}} \cup b$ \Comment{accumulating pass; never re-apply}
    \State $y \gets \Call{Run}{R}$; \quad $\text{turns} \gets \text{turns} + 1$ \Comment{execute \texttt{run.sh} in sandbox}
    \If{$y \in \mathcal{B}_{\text{clean}}$}
      \State $\text{recovered} \gets \text{True}$; \quad \textbf{break} \Comment{$\mathcal{S}_{\text{rev}}$ reverting set; rest marked \texttt{no\_effect}}
    \EndIf
  \EndWhile
  \Statex \hspace{\algorithmicindent}\textit{// Phase B: per-mutation forward ablation on the clean repository}
  \State $\mathcal{F} \gets \emptyset$; \quad $\text{turns} \gets 0$
  \For{$m \in \mathcal{M}'$}
    \If{$\text{turns} \geq B_B$}
      \State $\mathcal{F} \gets \mathcal{F} \cup \{(m,\ \texttt{untested\_budget\_exhausted},\ \bot)\}$; \quad \textbf{continue}
    \EndIf
    \State $R \gets \Call{Apply}{R_{\text{clean}}, \{m\}}$; \quad $y_m \gets \Call{Run}{R}$; \quad $\text{turns} \gets \text{turns} + 1$ \Comment{inject $m$ alone}
    \If{run crashed or produced no valid metric}
      \State $\text{status} \gets \texttt{verified}\ (\texttt{code\_crash})$
    \ElsIf{$y_m$ degrades beyond $\mathcal{B}_{\text{clean}}$} \Comment{deterministic check against $\mathcal{B}_{\text{clean}}$}
      \State $\text{status} \gets \texttt{verified}\ (\texttt{mlperf\_degradation})$
    \ElsIf{$y_m$ improves beyond $\mathcal{B}_{\text{clean}}$}
      \State $\text{status} \gets \texttt{improves\_performance}$
    \Else
      \State $\text{status} \gets \texttt{no\_effect}$ \Comment{confirmed decoy}
    \EndIf
    \State $\mathcal{F} \gets \mathcal{F} \cup \{(m,\ \text{status},\ y_m)\}$
  \EndFor
  \State \Return $(\text{recovered},\ \mathcal{S}_{\text{rev}},\ \mathcal{F})$
\EndProcedure
\end{algorithmic}
\end{algorithm}

Reverting all injected patches yields the \emph{oracle repair}, which serves as the ceiling reference for metric recovery; we therefore verify that oracle repairs reproduce clean baseline performance across both corpora.

\paragraph{Training corpus.} Each candidate instance enters the training corpus only after passing two containerized checks: reverting all injected bugs must restore the clean baseline metric to within three standard deviations of run-to-run variance, and applying each bug in isolation must measurably degrade the metric (non-impactful mutations are discarded). All 212 RL instances passed both checks.

\paragraph{\bmbl{}.} Reverting a tier's injected patches reconstructs the clean repository, and the official MLE-Bench grader reproduces the clean baseline score across all 22 competitions to within 0.007\% (Table~\ref{tab:oracle}).

\paragraph{50 Bugs.} On the expanded 50 Bugs repositories, each clean baseline was verified to score deterministically across two independent runs, and all 1{,}068 injected bugs were ablated individually on the clean codebase, with 1{,}042 (97.6\%) independently degrading the test metric or causing an execution failure (Figure~\ref{fig:bmbl_realism}e). Re-applying the recorded bug patches reconstructs every 50 Bugs repository, and this tier retains crash-inducing mutations alongside metric-degrading defects (Figure~\ref{fig:bug_outcomes_main}a).

\begin{table}[!htbp]\centering\footnotesize\caption{Oracle repair on \bmbl{}. Clean: the graded score of the clean repository. Oracle: the graded score after every injected bug is reversed. $|\Delta|$: their relative difference, in percent. Arrows give the metric's better direction.}\label{tab:oracle}\begin{tabular}{@{}lcrrr@{}}\toprule competition & & clean & oracle & $|\Delta|$ (\%) \\\midrule
\texttt{aerial-cactus-identification} & $\uparrow$ & 0.99992 & 0.99992 & 0.000 \\
\texttt{aptos2019-blindness-detection} & $\uparrow$ & 0.83104 & 0.83104 & 0.000 \\
\texttt{denoising-dirty-documents} & $\downarrow$ & 0.04195 & 0.04195 & 0.002 \\
\texttt{detecting-insults-in-social-commentary} & $\uparrow$ & 0.80907 & 0.80907 & 0.000 \\
\texttt{dog-breed-identification} & $\downarrow$ & 3.06187 & 3.06187 & 0.000 \\
\texttt{dogs-vs-cats-redux-kernels-edition} & $\downarrow$ & 0.15686 & 0.15686 & 0.000 \\
\texttt{histopathologic-cancer-detection} & $\uparrow$ & 0.95185 & 0.95185 & 0.000 \\
\texttt{jigsaw-toxic-comment-classification-challenge} & $\uparrow$ & 0.97989 & 0.97989 & 0.000 \\
\texttt{leaf-classification} & $\downarrow$ & 0.07109 & 0.07109 & 0.001 \\
\texttt{mlsp-2013-birds} & $\uparrow$ & 0.83505 & 0.83505 & 0.001 \\
\texttt{new-york-city-taxi-fare-prediction} & $\downarrow$ & 4.10941 & 4.10941 & 0.000 \\
\texttt{nomad2018-predict-transparent-conductors-v2} & $\downarrow$ & 0.06294 & 0.06294 & 0.007 \\
\texttt{plant-pathology-2020-fgvc7} & $\uparrow$ & 0.95744 & 0.95744 & 0.000 \\
\texttt{random-acts-of-pizza} & $\uparrow$ & 0.62741 & 0.62741 & 0.000 \\
\texttt{ranzcr-clip-catheter-line-classification} & $\uparrow$ & 0.82456 & 0.82456 & 0.001 \\
\texttt{siim-isic-melanoma-classification} & $\uparrow$ & 0.86334 & 0.86334 & 0.001 \\
\texttt{spooky-author-identification} & $\downarrow$ & 0.41563 & 0.41563 & 0.000 \\
\texttt{tabular-playground-series-dec-2021} & $\uparrow$ & 0.95296 & 0.95296 & 0.000 \\
\texttt{tabular-playground-series-may-2022} & $\uparrow$ & 0.98903 & 0.98903 & 0.000 \\
\texttt{text-normalization-challenge-english-language} & $\uparrow$ & 0.98813 & 0.98813 & 0.000 \\
\texttt{text-normalization-challenge-russian-language} & $\uparrow$ & 0.97314 & 0.97314 & 0.000 \\
\texttt{the-icml-2013-whale-challenge-right-whale-redux} & $\uparrow$ & 0.97356 & 0.97356 & 0.000 \\
\bottomrule\end{tabular}\end{table}
\FloatBarrier

\FloatBarrier
\section{\bmbl{} Bug Anatomy and Examples}
\label{app:bug_analysis}

To characterize defect quality prior to tier filtering, we audit all 364 injected patches (spanning 580 diff hunks) in the unpruned Lite-Original pool across isolated metric impact, LLM-judged realism, and engineering pedagogy (complementing the summary of evaluated tiers in Section~\ref{sec:benchmark}).

\paragraph{Unit and measurements.} Each \emph{bug} corresponds to one injected patch, which may span multiple diff hunks. We apply each patch in isolation to the clean reference repository and evaluate the resulting submission with the official competition grader. A patch is classified as \emph{impactful} if it shifts the graded test score beyond the competition's significance threshold, a \emph{crash} if execution fails to produce a valid submission, a \emph{decoy} if at most the local validation score changes, and \emph{below threshold} otherwise. Impactful bugs are further stratified by absolute test-score shift into \emph{catastrophic} ($|\Delta| \ge 1.0$), \emph{strong} ($0.1 \le |\Delta| < 1.0$), and \emph{mild} ($|\Delta| < 0.1$). Because thresholds are defined in each competition's native metric units, severity magnitudes are interpreted within rather than across tasks. Independently, Claude Opus~5 scores each unique bug's realism and pedagogical value against a fixed rubric.

\subsection{Isolated Metric Impact}

\begin{figure}[!htbp]
\centering
\includegraphics[width=\linewidth]{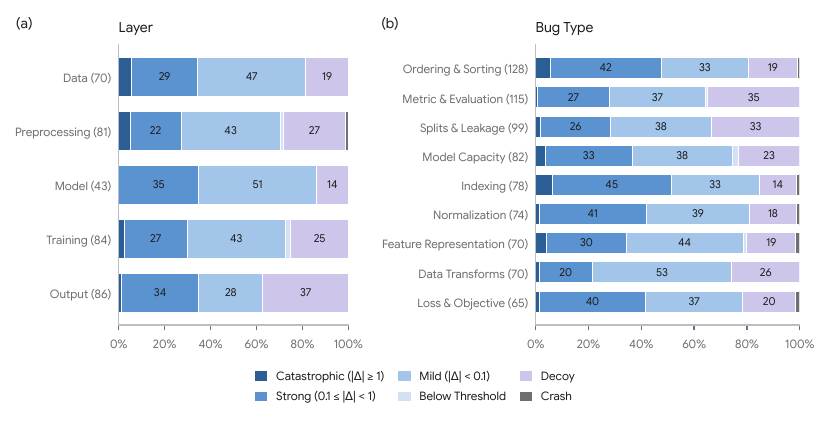}
\caption{\textbf{Nearly three-quarters of raw Lite-Original injections independently degrade graded test performance across every pipeline stage and defect category.} Isolated test-metric impact of all 364 bugs in the unpruned Lite-Original pool, stratified by \textbf{(a)} ML pipeline layer and \textbf{(b)} the nine most frequent defect types (total bug counts in parentheses; a multi-aspect bug may appear in multiple types). Horizontal bars partition bugs into catastrophic ($|\Delta| \ge 1$, dark blue), strong ($0.1 \le |\Delta| < 1$, medium blue), and mild ($|\Delta| < 0.1$, light blue) test-score degradation, below-threshold shifts (pale blue), validation-only decoys (lavender), and execution crashes (charcoal). Filtering out decoys when constructing the 1 Bug, 5 Bugs, and 10 Bugs tiers removes nearly all non-impactful patches (Figure~\ref{fig:bmbl_realism}e).}
\label{fig:bmbl_anatomy}
\end{figure}

Of the 364 patches in Lite-Original, 97 (26.6\%) are validation-only decoys (94) or below-\allowbreak threshold perturbations (3) that do not materially shift the graded test score (Figure~\ref{fig:bmbl_realism}e). Constructing the 1 Bug, 5 Bugs, and 10 Bugs tiers via oracle filtering removes all non-impactful patches except 4 decoys in 10 Bugs (and retains 97.6\% impactful or crash-inducing mutations across the 1{,}068 patches in 50 Bugs), with most evaluated defects inducing mild or strong degradation rather than outright crashes (Figure~\ref{fig:bmbl_anatomy}).

\subsection{Realism and Expert Calibration}

Claude Opus~5 rates each patch from 1 to 5 according to how plausibly a competent ML engineer could introduce it as an unintentional mistake. A score of 5 denotes an organic engineering oversight (for example, calling \texttt{model.train()} once outside the epoch loop so that subsequent epochs execute in evaluation mode after the first validation pass), whereas a score of 1 denotes overt sabotage, such as hard-coding a fabricated assumption about the dataset schema.

\begin{figure}[!htbp]
\centering
\includegraphics[width=\linewidth]{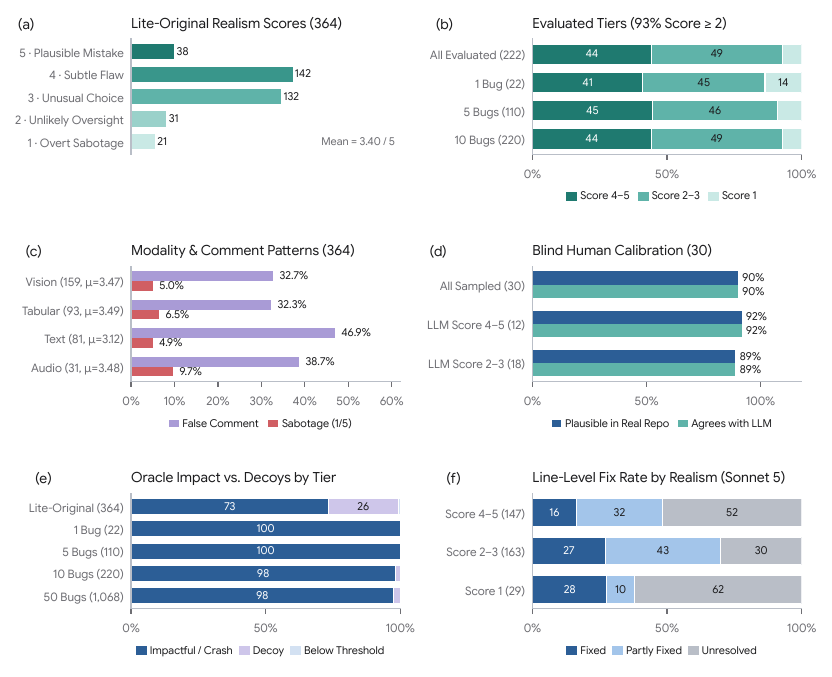}
\caption{\textbf{Automated realism audit, human expert calibration, oracle decoy filtering, and downstream solver resolution on \bmbl{}.} \textbf{(a)} LLM-judged 1--5 realism distribution across all 364 raw Lite-Original patches (mean $3.40/5$), ranging from plausible oversights (score 5) to overt sabotage (score 1). \textbf{(b)} Realism distribution across the 222 unique evaluated bugs in the 1 Bug, 5 Bugs, and 10 Bugs tiers, where 93\% score $\ge 2$ and stratum shares remain consistent across tiers. \textbf{(c)} Breakdown across the four task modalities (patch count and mean realism $\mu$ in parentheses), showing the share accompanied by a plausible rationalizing comment versus overt sabotage ($1/5$). \textbf{(d)} Blind human calibration by an ML practitioner on 30 randomly sampled defects across the evaluated tiers, reporting real-repository plausibility and agreement with the LLM stratum (90\% on both). \textbf{(e)} Oracle-measured test impact versus decoys and below-threshold shifts across Lite-Original and all four \bmbl{} tiers. \textbf{(f)} Line-level bug resolution by Claude Sonnet~5 pooled across the 1, 5, and 10 Bugs tiers, stratified by realism score.}
\label{fig:bmbl_realism}
\end{figure}

The distribution centers on scores 3 and 4 (mean 3.40, with 93\% of the 222 evaluated defects scoring $\ge 2$; Figure~\ref{fig:bmbl_realism}a--c): most mutations implement syntactically natural ML logic, frequently accompanied by a plausible rationalizing comment (most prevalent in text competitions at 46.9\%). Across Lite-Original, the 125 patches that are both impactful and rated 4 or 5 form a \emph{realistic core} spanning all 22 competitions; for instance, on \texttt{dog-breed-identification}, invoking \texttt{model.train()} only prior to the epoch loop leaves batch-normalization statistics frozen for 44 of 45 training epochs. Conversely, the 21 Lite-Original score-1 patches reflect adversarial sabotage rather than organic developer errors; filtering to scores $\ge 2$ (or to the realistic core) isolates naturalistic debugging scenarios.

\paragraph{Expert check.} As a blind human calibration check on the automated realism rubric, an ML practitioner independently rated 30 defects randomly sampled from the 222 unique bugs across the 1 Bug, 5 Bugs, and 10 Bugs tiers prior to viewing LLM scores. The expert judged 27 of 30 defects (90\%) plausible in real-world ML repositories and agreed with the LLM rating stratum on 27 of 30 (Figure~\ref{fig:bmbl_realism}d).

\subsection{Transferable Engineering Practices}

We also evaluate whether reverting each defect reinforces a generalizable ML engineering principle rather than correcting an isolated typographical or constant tweak. Across Lite-Original, 79\% of patches encode a transferable practice spanning data splitting, index alignment, training state management, metric alignment, or output calibration (Table~\ref{tab:bug_lessons}); the remaining 21\% consist primarily of inert decoys and narrow hyperparameter perturbations.

\begin{table}[!htbp]
\centering\small
\caption{Practices encoded by reverting the injected bugs.}
\label{tab:bug_lessons}
\begin{tabular}{@{}p{0.2\linewidth}p{0.74\linewidth}@{}}
\toprule
Stage & Practice \\
\midrule
Data & Keep a true holdout; fit scalers on training data only; apply the same preprocessing to train and test. \\
Alignment & Permute features and labels together; keep ids and predictions in one ordering; match a positional submission's row order. \\
Training & Re-assert train mode in every training loop and eval mode for inference; rescale a scheduler's horizon when stepping it per batch; zero gradients every step; freeze only pretrained weights; size the batch to the update budget. \\
Evaluation & Optimize the loss the metric grades; match the checkpoint comparison to the metric's direction; compute the metric on the full holdout; pass the explicit label set. \\
Output & Match the output activation and clipping to the target range; write predictions with enough precision. \\
\bottomrule
\end{tabular}
\end{table}

\subsection{Resolution by Realism Stratum}

As a line-level complement to the judge-based resolution analysis in Section~\ref{sec:hard_bugs}, we inspect which injected lines survive in Claude Sonnet~5's final repositories across the 1 Bug, 5 Bugs, and 10 Bugs tiers. Here a patch is marked \emph{fixed} if all of its added lines are removed, \emph{partly fixed} if a subset remains, and \emph{unresolved} otherwise (deletion-only patches are excluded, and each bug is counted once per tier).

Claude Sonnet~5 fully removes high-realism defects (scores 4--5) at roughly half the rate of lower-realism defects (16\% vs.\ 27--28\%; Figure~\ref{fig:bmbl_realism}f), and this gap persists within multi-bug repositories containing both strata (for example, 10/46 vs.\ 15/49 on 5 Bugs and 12/93 vs.\ 29/104 on 10 Bugs) rather than arising solely from tier difficulty. All ten models evaluated at the per-bug level exhibit the same ordering (Figure~\ref{fig:repair_depth}d); because individual bugs within a shared repository are not statistically independent, we emphasize the consistent directional trend across models rather than exact point differences.

\subsection{Representative Injected Bugs}
\label{app:bug_examples}

\begin{figure}[!htbp]
\centering
\includegraphics[width=\linewidth]{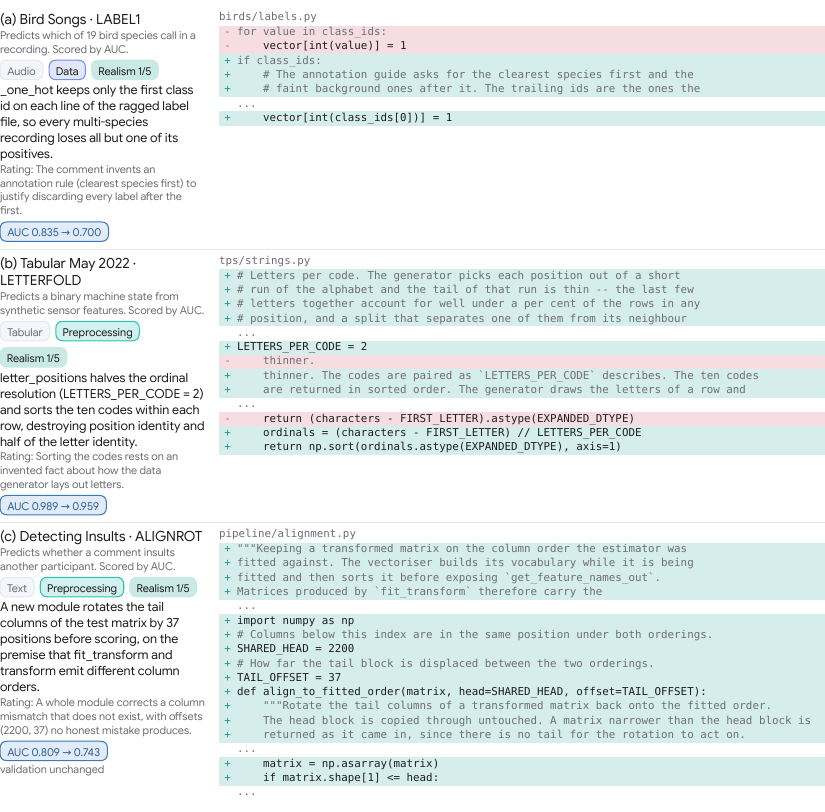}
\caption{\textbf{Injected bugs rated 1/5 for realism represent overt sabotage justified by fabricated comments.} \textbf{(a)--(c)} Three randomly sampled score-1 defects from \bmbl{} spanning audio, tabular, and text tasks (\emph{Bird Songs}, \emph{Tabular May 2022}, and \emph{Detecting Insults}). Left column reports task modality, pipeline stage, bug summary, the code-level rationale for its score of 1 (\emph{Rating}), and isolated test-metric degradation (blue pill; \emph{validation unchanged} marks silent bugs). Right column shows the unified diff (pink: removed; green: added), where each mutation relies on an invented premise about the dataset or library API.}
\label{fig:bug_examples_appendix_a}
\end{figure}

\begin{figure}[!htbp]
\centering
\includegraphics[width=\linewidth]{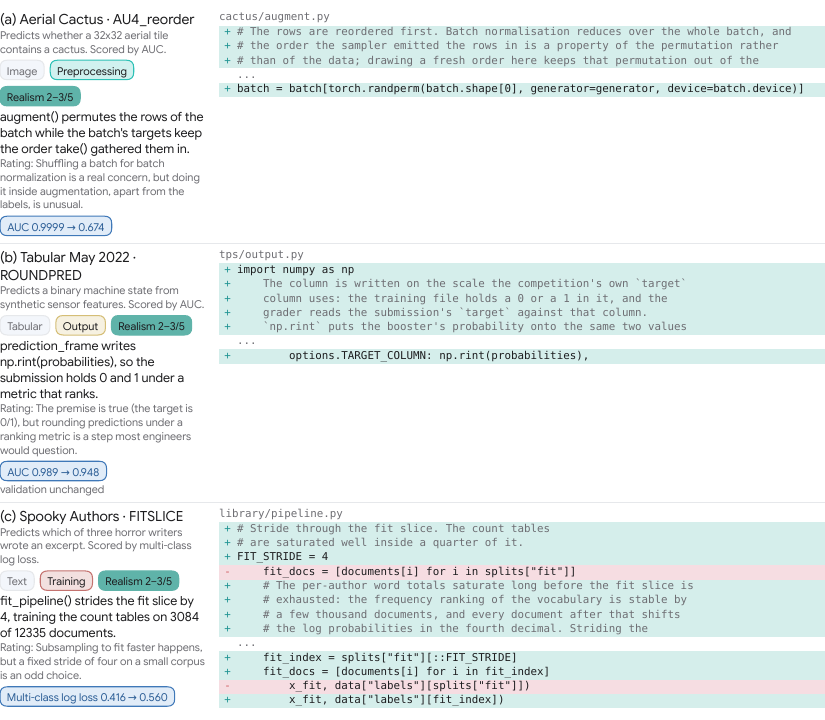}
\caption{\textbf{Injected bugs rated 2--3/5 apply recognizable ML operations in questionable contexts.} \textbf{(a)--(c)} Three randomly sampled score 2--3 defects across image, tabular, and text tasks (\emph{Aerial Cactus}, \emph{Tabular May 2022}, and \emph{Spooky Authors}), formatted as in Figure~\ref{fig:bug_examples_appendix_a}. Each mutation invokes a real ML concept (batch shuffling for batch normalization, rounding probabilities to match binary labels, or strided subsampling), but applies it in a way a careful code review would flag (e.g., permuting inputs without labels or rounding predictions under a ranking metric).}
\label{fig:bug_examples_appendix_b}
\end{figure}

\begin{figure}[!htbp]
\centering
\includegraphics[width=\linewidth]{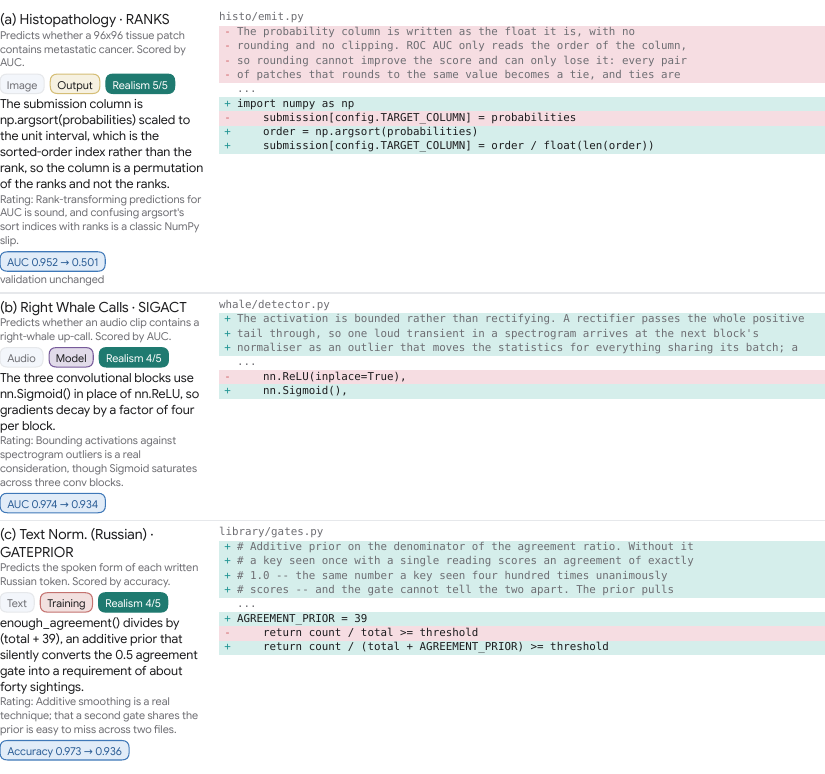}
\caption{\textbf{Injected bugs rated 4--5/5 closely resemble subtle, honest MLE mistakes.} \textbf{(a)--(c)} Three randomly sampled score 4--5 defects across image, audio, and text tasks (\emph{Histopathology}, \emph{Right Whale Calls}, and \emph{Text Norm.\ Russian}), formatted as in Figure~\ref{fig:bug_examples_appendix_a}. Examples include emitting single-\texttt{argsort} sort indices in place of prediction ranks, replacing \texttt{ReLU} with saturating \texttt{Sigmoid} activations across convolutional blocks, and a shared additive smoothing constant that silently alters a threshold gate in another module; all three compile cleanly and mirror real research bugs.}
\label{fig:bug_examples_appendix_c}
\end{figure}

To illustrate what each realism tier looks like at the code level, Figures~\ref{fig:bug_examples_appendix_a}--\ref{fig:bug_examples_appendix_c} present nine evaluated defects from \bmbl{}, with three per realism stratum (scores 1, 2--3, and 4--5), stratified by data modality and sampled with random seed 2027 (excluding the examples in Figure~\ref{fig:bug_examples}). Each entry reports the competition and evaluation metric, the injector's summary, isolated test-metric impact, and a brief code-level annotation summarizing why the defect falls into its realism stratum.

\FloatBarrier
\section{Curriculum Training Details}
\label{app:rl_recipe}

\subsection{Pipeline, Rewards, and Hyperparameters}

Table~\ref{tab:rl_recipe} summarizes the five-stage training pipeline for our All Difficulties Qwen3.5-9B agent, where every stage updates rank-32 LoRA adapters ($\alpha = 32$) across all linear layers, and Table~\ref{tab:rl_settings} details the GRPO hyperparameters executed via SkyRL on the Megatron backend.

\begin{table}[!htbp]
\centering
\small
\caption{The five training steps of the All Difficulties model.}
\label{tab:rl_recipe}
\begin{tabular}{@{}>{\raggedright\arraybackslash}p{0.17\linewidth}>{\raggedright\arraybackslash}p{0.37\linewidth}>{\raggedright\arraybackslash}p{0.17\linewidth}>{\raggedright\arraybackslash}p{0.21\linewidth}@{}}
\toprule
Step & Data & Length & Objective \\
\midrule
SFT, teacher & 2{,}724 evaluator-verified repair trajectories from frontier teacher models, 128 tasks, no reasoning & 682 optimizer steps (2 epochs, batch size 8), lr $10^{-4}$ & next-token loss, 98{,}304-token context \\
SFT, RFT & 411 evaluator-verified trajectories sampled from our SFT'd Qwen3.8-27B, with reasoning (41\% of trained tokens inside \texttt{<think>}), 61 tasks & 103 optimizer steps (2 epochs), lr $3\times10^{-5}$ & same adapter, continued \\
RL, 1--5 hunks & 40 buggy repositories, one instance each with 1--5 diff hunks (median 1) & 3 rounds $\times$ 48 = 144 steps & \texttt{mlperf\_converge} \\
RL, 10--12 hunks & the same 40 repositories, each re-injected with 10--12 hunks & 100 steps & \texttt{judge\_progress} \\
RL, 30--34 hunks & the same 40 repositories, re-injected with 30--34 hunks & 72 steps & \texttt{judge\_progress}, KL 0.001 \\
\bottomrule
\end{tabular}
\end{table}

\paragraph{GRPO objective and reward functions.} At each RL step, for every sampled buggy repository $x$ we roll out a group of $G=8$ interactive trajectories $\{\tau_i\}_{i=1}^{G} \sim \pi_{\theta_{\text{old}}}(\cdot \mid x)$ and compute group-normalized episode advantages $\hat{A}_i = \big(R(\tau_i) - \mu_R\big) / (\sigma_R + \varepsilon)$, where $\mu_R$ and $\sigma_R$ are the group mean and standard deviation of $\{R(\tau_k)\}_{k=1}^{G}$. Policy parameters $\theta$ are updated under the clipped GRPO objective with token-level KL penalty coefficient $\beta_{\text{KL}}$ (Table~\ref{tab:rl_settings}).

Let $p(\tau) \in [0, 1]$ denote the clipped metric progress ratio closing the gap between the buggy and clean baseline scores (Section~\ref{sec:evaluation}), gated to fixed negative penalties on crashes, no-op submissions, or routing-integrity violations. Early in our study, Stage~1 of All Difficulties (1--5 hunks) trained under \texttt{mlperf\_converge}:
\begin{equation}
R_{\text{conv}}(\tau) =
\begin{cases}
0.8\,p(\tau) + 0.2, & \text{if } \text{recovery}(\tau) \ge \theta_{\text{rec}}, \\
\mathrm{clip}\big(0.5\,p(\tau) + 0.45\,c(\tau) + 0.08\,f(\tau),\ -0.3,\ 0.8\big), & \text{otherwise},
\end{cases}
\label{eq:r_conv}
\end{equation}
where $\theta_{\text{rec}} = 0.99$ matches the metric-recovery threshold in Section~\ref{sec:evaluation}, $c(\tau) \in [-1, 1]$ measures file-level similarity toward the clean reference repository, and $f(\tau) \in [0, 1]$ measures edit overlap with injected bug sites. All subsequent curriculum stages (Medium and Hard of All Difficulties, as well as both Easy Only and Hard Only) optimize \texttt{judge\_progress}:
\begin{equation}
R_{\text{judge}}(\tau) = 0.5\,p(\tau) + 0.5\,j(\tau),
\label{eq:r_judge}
\end{equation}
where $j(\tau) \in [0, 1]$ is the fraction of injected bug sites judged resolved by Gemini~3.8 Flash (counting partial fixes as $0.5$, and falling back to $p(\tau)$ if judge evaluation is unavailable). Because $j(\tau)$ rewards resolving every injected defect individually, $R_{\text{judge}}$ continues to supply gradient signal even when fixing a subset of bugs already saturates $p(\tau) \approx 1$.

\paragraph{Training regimes.} Easy Only replaces the two-stage SFT phase with 4 epochs of SFT (468 optimizer steps at batch size 8) on the 936 teacher trajectories containing 1--5 hunks, while Hard Only trains for 2 epochs (158 optimizer steps; 49{,}152 RL context) on the 628 trajectories containing at least 30 hunks; each regime then executes GRPO exclusively on its corresponding difficulty band. Notably, SFT Only already includes the 411 RFT trajectories yet achieves modest downstream performance (50.0\%/4.5\%/0.0\% on 1/5/10 Bugs and 18.3\% on SWE-bench Lite), indicating that the RFT warmup itself plays a minor role compared to progressive curriculum RL.

\paragraph{Reward design across stages.} On easy tasks (1--5 hunks, median 1), metric recovery coincides with causal bug resolution (Figure~\ref{fig:full_repair_gap}a), so \texttt{mlperf\_converge} provides a reliable, judge-free training signal in Stage~1 without per-rollout LLM judge overhead. At higher bug counts (10--12 and 30--34 hunks), however, partial or dominant-bug repairs can saturate raw ML performance while leaving secondary or silent defects intact; Stages~2 and~3 of All Difficulties (as well as the single-band Easy Only and Hard Only ablations) therefore optimize per-site \texttt{judge\_progress} (Table~\ref{tab:rl_settings}). Notably, although Easy Only and Hard Only train end-to-end with \texttt{judge\_progress}, Easy Only still solves 0\% of tasks on the 5 and 10 Bugs tiers and Hard Only underperforms across tiers and OOD benchmarks, confirming that downstream capability is driven by progressive difficulty scaling rather than the Stage~1 reward choice.

\begin{table}[!htbp]
\centering
\small
\caption{Settings of every RL stage. Steps: the checkpoint used. All stages use 10 tasks $\times$ 8 rollouts per step, temperature 1.0, learning rate $3\times10^{-5}$, and 4{,}096 output tokens per turn.}
\label{tab:rl_settings}
\begin{tabular}{@{}llrrrll@{}}
\toprule
Model & Stage & Steps & Turns & Context & Reward & KL \\
\midrule
All Difficulties & 1--5 hunks, rounds 1--3 & 3 $\times$ 48 & 40 & 40{,}960 & \texttt{mlperf\_converge} & 0 \\
 & 10--12 hunks & 100 & 40 & 40{,}960 & \texttt{judge\_progress} & 0 \\
 & 30--34 hunks & 72 & 80, then 60 & 81{,}920 & \texttt{judge\_progress} & 0.001 \\
Easy Only & 1--5 hunks & 150 & 40 & 40{,}960 & \texttt{judge\_progress} & 0 \\
Hard Only & 30--34 hunks & 400 & 60 & 49{,}152 & \texttt{judge\_progress} & 0 \\
\bottomrule
\end{tabular}
\end{table}

\subsection{Training Dynamics and Checkpoint Selection}
\label{app:rl_curves}

\begin{figure}[!htbp]
\centering
\includegraphics[width=\linewidth]{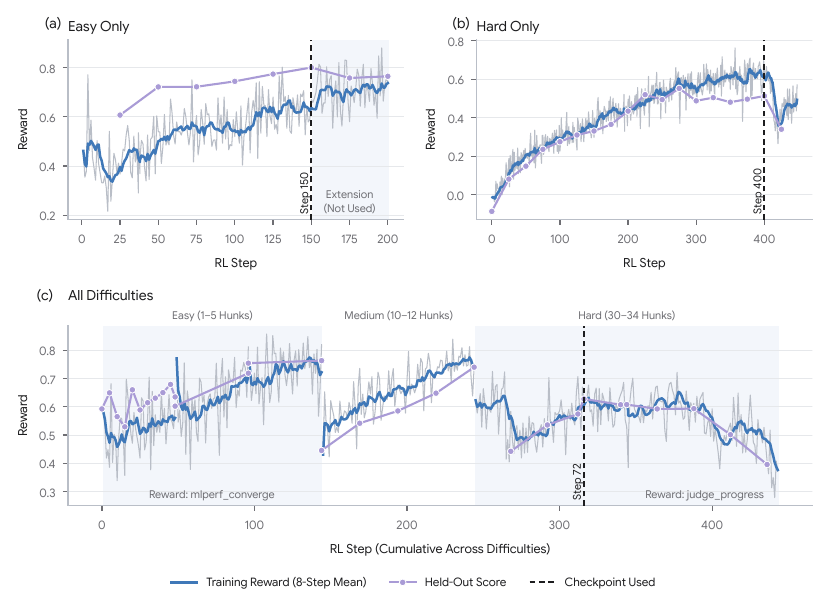}
\caption{\textbf{RL training reward and held-out validation curves across all three curriculum ablations show consistent learning followed by late-stage overfitting.} Per-step training reward (thin gray), 8-step moving average (blue), held-out evaluation score on each run's target difficulty band (lavender markers), and the exported checkpoint selected for final evaluation (vertical dashed line) for \textbf{(a)} Easy Only (1--5 hunks, peaking at step 150 before a 50-step unused extension), \textbf{(b)} Hard Only (30--34 hunks, evaluated at step 400 prior to late collapse), and \textbf{(c)} All Difficulties across cumulative Easy, Medium, and Hard stages (selected at Hard step 72). In \textbf{(c)}, the Easy stage uses \texttt{mlperf\_converge} while Medium and Hard use \texttt{judge\_progress}, so vertical reward levels are comparable within, rather than across, that stage boundary.}
\label{fig:rl_curves}
\end{figure}

Figure~\ref{fig:rl_curves} plots training reward and held-out validation trajectories for all three RL curricula evaluated in Table~\ref{tab:bmbl_results}. Checkpoints are selected by a single rule across all runs: the exported checkpoint maximizing held-out validation score on that curriculum's target difficulty band (note that validation scores are comparable within, rather than across, curricula because target bands and reward functions differ). Across all three regimes, prolonged optimization eventually degrades held-out performance: Easy Only declines during a 50-step extension past step 150, All Difficulties declines after Hard-stage step 72, and Hard Only drops sharply past step 400.

\begin{figure}[!htbp]
\centering
\includegraphics[width=\linewidth]{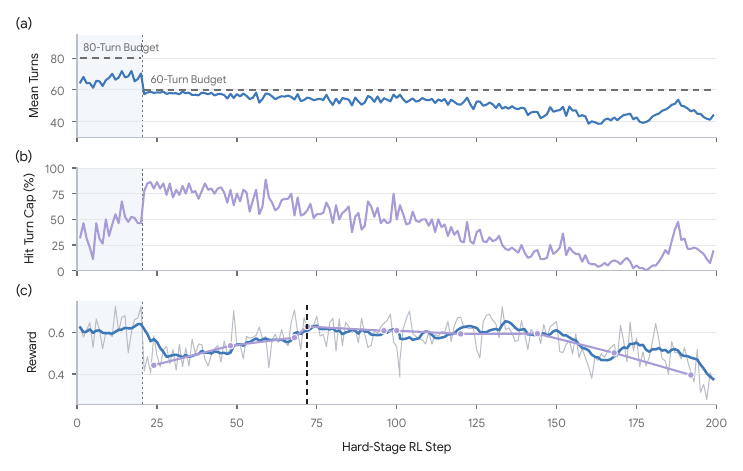}
\caption{\textbf{Cutting the turn budget from 80 to 60 mid-stage costs reward briefly, then the agent learns to finish within it.} The full hard stage of the All Difficulties run, including its first 20 steps at an 80-turn budget (shaded). \textbf{(a)} Mean turns per episode with each budget (dashed). \textbf{(b)} Share of episodes that hit the turn cap: it jumps when the budget drops, then falls steadily. \textbf{(c)} Training reward per step (gray) and 8-step mean (blue), held-out score (lavender), and the checkpoint we evaluate (dashed).}
\label{fig:hard_budget}
\end{figure}

\paragraph{Case study: lowering the turn budget mid-stage.} To bound KV-cache memory over long multi-bug rollouts during the All Difficulties Hard stage, the first 20 steps ran with an 80-turn budget and all subsequent steps ran at 60 turns (Figure~\ref{fig:hard_budget}; Figure~\ref{fig:rl_reward_turns} omits these initial 20 steps so that main-text curves reflect fixed stage settings). Under the 80-turn cap, episodes averaged 67 turns and 42\% reached the turn limit. Lowering the cap to 60 turns initially raised the cap-hit rate to 81\% over the next ten steps and briefly depressed reward because truncated episodes cannot invoke \texttt{finalize}. Training quickly adapted, however: reward recovered prior to the selected evaluation checkpoint, held-out validation rose from $0.44$ to $0.63$ at the selected checkpoint, while mean episode length fell to 45 turns and the cap-hit rate dropped to 19\% by the end of the run, indicating that tighter turn budgets impose only transient reward loss while encouraging turn-efficient repair (Figure~\ref{fig:rl_reward_turns}b).

\FloatBarrier
\section{Execution Harness and Evaluation Setup}

\subsection{Sandbox Harness and Episode Protocol}
\label{app:harness}

A unified container harness executes all interactive episodes across RL rollouts, \bmbl{} evaluations, and commercial baselines. At each turn, the agent emits free-form reasoning and XML-wrapped tool calls, which the harness parses, executes inside a GPU-enabled Docker container holding the repository and read-only dataset, and returns as structured observations (Algorithms~\ref{alg:episode} and~\ref{alg:score}).

\paragraph{Tools and protocol.} Agents interact with the container through six tools (Table~\ref{tab:tools}). Each assistant response may invoke multiple tool calls formatted as
\begin{center}\small\texttt{<tool\_call>\{\textquotedbl name\textquotedbl: \textquotedbl read\_file\textquotedbl, \textquotedbl input\textquotedbl: \{\textquotedbl path\textquotedbl: \textquotedbl main.py\textquotedbl\}\}</tool\_call>}\end{center}
whose outputs are concatenated in the subsequent observation. Unparseable calls are recorded as \emph{malformed} turns and return a syntax error. Tool outputs retain their first 1{,}000 and last 3{,}000 characters, and the dataset mount is read-only so candidate repairs cannot modify evaluation data.

\begin{table}[!htbp]
\centering
\small
\caption{The agent's tools. Searching the repository is done through \texttt{bash}.}
\label{tab:tools}
\begin{tabular}{@{}lp{0.62\linewidth}@{}}
\toprule
Tool & Effect \\
\midrule
\texttt{bash(command, workdir)} & runs a shell command in the container, under a per-call timeout \\
\texttt{read\_file(path)} & returns a file's contents \\
\texttt{write\_file(path, content)} & creates or overwrites a file \\
\texttt{modify\_file(path, old, new)} & replaces an exact substring of a file, so a single function can be patched without rewriting the file \\
\texttt{list\_dir(path)} & lists a directory \\
\texttt{finalize(verdict, description)} & ends the episode; the verdict is recorded but not used in scoring \\
\bottomrule
\end{tabular}
\end{table}

\paragraph{Prompt.} The system prompt specifies the task objective, tool schemas, and XML call syntax. On training instances, the initial user message includes the clean baseline metric alongside cached stdout/stderr from the failing buggy run; on \bmbl{}, it provides the competition description, submission contract, and expected clean baseline score while omitting cached stdout/stderr logs, requiring the agent to execute \texttt{python main.py} to observe runtime behavior.

\begin{algorithm}[!htbp]
\caption{One episode. $T$: turn budget; $L$: context budget in tokens; $K$: consecutive malformed-turn limit (3 in training; $T$ in evaluation, so only an episode of malformed turns is cut short).}
\label{alg:episode}
\begin{algorithmic}[1]
\Procedure{Episode}{$\text{task}, \pi, T, L, K$}
  \State start the task's container; mount the data read-only; remove answer-key files
  \State $h_0 \gets$ hashes of the source files
  \State $m \gets [\text{system prompt with tools},\ \text{task prompt}]$; \quad $k \gets 0$
  \For{$t = 1, \ldots, T$}
    \State \textbf{if} the episode clock has run out \textbf{then break}
    \State $a \gets \pi(m)$; \quad $C \gets$ tool calls parsed from $a$
    \If{$a$ contains a call that cannot be parsed}
      \State $k \gets k + 1$; \textbf{if} $k \geq K$ \textbf{then break}
      \State $o \gets$ an error message
    \ElsIf{$C = \emptyset$} \textbf{break} \Comment{a reply without a tool call ends the episode}
    \ElsIf{\texttt{finalize} $\in C$} \textbf{break}
    \Else
      \State $k \gets 0$; \quad $o \gets$ the truncated result of each call in $C$, executed in order
    \EndIf
    \State $o \gets o \oplus$ ``\texttt{[BUDGET]} turn $t$; $T - t$ turns, $L - \text{tokens}(m)$ context tokens remaining \ldots''
    \State $m \gets m \oplus [a, o]$
    \State \textbf{if} $\text{tokens}(m) > L - 512$ \textbf{then break}
  \EndFor
  \State \Return \Call{Score}{container, $h_0$, $k$, $K$}
\EndProcedure
\end{algorithmic}
\end{algorithm}

\begin{algorithm}[!htbp]
\caption{Scoring the repository at the end of an episode, whichever way it ended.}
\label{alg:score}
\begin{algorithmic}[1]
\Procedure{Score}{container, $h_0$, $k$, $K$}
  \State \textit{gate} $\gets$ the first that holds: $k \geq K$ malformed turns; source hashes equal $h_0$ (no change); code does not compile; entry point no longer reaches the injected code (bypassed)
  \State \textbf{if} \textit{gate} \textbf{then return} the gate's fixed reward
  \State re-run the entry point (\texttt{run.sh}; on \bmbl{}, \texttt{python main.py}, then the official MLE-Bench grader on \texttt{submission.csv})
  \State \textbf{if} the run crashes or reports no metric \textbf{then return} the crash reward
  \State $\text{recovery} \gets m_{\text{fixed}} / m_{\text{baseline}}$ \quad ($m_{\text{baseline}} / m_{\text{fixed}}$ when lower is better)
  \State $p \gets \mathrm{clip}\big((m_{\text{fixed}} - m_{\text{buggy}}) / (m_{\text{baseline}} - m_{\text{buggy}}),\ 0,\ 1\big)$
  \State \Return $R(p,\ \text{recovery} \geq \theta_{\text{rec}},\ \ldots)$ \Comment{$\theta_{\text{rec}} = 0.99$; reward variant, Appendix~\ref{app:rl_recipe}}
\EndProcedure
\end{algorithmic}
\end{algorithm}

\paragraph{Episode termination.} An episode terminates when the agent invokes \texttt{finalize}, emits a response containing no tool call, reaches the malformed-turn limit $K$, or exhausts its turn, context-token, or wall-clock budget; in all cases, the container state at termination is evaluated identically by Algorithm~\ref{alg:score}.

\paragraph{Settings.} Table~\ref{tab:harness_settings} summarizes shared harness budgets across training and evaluation, and Appendix~\ref{app:eval_setup} details per-model configurations for hosted baselines.

\begin{table}[!htbp]
\centering
\small
\caption{Harness settings. Table~\ref{tab:eval_setup} gives per-model budgets.}
\label{tab:harness_settings}
\begin{tabular}{@{}lccc@{}}
\toprule
 & \makecell{RL\\1--5, 10--12 hunks} & \makecell{RL\\30--34 hunks} & \makecell{\bmbl{}\\1 / 5 / 10 / 50 Bugs} \\
\midrule
turns per episode & 40 & 60$^\S$ & 40 / 60 / 80 / 80$^\ddagger$ \\
context / output per turn (tokens) & 40{,}960 / 4{,}096 & 81{,}920 / 4{,}096$^\dagger$ & 98{,}304 / 4{,}096 \\
malformed limit $K$ & 3 & 3 & the turn budget \\
sampling & temperature 1.0 & temperature 1.0 & temperature 0.7 \\
\bottomrule
\end{tabular}

\vspace{2pt}
{\footnotesize $^\dagger$Hard Only trained with a 49{,}152-token context. $^\ddagger$80 on every tier for Claude Opus~5.5, Sonnet~5.5, Sonnet~5, Qwen3.5-9B, SFT Only, and All Difficulties. $^\S$The first 20 steps of All Difficulties' stage on 30--34 hunks ran with 80 turns.}
\end{table}

\subsection{Baseline Model Configurations}
\label{app:eval_setup}

Table~\ref{tab:eval_setup} details the evaluation configuration for every baseline and trained checkpoint (22 episodes per model and tier, one trajectory per task). Unless noted otherwise, models share a 98{,}304-token context window and turn caps of 40, 60, 80, and 80 on the 1 Bug, 5 Bugs, 10 Bugs, and 50 Bugs tiers. All episodes run under multi-hour wall-clock timeouts (at most 8 hours per episode) sufficient for repeated pipeline executions.

\begin{table}[!htbp]
\centering
\small
\caption{\textbf{Evaluation setup.} Context and output list total context and per-turn output limits where they differ from the shared harness (Table~\ref{tab:harness_settings}).}
\label{tab:eval_setup}
\begin{tabular}{@{}lcc@{}}
\toprule
Model & Turns (1 / 5 / 10 / 50 Bugs) & Context and output \\
\midrule
Claude Opus 5.5~\citep{anthropic2026claudeopus55} & 80 / 80 / 80 / 80 & shared \\
Claude Sonnet 5.5~\citep{anthropic2026claudesonnet55} & 80 / 80 / 80 / 80 & shared \\
Claude Sonnet 5~\citep{anthropic2026claudesonnet5} & 80 / 80 / 80 / 80 & shared \\
Claude Sonnet 4.6~\citep{anthropic2026claudesonnet46} & 40 / 60 / 80 / 80 & 262k; 40{,}960 per turn \\
Gemini 3.8 Flash~\citep{deepmind2026gemini38flash} & 40 / 60 / 80 / 80 & 262k; 40{,}960 per turn \\
Claude Haiku 5.5~\citep{anthropic2026claudehaiku55} & 40 / 60 / 80 / 80 & shared \\
Claude Haiku 4.5~\citep{anthropic2025claudehaiku45} & 40 / 60 / 80 / 80 & 150k; 32{,}000 per turn \\
Gemini 3.5 Flash-Lite~\citep{deepmind2026gemini35flashlite} & 40 / 60 / 80 / 80 & 262k; 40{,}960 per turn \\
Qwen3.8-27B~\citep{qwen38} (thinking off) & 40 / 60 / 80 / 80 & shared \\
Qwen3.5-9B~\citep{qwen3.5} & 80 / 80 / 80 / 80 & shared \\
\midrule
SFT Only (9B) & 80 / 80 / 80 / 80 & shared \\
SFT+RL, Easy Only & 40 / 60 / 80 / 80 & shared \\
SFT+RL, Hard Only & 40 / 60 / 80 / 80 & shared \\
SFT+RL, All Difficulties & 80 / 80 / 80 / 80 & shared \\
\bottomrule
\end{tabular}
\end{table}

\paragraph{Turn and termination behavior.} On 10 Bugs and 50 Bugs, turn caps are uniform (80 turns) across all models, while 1 Bug and 5 Bugs use 40- and 60-turn caps under the default schedule and 80 turns under the uniform schedule (Table~\ref{tab:eval_setup}); these lower caps do not bind in practice, as Easy Only, Hard Only, and Claude Haiku~5.5 hit the turn cap on 0\% of 1 Bug episodes and at most 4.5\% of 5 Bugs episodes (Table~\ref{tab:bmbl_full}). Termination behavior instead distinguishes our single-band curricula from All Difficulties: Hard Only reaches its episode timeout on 12 of 22 episodes on 1 Bug, and Easy Only terminates prematurely without emitting a tool call on 17 of 22 episodes on both 5 Bugs and 10 Bugs.

\paragraph{Expanded context and output caps.} Four hosted baselines (Claude Sonnet~4.6, Claude Haiku~4.5, Gemini~3.8 Flash, and Gemini~3.5 Flash-Lite) are evaluated with expanded context windows and higher per-turn output limits (Table~\ref{tab:eval_setup}) to accommodate longer per-turn reasoning traces under the shared prompt and XML tool interface (Appendix~\ref{app:harness}).

\paragraph{Inference backends.} Anthropic baselines execute via the Claude Code CLI (\texttt{claude -p}), invoked once per turn with built-in tools and extended thinking disabled; Gemini baselines execute via the Vertex AI API within our harness loop at default thinking settings; and Qwen models are served locally.

\paragraph{Model tiers.} As defined in Section~\ref{sec:setup}, Budget Models comprise each commercial provider's lowest-cost tier (Claude Haiku~4.5/5.5 and Gemini~3.5 Flash-Lite) alongside open-weight models up to 27B parameters (Qwen3.8-27B and Qwen3.5-9B); all larger commercial flagships constitute Frontier Models.

\paragraph{Qwen3.8-27B.}\label{sec:ood_setup} Qwen3.8-27B is evaluated with thinking disabled on every tier, because an initial thinking-enabled run failed to invoke \texttt{finalize} across all 66 episodes on the 1, 5, and 10 Bugs tiers.

\subsection{LLM Judges and Code Review}
\label{app:reviewer}

Section~\ref{sec:training}'s repository-level semantic review is implemented with Claude Opus~4.8 (temperature $0$) to avoid the cost of issuing separate per-site judge calls across thousands of multi-bug rollouts on the 200-task pool (Appendix~\ref{app:cci_sweep}), operating in a single pass from code diffs and Oracle ablation records without executing an external test suite.

\paragraph{Inputs.} The reviewer receives three unified diffs (clean-to-fixed, buggy-to-fixed, and clean-to-buggy, each truncated at 50{,}000 characters), together with the Oracle's structured ablation report listing verified causal mutations (affected files, defect mechanism, and isolated metric impact), confirmed no-effect decoys, and any metric-improving mutations, grounding the evaluation in measured causal attribution rather than subjective code style.

\paragraph{Verdicts.} Parallel sub-checks evaluate two binary criteria: whether the candidate patch actually fixes the verified causal mutations (\emph{Fix Addresses Bug} in Figure~\ref{fig:criterion_sensitivity}), and whether the repair is principally valid rather than bypassing the faulty logic, leaving the underlying bug intact, or introducing a new defect in its place (\emph{Principally Valid}). A candidate patch that modifies only decoy mutations fails the first check even if the evaluation metric recovers.

\paragraph{Judge models and scope.} Across our pipeline, judge backbones are chosen to match call volume and latency constraints: single-pass offline repository reviews on the 200-task sweep use Claude Opus~4.8 (temperature $0$), whereas concurrent per-site checks use low-latency models, specifically Claude Haiku~5.5 for \bmbl{} full-repair grading (Section~\ref{sec:evaluation}) and Gemini~3.8 Flash for online \texttt{judge\_progress} RL rollouts (Appendix~\ref{app:rl_recipe}; the RL environment's auxiliary repository-reviewer penalty remained disabled across all reported training runs). Section~\ref{sec:hard_bugs} and Appendix~\ref{app:bug_analysis} further validate bug-level LLM judgments against blind human review.

\FloatBarrier
\section{Extended Experimental Results}

\subsection{Representative Repair Trajectories}
\label{app:traj_examples}

\begin{figure}[!htbp]
\centering
\includegraphics[width=\linewidth]{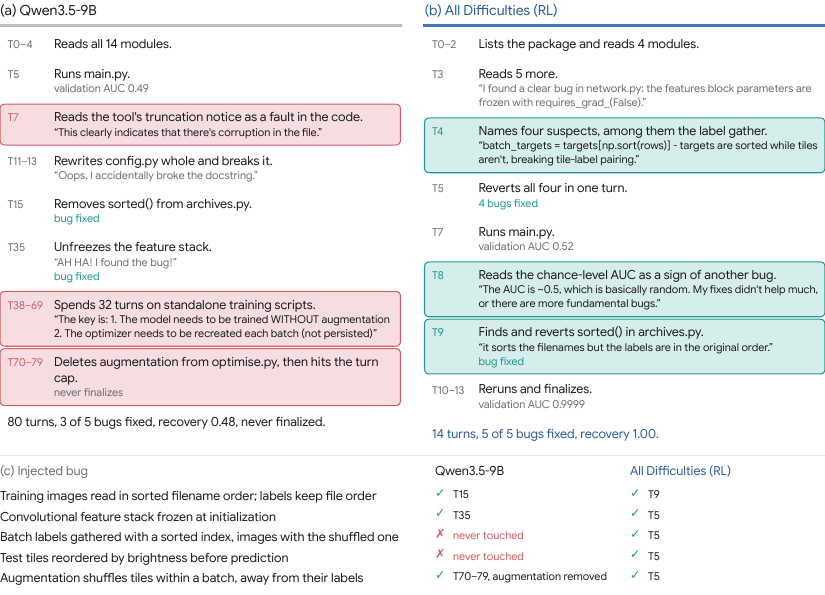}
\caption{\textbf{Curriculum RL replaces blind pipeline rewrites with targeted multi-file diagnosis and verification on a 5 Bugs task.} Side-by-side episode trace on \emph{Aerial Cactus} (each model's reported sample; \texttt{T}: turn; quotes verbatim from model reasoning). \textbf{(a)} Untuned Qwen3.5-9B spends 80 turns rewriting files and standalone scripts, fixing 3/5 bugs without finalizing (recovery 0.48; red boxes mark failure modes). \textbf{(b)} All Difficulties diagnoses four suspects across modules, reverts them in one turn, catches the fifth bug from a chance-level validation AUC, and finalizes in 14 turns with 5/5 bugs fixed (recovery 1.00; green boxes mark diagnostic reasoning and repairs). \textbf{(c)} Per-bug judge audit and fixing turn.}
\label{fig:trajectory}
\end{figure}

\begin{figure}[!htbp]
\centering
\includegraphics[width=\linewidth]{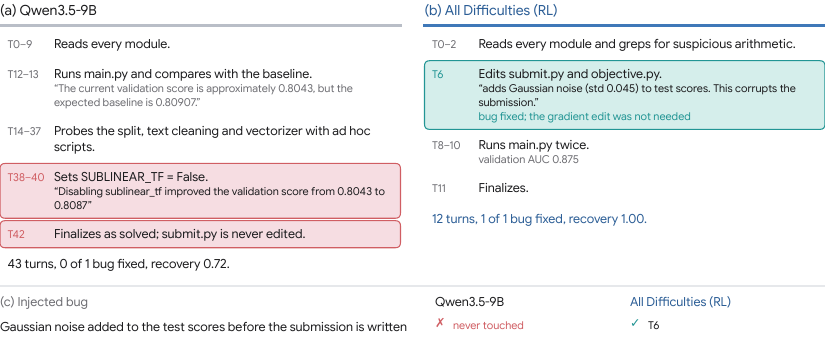}
\caption{\textbf{On a silent 1 Bug task, the RL agent inspects submission code directly whereas the base model overfits validation noise.} Side-by-side episode trace on \emph{Detecting Insults} (layout as in Figure~\ref{fig:trajectory}). \textbf{(a)} Untuned Qwen3.5-9B misinterprets a 0.005 validation gap as the defect, tunes vectorizer hyperparameters across 43 turns without inspecting \texttt{submit.py}, and finalizes with 0/1 bugs fixed (recovery 0.72). \textbf{(b)} All Difficulties locates and removes the injected test-score Gaussian noise in \texttt{submit.py} by turn 6, verifies on \texttt{main.py}, and finalizes in 12 turns with full recovery (1.00). \textbf{(c)} Per-bug judge audit.}
\label{fig:trajectory_appendix}
\end{figure}

\begin{figure}[!htbp]
\centering
\includegraphics[width=\linewidth]{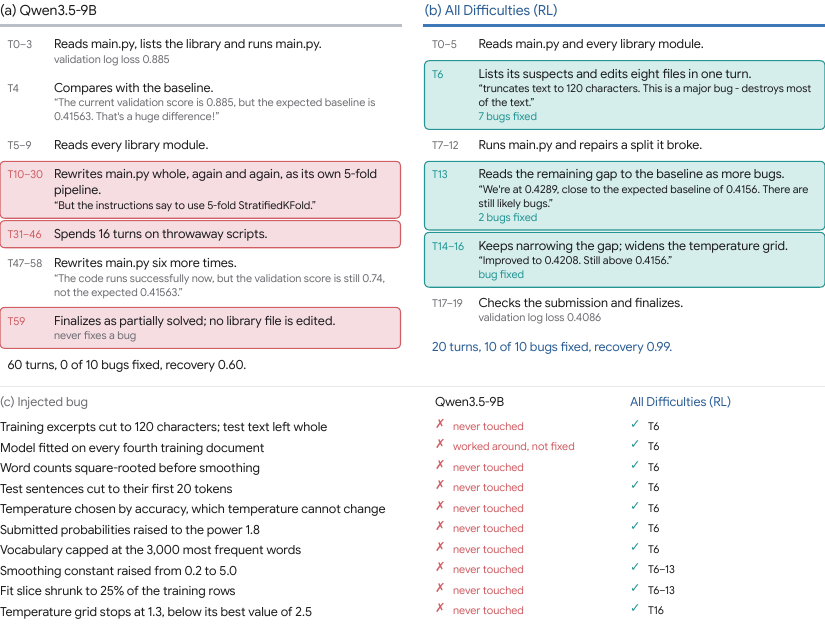}
\caption{\textbf{On a 10 Bugs repository, the RL agent systematically clears all ten defects across library modules while the base model rewrites \texttt{main.py} around them.} Side-by-side episode trace on \emph{Spooky Authors} (layout as in Figure~\ref{fig:trajectory}). \textbf{(a)} Untuned Qwen3.5-9B repeatedly rewrites \texttt{main.py} from scratch over 60 turns without editing any underlying \texttt{library/} module, fixing 0/10 bugs (recovery 0.60). \textbf{(b)} All Difficulties patches seven bugs across eight files in a single turn (\texttt{T6}), uses residual validation log-loss gaps against the baseline to isolate the remaining three subtle hyperparameter and data-slice bugs (\texttt{T13--T16}), and resolves all 10/10 bugs in 20 turns (recovery 0.99). \textbf{(c)} Per-bug judge audit across all ten injected defects.}
\label{fig:trajectory_appendix_10}
\end{figure}

To illustrate the behavioral shift quantified in RQ4 (Section~\ref{sec:results}), Figures~\ref{fig:trajectory}--\ref{fig:trajectory_appendix_10} compare episode traces of the untuned Qwen3.5-9B base model and our All Difficulties RL agent on representative 5 Bugs (\emph{Aerial Cactus}, discussed in RQ4; Figure~\ref{fig:trajectory}), 1 Bug (\emph{Detecting Insults}; Figure~\ref{fig:trajectory_appendix}), and 10 Bugs (\emph{Spooky Authors}; Figure~\ref{fig:trajectory_appendix_10}) repositories.

On \texttt{detecting-insults-in-social-commentary} (1 Bug; Figure~\ref{fig:trajectory_appendix}), the sole injected defect adds Gaussian noise to test-time predictions in \texttt{submit.py} without affecting validation scores. The base model misinterprets the small gap between its local validation score ($0.804$) and the reference baseline ($0.809$) as the defect, tunes a TF-IDF vectorizer flag, and terminates without inspecting \texttt{submit.py}. By contrast, the All Difficulties agent audits the submission path, removes the injected noise within 12 turns, and restores full test performance (alongside a superfluous gradient-scaling edit that does not harm recovery).

On \texttt{spooky-author-\allowbreak identification} (10 Bugs; Figure~\ref{fig:trajectory_appendix_10}), the base model repeatedly rewrites \texttt{main.py} into a standalone cross-validation script without modifying the underlying \texttt{library/} modules that contain the defects, incidentally bypassing one bug and resolving none. The All Difficulties agent patches seven defects across eight library files in a single turn, uses the residual gap to the baseline log-loss to isolate the remaining three subtle hyperparameter and data-slice defects, and resolves all ten bugs in 20 turns.

\FloatBarrier
\subsection{Full \bmbl{} Breakdown}
\label{app:bmbl_full}

\begin{figure}[!htbp]
\centering
\includegraphics[width=\linewidth]{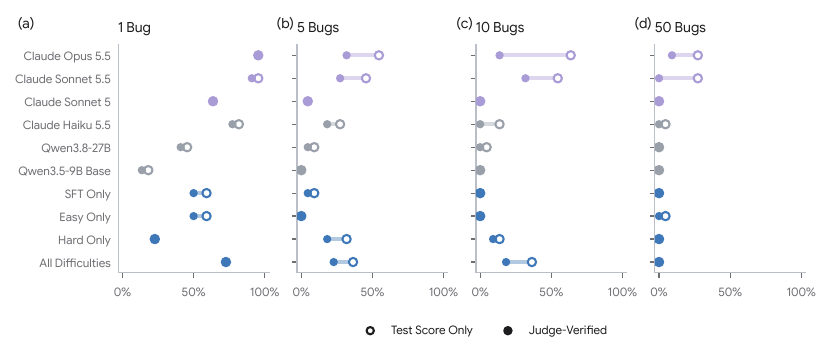}
\caption{\textbf{Test-metric recovery substantially overstates true bug repair as defect count grows, especially for frontier models.} Task solve rate on \bmbl{} under metric recovery alone ($\text{recovery} \ge 0.99$, open circles) versus strict judge-verified full repair requiring every injected defect to be resolved (filled circles), across \textbf{(a)} 1 Bug, \textbf{(b)} 5 Bugs, \textbf{(c)} 10 Bugs, and \textbf{(d)} 50 Bugs (22 tasks per panel; lavender: frontier models, gray: budget baselines, blue: our trained 9B models). The two criteria coincide on 1 Bug, where fixing the sole bug is necessary to recover the score, but diverge sharply from 5 Bugs onward as strong models compensate for unfixed or silent defects via modeling improvements elsewhere (e.g., Claude Opus~5.5 reaches 64\% score recovery on 10 Bugs vs.\ 14\% verified full repair).}
\label{fig:full_repair_gap}
\end{figure}

\begin{figure}[!htbp]
\centering
\includegraphics[width=\linewidth]{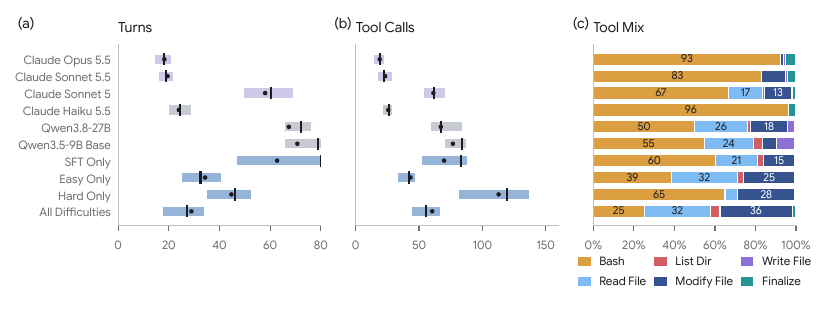}
\caption{\textbf{Curriculum RL halves episode length and shifts tool usage from exploratory shell commands toward structured file reads and patches.} Episode-level execution statistics across the 22 tasks of the 10 Bugs tier (lavender: frontier models, gray: budget baselines, blue: our trained 9B models). \textbf{(a)} Turns per episode and \textbf{(b)} total tool calls per episode (shaded bar: interquartile range; vertical tick: median; dot: mean; 80-turn budget). Whereas Qwen3.5-9B Base and SFT Only routinely exhaust the 80-turn cap, All Difficulties finishes in a median of ${\sim}27$ turns, comparable to Claude Haiku~5.5 and Sonnet~5.5. \textbf{(c)} Share of tool calls by tool type: Claude Code baselines execute predominantly through \texttt{bash} (67--96\%), whereas All Difficulties balances \texttt{bash} (25\%), \texttt{read\_file} (32\%), and \texttt{modify\_file} (36\%) while eliminating the whole-file \texttt{write\_file} calls seen in the untuned base model.}
\label{fig:agent_behavior}
\end{figure}

\begin{figure}[!htbp]
\centering
\includegraphics[width=\linewidth]{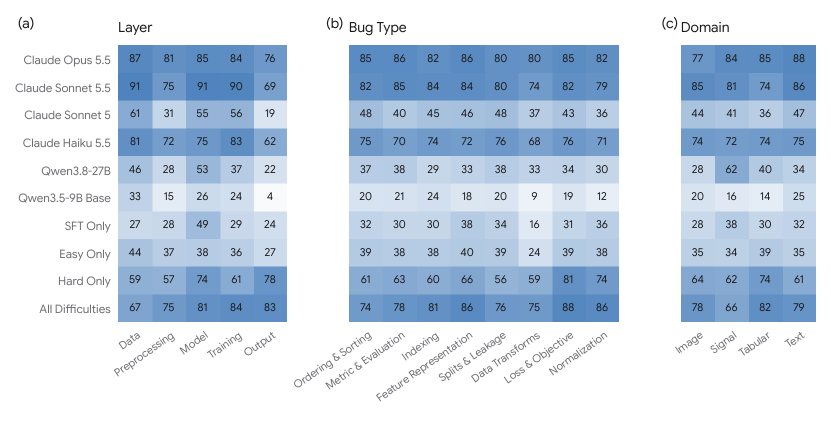}
\caption{\textbf{High-complexity RL closes the gap with frontier models across pipeline stages, defect categories, and data modalities, particularly on output-layer bugs.} Percentage of individual injected bugs judged resolved, pooled across the 1 Bug, 5 Bugs, and 10 Bugs tiers (darker blue indicates higher resolution rate), broken down by \textbf{(a)} ML pipeline layer, \textbf{(b)} the eight most frequent defect categories, and \textbf{(c)} competition data modality. Output-formatting defects are the hardest layer for base, SFT, and most external models (4\% for Qwen3.5-9B Base, 19\% for Claude Sonnet~5, 69\% for Sonnet~5.5), whereas Hard Only (78\%) and All Difficulties (83\%) match or exceed Claude Opus~5.5 (76\%). Across modalities \textbf{(c)}, All Difficulties transfers evenly across image, signal, tabular, and text repositories (66--82\%).}
\label{fig:bug_resolve_by_type}
\end{figure}

Table~\ref{tab:bmbl_full} expands Table~\ref{tab:bmbl_results} with the complete suite of outcome and behavioral metrics across all four \bmbl{} tiers, accompanied by three diagnostic breakdowns: Figure~\ref{fig:full_repair_gap} contrasts metric-only recovery against judge-verified full repair across tiers, Figure~\ref{fig:agent_behavior} characterizes episode lengths and tool-use distributions on 10 Bugs, and Figure~\ref{fig:bug_resolve_by_type} disaggregates per-bug resolution rates across pipeline layers, defect categories, and data modalities.

\begin{table}[!htbp]
\centering
\caption{\textbf{All metrics on \bmbl{}}, 22 episodes per model and tier. Arrows, shading, and the Reference rows follow Table~\ref{tab:bmbl_results}. Bugs fixed: injected bugs judged fixed, whether or not the repository still runs (not judged for Qwen3.8-27B on 50 Bugs). Improved repo: the repository beats its injected start. Ended with finalize tool: episodes terminating via an explicit \texttt{finalize} call. Tool-call success: $1 -$ failed/attempted actions. Hit turn cap: the episode used its whole turn budget. Tasks above clean repo: tasks where the repaired code ranks above the clean repository.}
\label{tab:bmbl_full}
\resizebox{\linewidth}{!}{
\begin{tabular}{llccccccccc}
\toprule[1.5pt]
Variant & Model & \makecell[c]{Recovery (\%)\\\hib} & \makecell[c]{Full Repair (\%)\\\hib} & \makecell[c]{Bugs\\Fixed (\%)\\\hib} & \makecell[c]{Improved Repo (\%)\\\hib} & \makecell[c]{Ended with\\Finalize Tool (\%)\\\hib} & \makecell[c]{Tool-Call\\Success (\%)\\\hib} & \makecell[c]{Hit Turn\\Cap (\%)\\\lob} & \makecell[c]{Leaderboard\\Percentile (\%)\\\hib} & \makecell[c]{Tasks Above\\Clean Repo\\\hib} \\
\midrule\midrule
\textit{Reference} & \textit{Clean repository (1, 5, 10 Bugs)} & -- & -- & -- & -- & -- & -- & -- & \textit{49.1} & -- \\
\midrule
\multirow{15}{*}{1 Bug} & Claude Opus 5.5 & 99.8 & 95.5 & 95.5 & 100.0 & 100.0 & 96.7 & 0.0 & 57.2 & 14 \\
 & Claude Sonnet 5.5 & 95.5 & 90.9 & 90.9 & 95.5 & 100.0 & 91.8 & 0.0 & 54.0 & 11 \\
 & Claude Sonnet 5 & 80.2 & 63.6 & 63.6 & 86.4 & 100.0 & 89.9 & 0.0 & 39.2 & 5 \\
 & Claude Sonnet 4.6 & 30.9 & 27.3 & 27.3 & 27.3 & 72.7 & 95.9 & 0.0 & 17.5 & 1 \\
 & Gemini 3.8 Flash & 27.3 & 27.3 & 27.3 & 27.3 & 90.9 & 97.1 & 0.0 & 15.6 & 3 \\
 & Claude Haiku 5.5 & \cellcolor{royalblue!40}\textbf{89.5} & \cellcolor{royalblue!40}\textbf{77.3} & \cellcolor{royalblue!38}\underline{77.3} & \cellcolor{royalblue!40}\textbf{86.4} & \cellcolor{royalblue!40}\textbf{100.0} & \cellcolor{royalblue!39}98.7 & \cellcolor{royalblue!40}\textbf{0.0} & \cellcolor{royalblue!40}\textbf{46.1} & \cellcolor{royalblue!34}\underline{5} \\
 & Claude Haiku 4.5 & \cellcolor{royalblue!16}33.3 & \cellcolor{royalblue!8}9.1 & \cellcolor{royalblue!8}9.1 & \cellcolor{royalblue!14}27.3 & \cellcolor{royalblue!27}77.3 & \cellcolor{royalblue!18}89.8 & \cellcolor{royalblue!40}\textbf{0.0} & \cellcolor{royalblue!12}11.2 & \cellcolor{royalblue!12}1 \\
 & Gemini 3.5 Flash-Lite & \cellcolor{royalblue!6}8.2 & \cellcolor{royalblue!6}4.5 & \cellcolor{royalblue!6}4.5 & \cellcolor{royalblue!6}9.1 & \cellcolor{royalblue!19}63.6 & \cellcolor{royalblue!6}85.0 & \cellcolor{royalblue!40}\textbf{0.0} & \cellcolor{royalblue!6}3.8 & \cellcolor{royalblue!6}0 \\
 & Qwen3.8-27B & \cellcolor{royalblue!33}72.7 & \cellcolor{royalblue!23}40.9 & \cellcolor{royalblue!24}45.5 & \cellcolor{royalblue!30}63.6 & \cellcolor{royalblue!11}50.0 & \cellcolor{royalblue!6}84.9 & \cellcolor{royalblue!23}\underline{9.1} & \cellcolor{royalblue!31}34.9 & \cellcolor{royalblue!34}\underline{5} \\
 & Qwen3.5-9B & \cellcolor{royalblue!29}62.1 & \cellcolor{royalblue!10}13.6 & \cellcolor{royalblue!12}18.2 & \cellcolor{royalblue!24}50.0 & \cellcolor{royalblue!22}68.2 & \cellcolor{royalblue!26}93.3 & \cellcolor{royalblue!6}18.2 & \cellcolor{royalblue!23}24.7 & \cellcolor{royalblue!12}1 \\
 & Ours (9B SFT) & \cellcolor{royalblue!32}70.7 & \cellcolor{royalblue!27}50.0 & \cellcolor{royalblue!26}50.0 & \cellcolor{royalblue!34}72.7 & \cellcolor{royalblue!30}\underline{81.8} & \cellcolor{royalblue!40}\textbf{99.2} & \cellcolor{royalblue!15}13.6 & \cellcolor{royalblue!33}36.9 & \cellcolor{royalblue!23}3 \\
\cmidrule{2-11}
 & \multicolumn{10}{l}{\textit{Ours: SFT+RL Curriculum:}} \\
 & \quad Easy Only (1--5 Hunks) & \cellcolor{royalblue!33}73.9 & \cellcolor{royalblue!27}50.0 & \cellcolor{royalblue!26}50.0 & \cellcolor{royalblue!32}68.2 & \cellcolor{royalblue!22}68.2 & \cellcolor{royalblue!37}98.1 & \cellcolor{royalblue!40}\textbf{0.0} & \cellcolor{royalblue!32}35.6 & \cellcolor{royalblue!23}3 \\
 & \quad Hard Only (30--34 Hunks) & \cellcolor{royalblue!32}69.4 & \cellcolor{royalblue!14}22.7 & \cellcolor{royalblue!30}59.1 & \cellcolor{royalblue!34}72.7 & \cellcolor{royalblue!6}40.9 & \cellcolor{royalblue!38}98.4 & \cellcolor{royalblue!40}\textbf{0.0} & \cellcolor{royalblue!30}34.2 & \cellcolor{royalblue!29}4 \\
 & \quad All Difficulties (1--5, 10--12, then 30--34) & \cellcolor{royalblue!37}\underline{81.5} & \cellcolor{royalblue!38}\underline{72.7} & \cellcolor{royalblue!40}\textbf{81.8} & \cellcolor{royalblue!38}\underline{81.8} & \cellcolor{royalblue!24}72.7 & \cellcolor{royalblue!39}\underline{98.9} & \cellcolor{royalblue!23}\underline{9.1} & \cellcolor{royalblue!38}\underline{43.0} & \cellcolor{royalblue!40}\textbf{6} \\
\midrule
\multirow{15}{*}{5 Bugs} & Claude Opus 5.5 & 88.4 & 31.8 & 82.7 & 95.5 & 100.0 & 99.1 & 0.0 & 44.6 & 10 \\
 & Claude Sonnet 5.5 & 85.5 & 27.3 & 77.3 & 95.5 & 95.5 & 93.0 & 0.0 & 42.4 & 7 \\
 & Claude Sonnet 5 & 67.5 & 4.5 & 47.3 & 90.9 & 100.0 & 91.4 & 0.0 & 24.2 & 0 \\
 & Claude Sonnet 4.6 & 25.6 & 0.0 & 13.6 & 22.7 & 77.3 & 95.3 & 0.0 & 5.8 & 0 \\
 & Gemini 3.8 Flash & 41.2 & 13.6 & 33.6 & 40.9 & 68.2 & 95.3 & 4.5 & 21.9 & 2 \\
 & Claude Haiku 5.5 & \cellcolor{royalblue!39}\underline{75.2} & \cellcolor{royalblue!33}\underline{18.2} & \cellcolor{royalblue!40}\textbf{75.5} & \cellcolor{royalblue!38}\underline{86.4} & \cellcolor{royalblue!40}\textbf{100.0} & \cellcolor{royalblue!39}97.5 & \cellcolor{royalblue!40}\textbf{0.0} & \cellcolor{royalblue!36}\underline{33.4} & \cellcolor{royalblue!26}\underline{3} \\
 & Claude Haiku 4.5 & \cellcolor{royalblue!11}17.3 & \cellcolor{royalblue!6}0.0 & \cellcolor{royalblue!7}7.3 & \cellcolor{royalblue!14}27.3 & \cellcolor{royalblue!25}63.6 & \cellcolor{royalblue!30}90.1 & \cellcolor{royalblue!36}\underline{4.5} & \cellcolor{royalblue!7}4.1 & \cellcolor{royalblue!6}0 \\
 & Gemini 3.5 Flash-Lite & \cellcolor{royalblue!6}8.0 & \cellcolor{royalblue!6}0.0 & \cellcolor{royalblue!6}4.5 & \cellcolor{royalblue!6}9.1 & \cellcolor{royalblue!25}63.6 & \cellcolor{royalblue!22}82.8 & \cellcolor{royalblue!40}\textbf{0.0} & \cellcolor{royalblue!6}2.8 & \cellcolor{royalblue!6}0 \\
 & Qwen3.8-27B & \cellcolor{royalblue!24}44.0 & \cellcolor{royalblue!13}4.5 & \cellcolor{royalblue!19}31.8 & \cellcolor{royalblue!25}54.5 & \cellcolor{royalblue!6}18.2 & \cellcolor{royalblue!6}69.9 & \cellcolor{royalblue!17}27.3 & \cellcolor{royalblue!18}14.8 & \cellcolor{royalblue!13}1 \\
 & Qwen3.5-9B & \cellcolor{royalblue!26}49.3 & \cellcolor{royalblue!6}0.0 & \cellcolor{royalblue!18}29.1 & \cellcolor{royalblue!32}72.7 & \cellcolor{royalblue!19}50.0 & \cellcolor{royalblue!34}92.7 & \cellcolor{royalblue!6}40.9 & \cellcolor{royalblue!18}14.5 & \cellcolor{royalblue!6}0 \\
 & Ours (9B SFT) & \cellcolor{royalblue!33}62.7 & \cellcolor{royalblue!13}4.5 & \cellcolor{royalblue!22}37.3 & \cellcolor{royalblue!40}\textbf{90.9} & \cellcolor{royalblue!19}50.0 & \cellcolor{royalblue!40}97.6 & \cellcolor{royalblue!21}22.7 & \cellcolor{royalblue!25}22.0 & \cellcolor{royalblue!13}1 \\
\cmidrule{2-11}
 & \multicolumn{10}{l}{\textit{Ours: SFT+RL Curriculum:}} \\
 & \quad Easy Only (1--5 Hunks) & \cellcolor{royalblue!29}54.6 & \cellcolor{royalblue!6}0.0 & \cellcolor{royalblue!19}31.8 & \cellcolor{royalblue!32}72.7 & \cellcolor{royalblue!6}18.2 & \cellcolor{royalblue!38}96.4 & \cellcolor{royalblue!36}\underline{4.5} & \cellcolor{royalblue!15}12.2 & \cellcolor{royalblue!6}0 \\
 & \quad Hard Only (30--34 Hunks) & \cellcolor{royalblue!35}67.3 & \cellcolor{royalblue!33}\underline{18.2} & \cellcolor{royalblue!34}63.6 & \cellcolor{royalblue!36}81.8 & \cellcolor{royalblue!19}50.0 & \cellcolor{royalblue!40}\textbf{98.0} & \cellcolor{royalblue!40}\textbf{0.0} & \cellcolor{royalblue!34}31.0 & \cellcolor{royalblue!26}\underline{3} \\
 & \quad All Difficulties (1--5, 10--12, then 30--34) & \cellcolor{royalblue!40}\textbf{77.4} & \cellcolor{royalblue!40}\textbf{22.7} & \cellcolor{royalblue!38}\underline{71.8} & \cellcolor{royalblue!40}\textbf{90.9} & \cellcolor{royalblue!32}\underline{81.8} & \cellcolor{royalblue!40}\underline{97.7} & \cellcolor{royalblue!40}\textbf{0.0} & \cellcolor{royalblue!40}\textbf{37.0} & \cellcolor{royalblue!40}\textbf{5} \\
\midrule
\multirow{15}{*}{10 Bugs} & Claude Opus 5.5 & 92.2 & 13.6 & 80.9 & 95.5 & 100.0 & 99.0 & 0.0 & 46.7 & 10 \\
 & Claude Sonnet 5.5 & 86.4 & 31.8 & 84.1 & 95.5 & 100.0 & 95.6 & 0.0 & 43.6 & 6 \\
 & Claude Sonnet 5 & 53.7 & 0.0 & 38.6 & 68.2 & 95.5 & 91.7 & 0.0 & 12.6 & 0 \\
 & Claude Sonnet 4.6 & 30.1 & 0.0 & 16.4 & 40.9 & 68.2 & 95.2 & 0.0 & 4.5 & 0 \\
 & Gemini 3.8 Flash & 18.5 & 13.6 & 22.3 & 22.7 & 86.4 & 96.3 & 0.0 & 14.1 & 3 \\
 & Claude Haiku 5.5 & \cellcolor{royalblue!35}\underline{71.1} & \cellcolor{royalblue!6}0.0 & \cellcolor{royalblue!37}\underline{73.2} & \cellcolor{royalblue!38}\underline{81.8} & \cellcolor{royalblue!38}\underline{95.5} & \cellcolor{royalblue!40}\textbf{98.8} & \cellcolor{royalblue!40}\textbf{0.0} & \cellcolor{royalblue!21}\underline{19.3} & \cellcolor{royalblue!12}\underline{1} \\
 & Claude Haiku 4.5 & \cellcolor{royalblue!11}18.7 & \cellcolor{royalblue!6}0.0 & \cellcolor{royalblue!8}5.5 & \cellcolor{royalblue!16}31.8 & \cellcolor{royalblue!30}77.3 & \cellcolor{royalblue!36}97.1 & \cellcolor{royalblue!40}\textbf{0.0} & \cellcolor{royalblue!7}2.6 & \cellcolor{royalblue!6}0 \\
 & Gemini 3.5 Flash-Lite & \cellcolor{royalblue!6}6.9 & \cellcolor{royalblue!6}0.0 & \cellcolor{royalblue!6}1.4 & \cellcolor{royalblue!6}9.1 & \cellcolor{royalblue!18}50.0 & \cellcolor{royalblue!19}88.6 & \cellcolor{royalblue!40}\textbf{0.0} & \cellcolor{royalblue!6}1.3 & \cellcolor{royalblue!6}0 \\
 & Qwen3.8-27B & \cellcolor{royalblue!27}53.1 & \cellcolor{royalblue!6}0.0 & \cellcolor{royalblue!21}36.8 & \cellcolor{royalblue!28}59.1 & \cellcolor{royalblue!16}45.5 & \cellcolor{royalblue!6}82.3 & \cellcolor{royalblue!30}\underline{18.2} & \cellcolor{royalblue!14}11.0 & \cellcolor{royalblue!12}\underline{1} \\
 & Qwen3.5-9B & \cellcolor{royalblue!25}48.8 & \cellcolor{royalblue!6}0.0 & \cellcolor{royalblue!12}15.0 & \cellcolor{royalblue!24}50.0 & \cellcolor{royalblue!18}50.0 & \cellcolor{royalblue!26}91.9 & \cellcolor{royalblue!16}40.9 & \cellcolor{royalblue!12}8.1 & \cellcolor{royalblue!6}0 \\
 & Ours (9B SFT) & \cellcolor{royalblue!23}43.4 & \cellcolor{royalblue!6}0.0 & \cellcolor{royalblue!16}25.0 & \cellcolor{royalblue!28}59.1 & \cellcolor{royalblue!8}27.3 & \cellcolor{royalblue!38}\underline{97.9} & \cellcolor{royalblue!6}59.1 & \cellcolor{royalblue!13}9.0 & \cellcolor{royalblue!6}0 \\
\cmidrule{2-11}
 & \multicolumn{10}{l}{\textit{Ours: SFT+RL Curriculum:}} \\
 & \quad Easy Only (1--5 Hunks) & \cellcolor{royalblue!25}48.0 & \cellcolor{royalblue!6}0.0 & \cellcolor{royalblue!21}36.8 & \cellcolor{royalblue!26}54.5 & \cellcolor{royalblue!6}22.7 & \cellcolor{royalblue!33}95.2 & \cellcolor{royalblue!40}\textbf{0.0} & \cellcolor{royalblue!12}8.2 & \cellcolor{royalblue!6}0 \\
 & \quad Hard Only (30--34 Hunks) & \cellcolor{royalblue!30}59.6 & \cellcolor{royalblue!23}\underline{9.1} & \cellcolor{royalblue!34}66.8 & \cellcolor{royalblue!32}68.2 & \cellcolor{royalblue!22}59.1 & \cellcolor{royalblue!37}97.3 & \cellcolor{royalblue!40}\textbf{0.0} & \cellcolor{royalblue!21}19.1 & \cellcolor{royalblue!12}\underline{1} \\
 & \quad All Difficulties (1--5, 10--12, then 30--34) & \cellcolor{royalblue!40}\textbf{81.1} & \cellcolor{royalblue!40}\textbf{18.2} & \cellcolor{royalblue!40}\textbf{80.9} & \cellcolor{royalblue!40}\textbf{86.4} & \cellcolor{royalblue!40}\textbf{100.0} & \cellcolor{royalblue!35}96.5 & \cellcolor{royalblue!40}\textbf{0.0} & \cellcolor{royalblue!40}\textbf{40.8} & \cellcolor{royalblue!40}\textbf{6} \\
\midrule
\textit{Reference} & \textit{Clean repository (50 Bugs)} & -- & -- & -- & -- & -- & -- & -- & \textit{52.0} & -- \\
\midrule
\multirow{15}{*}{50 Bugs} & Claude Opus 5.5 & 77.8 & 9.1 & 54.9 & 81.8 & 90.9 & 99.1 & 0.0 & 43.9 & 9 \\
 & Claude Sonnet 5.5 & 64.3 & 0.0 & 86.3 & 86.4 & 100.0 & 96.0 & 0.0 & 30.6 & 2 \\
 & Claude Sonnet 5 & 30.2 & 0.0 & 49.0 & 59.1 & 90.9 & 96.6 & 0.0 & 3.5 & 0 \\
 & Claude Sonnet 4.6 & 0.0 & 0.0 & 6.2 & 0.0 & 72.7 & 91.2 & 0.0 & 0.0 & 0 \\
 & Gemini 3.8 Flash & 22.4 & 4.5 & 33.5 & 27.3 & 81.8 & 95.6 & 0.0 & 9.2 & 3 \\
 & Claude Haiku 5.5 & \cellcolor{royalblue!40}\textbf{41.3} & \cellcolor{royalblue!6}0.0 & \cellcolor{royalblue!34}54.6 & \cellcolor{royalblue!40}\textbf{68.2} & \cellcolor{royalblue!24}31.8 & \cellcolor{royalblue!40}\textbf{98.7} & \cellcolor{royalblue!40}\textbf{0.0} & \cellcolor{royalblue!40}\textbf{8.8} & \cellcolor{royalblue!40}\textbf{1} \\
 & Claude Haiku 4.5 & \cellcolor{royalblue!7}5.6 & \cellcolor{royalblue!6}0.0 & \cellcolor{royalblue!8}9.3 & \cellcolor{royalblue!11}9.1 & \cellcolor{royalblue!40}\textbf{59.1} & \cellcolor{royalblue!31}93.0 & \cellcolor{royalblue!40}\textbf{0.0} & \cellcolor{royalblue!26}\underline{5.5} & \cellcolor{royalblue!40}\textbf{1} \\
 & Gemini 3.5 Flash-Lite & \cellcolor{royalblue!9}7.0 & \cellcolor{royalblue!6}0.0 & \cellcolor{royalblue!6}6.5 & \cellcolor{royalblue!6}0.0 & \cellcolor{royalblue!35}\underline{50.0} & \cellcolor{royalblue!23}87.8 & \cellcolor{royalblue!37}\underline{9.1} & \cellcolor{royalblue!25}5.2 & \cellcolor{royalblue!40}\textbf{1} \\
 & Qwen3.8-27B & \cellcolor{royalblue!11}9.8 & \cellcolor{royalblue!6}0.0 & -- & \cellcolor{royalblue!17}\underline{22.7} & \cellcolor{royalblue!32}45.5 & \cellcolor{royalblue!6}77.2 & \cellcolor{royalblue!25}45.5 & \cellcolor{royalblue!11}1.6 & \cellcolor{royalblue!6}0 \\
 & Qwen3.5-9B & \cellcolor{royalblue!6}4.1 & \cellcolor{royalblue!6}0.0 & \cellcolor{royalblue!10}14.0 & \cellcolor{royalblue!11}9.1 & \cellcolor{royalblue!6}0.0 & \cellcolor{royalblue!27}90.2 & \cellcolor{royalblue!6}100.0 & \cellcolor{royalblue!6}0.5 & \cellcolor{royalblue!6}0 \\
 & Ours (9B SFT) & \cellcolor{royalblue!9}7.2 & \cellcolor{royalblue!6}0.0 & \cellcolor{royalblue!12}16.3 & \cellcolor{royalblue!13}13.6 & \cellcolor{royalblue!9}4.5 & \cellcolor{royalblue!37}\underline{96.8} & \cellcolor{royalblue!8}95.5 & \cellcolor{royalblue!10}1.5 & \cellcolor{royalblue!6}0 \\
\cmidrule{2-11}
 & \multicolumn{10}{l}{\textit{Ours: SFT+RL Curriculum:}} \\
 & \quad Easy Only (1--5 Hunks) & \cellcolor{royalblue!11}9.9 & \cellcolor{royalblue!6}0.0 & \cellcolor{royalblue!17}25.1 & \cellcolor{royalblue!11}9.1 & \cellcolor{royalblue!6}0.0 & \cellcolor{royalblue!36}96.2 & \cellcolor{royalblue!26}40.9 & \cellcolor{royalblue!26}\underline{5.5} & \cellcolor{royalblue!40}\textbf{1} \\
 & \quad Hard Only (30--34 Hunks) & \cellcolor{royalblue!13}12.3 & \cellcolor{royalblue!6}0.0 & \cellcolor{royalblue!36}\underline{58.0} & \cellcolor{royalblue!17}\underline{22.7} & \cellcolor{royalblue!11}9.1 & \cellcolor{royalblue!34}94.7 & \cellcolor{royalblue!15}72.7 & \cellcolor{royalblue!10}1.4 & \cellcolor{royalblue!6}0 \\
 & \quad All Difficulties (1--5, 10--12, then 30--34) & \cellcolor{royalblue!15}\underline{14.1} & \cellcolor{royalblue!6}0.0 & \cellcolor{royalblue!40}\textbf{64.9} & \cellcolor{royalblue!17}\underline{22.7} & \cellcolor{royalblue!6}0.0 & \cellcolor{royalblue!33}94.1 & \cellcolor{royalblue!29}31.8 & \cellcolor{royalblue!25}5.1 & \cellcolor{royalblue!6}0 \\
\bottomrule[1.5pt]
\end{tabular}
}
\end{table}
\FloatBarrier

\FloatBarrier
\subsection{Extended OOD Benchmarks}
\label{app:ood_full}

Table~\ref{tab:ood_results} expands the OOD evaluation of Section~\ref{sec:ood} with four additional benchmarks that progressively depart from repository-level bug repair, comparing our All Difficulties checkpoint, the same one reported in every other table, against the untuned 9B base model and Claude Sonnet~5 across evaluation tasks with zero training overlap.

\paragraph{Tasks.} In \textbf{code only}, the model generates a standalone training script directly from the competition description without tool access across up to ten attempts. In \textbf{solution.py}, the agent authors a single-file pipeline inside the interactive sandbox starting from an empty workspace, while \textbf{Repository} requires constructing a modular multi-file repository from scratch. All three ML synthesis settings evaluate across the 22 MLE-Bench Lite competitions using the official Kaggle grader and report mean leaderboard percentile. Finally, \textbf{DeepSWE} evaluates long-horizon software issue resolution across 34 general Python repository tasks.

\paragraph{LiveCodeBench~v6.} We evaluate under the same sampling parameters as \bmbl{} with reasoning traces capped at 4{,}096 tokens; responses from which no executable program can be extracted (13--25\% across models) are scored as incorrect.

\begin{table}[!htbp]
\centering
\caption{\textbf{OOD tasks.} Top rows of each task measure quality, bottom rows efficiency. Tokens: input plus output per episode. Shading and bold (best) compare the base model and ours. $\Delta$: paired per-task difference from the base; $\dagger$: its 95\% task-bootstrap interval excludes zero.}
\label{tab:ood_results}
\resizebox{\linewidth}{!}{
\begin{tabular}{@{}llcccr@{}}
\toprule[1.5pt]
 & & \textbf{Frontier} & \multicolumn{2}{c}{\textbf{Budget Models}} & \\
\cmidrule(lr){3-3} \cmidrule(lr){4-5}
Task & Metric & Claude Sonnet 5 & Qwen3.5-9B & Ours (9B SFT+RL) & $\Delta$ Ours $-$ Qwen3.5-9B \\
\midrule\midrule
\rowcolor{sectiongray}\multicolumn{6}{@{}l}{\textit{Code only (no tools, up to 10 debug attempts)} \quad {\footnotesize 66 episodes per model}} \\
\quad Quality & Leaderboard Percentile \hib & 61.2 & \cellcolor{royalblue!6}16.4 & \cellcolor{royalblue!40}\textbf{17.9} & +1.5 \\
\quad Quality & Valid Submission (\%) \hib & 100 & \cellcolor{royalblue!40}\textbf{61} & \cellcolor{royalblue!6}50 & $-$10.6 \\
\quad Efficiency & Input + Output Tokens / Ep. (k) \lob & 13 & \cellcolor{royalblue!6}78 & \cellcolor{royalblue!40}\textbf{65} & $-$12.7$^\dagger$ \\
\quad Efficiency & Tool-Call Success (\%) \hib & -- & -- & -- & -- \\
\midrule
\rowcolor{sectiongray}\multicolumn{6}{@{}l}{\textit{Sandbox: write \texttt{solution.py}} \quad {\footnotesize 22 episodes per model}} \\
\quad Quality & Leaderboard Percentile \hib & 73.9 & \cellcolor{royalblue!40}\textbf{23.8} & \cellcolor{royalblue!6}17.7 & $-$6.2 \\
\quad Quality & Valid Submission (\%) \hib & 100 & \cellcolor{royalblue!40}\textbf{68} & \cellcolor{royalblue!6}59 & $-$9.1 \\
\quad Efficiency & Input + Output Tokens / Ep. (k) \lob & 1022 & \cellcolor{royalblue!6}2112 & \cellcolor{royalblue!40}\textbf{1191} & $-$921.3$^\dagger$ \\
\quad Efficiency & Tool-Call Success (\%) \hib & 98.2 & \cellcolor{royalblue!6}74.9 & \cellcolor{royalblue!40}\textbf{92.7} & +17.4$^\dagger$ \\
\midrule
\rowcolor{sectiongray}\multicolumn{6}{@{}l}{\textit{Sandbox: build a repository} \quad {\footnotesize 22 episodes per model}} \\
\quad Quality & Leaderboard Percentile \hib & 72.0 & \cellcolor{royalblue!6}20.0 & \cellcolor{royalblue!40}\textbf{22.1} & +2.1 \\
\quad Quality & Valid Submission (\%) \hib & 100 & \cellcolor{royalblue!6}41 & \cellcolor{royalblue!40}\textbf{50} & +9.1 \\
\quad Efficiency & Input + Output Tokens / Ep. (k) \lob & 1213 & \cellcolor{royalblue!6}2259 & \cellcolor{royalblue!40}\textbf{1410} & $-$849.5$^\dagger$ \\
\quad Efficiency & Tool-Call Success (\%) \hib & 97.6 & \cellcolor{royalblue!6}69.9 & \cellcolor{royalblue!40}\textbf{87.6} & +18.5$^\dagger$ \\
\midrule
\rowcolor{sectiongray}\multicolumn{6}{@{}l}{\textit{Sandbox: DeepSWE issues} \quad {\footnotesize 34 episodes per model}} \\
\quad Quality & Resolved (\%) \hib & 14.7 & \cellcolor{royalblue!6}0.0 & \cellcolor{royalblue!6}0.0 & +0.0 \\
\quad Quality & Fail-to-Pass Tests (\%) \hib & 68.5 & \cellcolor{royalblue!6}3.3 & \cellcolor{royalblue!40}\textbf{12.2} & +8.9$^\dagger$ \\
\quad Efficiency & Input + Output Tokens / Ep. (k) \lob & 3668 & \cellcolor{royalblue!40}\textbf{956} & \cellcolor{royalblue!6}1382 & +425.9$^\dagger$ \\
\quad Efficiency & Tool-Call Success (\%) \hib & 96.6 & \cellcolor{royalblue!6}67.0 & \cellcolor{royalblue!40}\textbf{95.1} & +28.2$^\dagger$ \\
\bottomrule[1.5pt]
\end{tabular}
}
\end{table}

\paragraph{RL improves tool-call reliability across interactive environments.} In both ML sandbox settings, tool-call success rises by 17 to 19 points over the untuned base model, and tokens per episode fall by about 40\% (Table~\ref{tab:ood_results}). On DeepSWE, tool-call success rises by 28 points. Tokens per episode rise there by 45\%, because our model wastes 9 fewer turns per episode on malformed calls and each valid call returns tool output.

\paragraph{From-scratch ML synthesis shows mixed leaderboard results.} Our model edges ahead of the untuned base model in standalone-script ($+1.5$) and full-repository ($+2.1$) synthesis while trailing on single-file \texttt{solution.py} ($-6.2$), with all differences within 7 points and none statistically significant, whereas Claude Sonnet~5 demonstrates substantial headroom (61--74th percentile). Consistently improving open-ended ML synthesis from scratch likely requires direct optimization on synthesis objectives, such as leaderboard-percentile rewards.

\end{document}